\documentclass{article}

\usepackage[preprint]{neurips_2026}
\usepackage{tabularx, booktabs}
 
\usepackage{amsmath,amssymb}

\usepackage[utf8]{inputenc} % allow utf-8 input
\usepackage[T1]{fontenc}    % use 8-bit T1 fonts
\usepackage{hyperref}       % hyperlinks
\usepackage{url}            % simple URL typesetting
\usepackage{booktabs}       % professional-quality tables
\usepackage{amsfonts}       % blackboard math symbols
\usepackage{nicefrac}       % compact symbols for 1/2, etc.
\usepackage{microtype}      % microtypography
\usepackage{xcolor}         % colors
\usepackage[most]{tcolorbox}

\usepackage{tabularx}
\usepackage{amsmath,amssymb}
\usepackage{graphicx}
\usepackage[utf8]{inputenc}
\usepackage[T1]{fontenc}
\usepackage{hyperref}
\usepackage{url}
\usepackage{algpseudocode}
\usepackage{booktabs}
\usepackage{amsfonts}
\usepackage{algorithm}
\usepackage{nicefrac}
\usepackage{microtype}
\usepackage{xcolor}
\usepackage[most]{tcolorbox}

\usepackage{amsthm}
\theoremstyle{definition}

\theoremstyle{plain}

\usepackage[capitalise,noabbrev]{cleveref}

\usepackage{titletoc}

\newtcolorbox{researchquestion}[2][]{%
  enhanced,
  colback=gray!8,
  colframe=gray!45,
  coltitle=black,
  fonttitle=\bfseries,
  title=#2,
  boxrule=0.6pt,
  arc=2mm,
  left=3mm,right=3mm,top=2mm,bottom=2mm,
  attach title to upper,
  boxed title style={
    colback=gray!18,
    colframe=gray!45,
    boxrule=0.6pt,
    arc=2mm,
    left=2mm,right=2mm,top=1mm,bottom=1mm
  },
  after title={\par\noindent{\color{gray!45}\rule{\linewidth}{0.6pt}}\par\vspace{4pt}},
  #1
}

\usepackage{tikz}
\usetikzlibrary{positioning, fit, backgrounds, calc, arrows.meta, shapes.geometric}

\usepackage{colortbl}   % \rowcolor for table-row highlighting
\usepackage{overpic}    % hyperlink overlays on the hero figure
\usepackage{wrapfig}    % text wrapping around gen_figure
\definecolor{almCoreText}{HTML}{70451A}
\definecolor{almEditText}{HTML}{1A3E70}
\definecolor{almEditMid}{HTML}{3C649A}
\definecolor{almGenText}{HTML}{701A69}
\definecolor{almGenMid}{HTML}{9A3C93}
\definecolor{almCite}{HTML}{0071BC}
\definecolor{almSlate}{HTML}{5E757C}
\definecolor{almCoreBg}{HTML}{F5EDE3}
\definecolor{almEditBg}{HTML}{E4ECF5}
\definecolor{almGenBg}{HTML}{F5E4F4}
\colorlet{almCoreFill}{almCoreText!20}
\colorlet{almCoreItem}{almCoreText!9}
\colorlet{almGenFill}{almGenText!20}
\colorlet{almGenItem}{almGenText!9}
\colorlet{almEditFill}{almEditText!20}
\colorlet{almEditItem}{almEditText!9}

\colorlet{almLinkBox}{almEditText!50}
\colorlet{almCiteBox}{almCite!50}
\hypersetup{
  colorlinks=false,
  linkbordercolor=almLinkBox,
  citebordercolor=almCiteBox,
  urlbordercolor=almCiteBox,
  filebordercolor=almCiteBox,
  pdfborder={0 0 0.7},
}

\newcommand{\secbadge}[2]{\hyperref[#2]{\textbf{\textcolor{#1}{\S}\textcolor{almCite}{\ref*{#2}}}}}

\newcommand{\figseclink}[2]{\hyperref[#2]{\scriptsize\bfseries\textcolor{#1}{\S\,}\textcolor{almCite}{\ref*{#2}}}}
\newcommand{\figtablink}[2]{\hyperref[#2]{\scriptsize\bfseries\textcolor{#1}{Tab.\,}\textcolor{almCite}{\ref*{#2}}}}

\makeatletter
\def\alm@cite@opt[#1]#2{\textcolor{almCite}{\almorig@cite[#1]{#2}}}
\def\alm@cite@noopt#1{\textcolor{almCite}{\almorig@cite{#1}}}
\AtBeginDocument{%
  \NewCommandCopy\almorig@cite\cite  % full copy (robust-safe) so no self-recursion
  \DeclareRobustCommand{\cite}{\@ifnextchar[{\alm@cite@opt}{\alm@cite@noopt}}%
}
\makeatother

\titlecontents{section}[0em]
  {\addvspace{6pt}\bfseries}
  {\textcolor{almCite}{\thecontentslabel}\hspace{0.75em}}
  {}
  {\hspace{0.5em}\titlerule*[0.6pc]{.}\contentspage}
  [\addvspace{2pt}]
\titlecontents{subsection}[1.9em]
  {\addvspace{1pt}}
  {\textcolor{almCite}{\thecontentslabel}\hspace{0.6em}}
  {}
  {\hspace{0.5em}\titlerule*[0.6pc]{.}\contentspage}
\titlecontents{subsubsection}[4.2em]
  {\addvspace{0.25pt}\small}
  {\textcolor{almCite}{\thecontentslabel}\hspace{0.6em}}
  {}
  {\hspace{0.5em}\titlerule*[0.6pc]{.}{\small\contentspage}}

\newcommand{\almGen}{\textbf{ALM Gen}}      % the DNG model (MatterGen-base backbone)
\newcommand{\almEdit}{\textbf{ALM Edit}}    % the CSP model (broader\_csp backbone)
\newcommand{\ci}[1]{\,{\scriptscriptstyle\pm#1}}

\title{Atomistic Language Models \\ Understand and Generate Materials}

\author{%
  Sathya Edamadaka$^{1}$, \;\,
  Krithik Ramesh$^{2}$, \;\,
  Ju Li$^{1}$, \;\,
  Rafael G\'omez-Bombarelli$^{1,3*}$
  \\[0.6ex]
  $^{1}$Massachusetts Institute of Technology, \quad
  $^{2}$Lyra Labs, \quad
  $^{3}$Lila Sciences
  \\[0.3ex]
  $^{*}$Correspondence to \texttt{rafagb@mit.edu}
}

\begin{document}
\vspace{-10mm}
\maketitle
\vspace{-10mm}
\begin{figure}[htp!]
  \centering
  \begin{overpic}[trim={9.5cm 13cm 14.5cm 12cm},clip,width=1\linewidth]{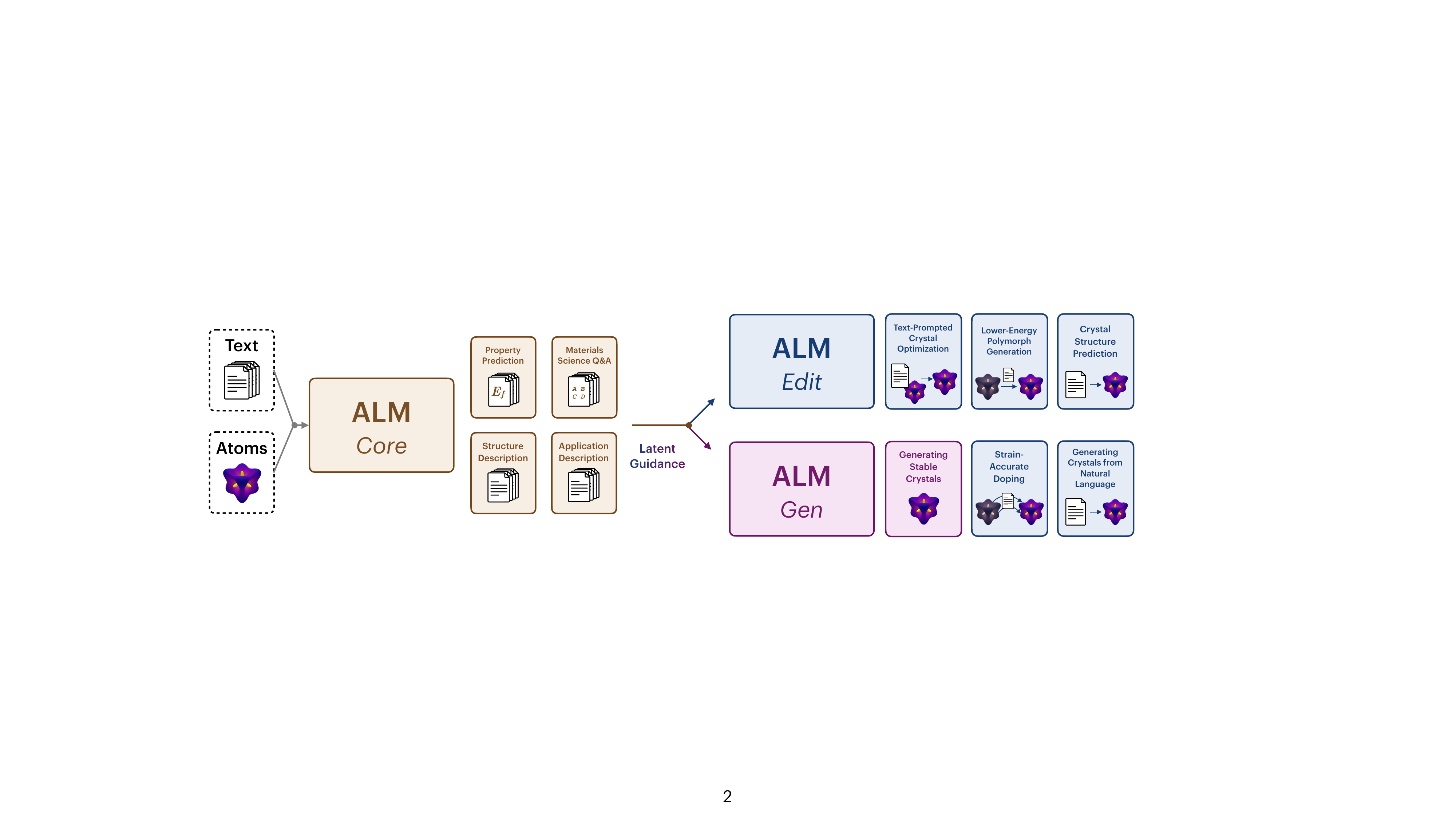}
    % overpic default coords: x in 0..100 (% width); the y unit is ALSO 1% of the WIDTH,
    % so the image top is at y = 100*(h/w) ~= 29.97. y-values below = (%height) x 0.2997.
    % Capability boxes -> their Methods section (top-right corner of each box).
    \put(22.5,15.2){\figseclink{almCoreText}{sec:methods_core}}            % ALM Core -> Methods (ALM Core)
    \put(67.3,22){\figseclink{almEditText}{sec:generator}}               % ALM Edit -> Methods (ALM Edit)
    \put(67.3,8.4){\figseclink{almGenText}{sec:methods_dng}}               % ALM Gen  -> Methods (ALM Gen)
    % Understanding task boxes (above) + the latent-guidance bridge (methods).
    \put(34.5,22.3){\figseclink{almCoreText}{sec:results_understanding}}   % above brown task boxes
    \put(46.5,10.5){\figseclink{almSlate}{sec:generator}}                  % Latent Guidance -> bridge (methods)
    % Edit (CSP) task boxes -> their result sections / tables.
    \put(74.75,24.4){\figseclink{almEditText}{sec:results_mode4}}           % Text-prompted optimization
    \put(83.75,24.4){\figseclink{almEditText}{sec:results_mode4}}    % Lower-energy polymorph
    \put(92.75,24.4){\figseclink{almEditText}{sec:results_csp}}           % Crystal structure prediction
    % Gen (DNG) task boxes -> their result sections / tables.
    \put(74.75,10.9){\figseclink{almGenText}{sec:results_dng}}          % Generating stable crystals
    \put(83.75,10.9){\figseclink{almGenText}{sec:results_generation}}     % Strain-accurate doping
    \put(92.75,10.9){\figseclink{almGenText}{sec:results_generation}}            % Crystals from natural language
  \end{overpic}
  \caption{\textbf{Atomistic Language Models bridge natural language and 3D atomic coordinates to \textit{understand, generate, and optimize materials}.}  This new paradigm allows a single autoregressive backbone to characterize the structure, properties, and applications of a material, as well as guide the discovery of new ones, all without lossy text representations.}
  \label{fig:hero}
\end{figure}
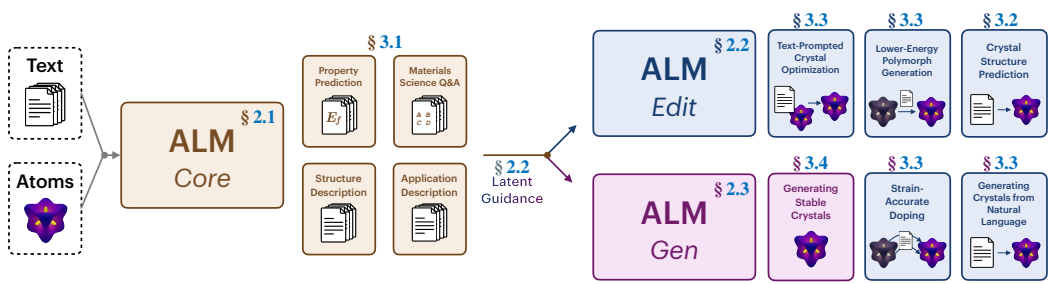

\begin{abstract}
Atomistic structure and natural language have long been modeled separately, with language models either calling atomistic models as tools or being fine-tuned on lossy textual encodings that discard atomistic information. We introduce \textbf{Atomistic Language Models (ALMs)} to pursue native multimodality, in which a single language backbone understands atomistic structures, generates materials from natural language, and optimizes crystal structures as instructed by text. By unifying a pretrained atomistic encoder, large language model, and denoising diffusion model through purely continuous projectors and staged training, ALMs achieve state-of-the-art results on crystal structure prediction and \textit{de novo} generation. ALMs are enabled by a \textbf{continuous bridge that maps language model embeddings directly into the steering space of atomistic diffusion}, and are assisted by \textbf{Text-to-Crystal Feynman--Kac (T2C-FK)}, a particle-based sampler that scores partial denoising trajectories to enforce stoichiometric targets at inference time. To evaluate the ability of ALMs to optimize and generate materials from natural-language prompts and 3D atom-coordinate inputs, we introduce \textbf{ALM Bench}, the first benchmark for text-conditioned crystal generation and optimization. Code, training data, and model weights will be released soon.
\end{abstract}

\section{Introduction}
\label{sec:intro}

Domain experts in materials science reason in two modes. One is \textit{continuous} and geometric: atomic coordinates, a periodic unit cell, or the actual structure on which physical laws act. The other is \textit{discrete} and linguistic: a phase, a processing history, or the experimental context that determines whether and how the material can be made. An ideal model of materials would move fluently between these modalities, parsing a structure to predict its properties, generating a structure from a natural language specification, and optimizing materials exactly as desired.

The two modalities carry fundamentally different representational biases. State-of-the-art (SoTA) atomistic models~\cite{wood_uma_2025, batatia_foundation_2024,rhodes_orb-v3_2025} operate on graphs built from 3D coordinates and periodic boundary conditions, many having inductive priors like symmetry, locality, and stoichiometry baked into their architectures. Language models operate on discrete tokens with no native notion of geometry or periodicity. Beneath these surface incompatibilities lies a deeper one: in matter, distances in real space and distances in semantic space are decoupled. A perturbation of a few atoms can leave a crystal structure nearly unchanged geometrically while moving it across a phase boundary; different polymorphs of the same composition can differ by less than an angstrom in coordinates and by orders of magnitude in conductivity or hardness. Each modality must therefore be modeled in its own, native representation. Previous approaches bridge this gap by training on lossy textual encodings of materials, like CIF files~\cite{antunes_crystal_2024}, Wyckoff strings~\cite{plaid_xu_2025}, or other custom string representations~\cite{gruver_fine-tuned_2024,alampara2024mattext}. However, they lose crucial geometric context that determines material behavior~\cite{gupta_matscibert_2021,edamadaka_universally_2025} or catastrophically forget their natural language abilities \cite{ozawa_graph-text_2024}. 

The natural next step is to compose pipelines of unimodal specialists: large language models (LLMs) call structure generators as tools~\cite{lu_towards_2026,xie_crystal_2022, jiao_crystal_2023, zeni_generative_2025}, generative models post-process language model outputs~\cite{yang_generative_2024}, and property predictors are wrapped behind text interfaces~\cite{deng2026harnessingatomisticskillsagenticatomistic}. Such pipelines inherit \textit{three failure modes}: components communicate only through discrete surface forms, discarding continuous geometric information; they train independently, so structure and language never develop a shared latent space; and generation can only be steered through what a text prompt can express, not through the latent space the model uses to reason. On the other hand, multimodal language models in literature that directly understand 3D atomic coordinates are purely property predictors, and cannot generate new materials~\cite{moro_multimodal_2025,suzuki_bridging_2025,tang_multimodal_2026,cui_l2m3of_2025}.

A different paradigm avoids these failure modes. Atomistic encoders, language models, and atomistic structure generators each excel and fail at different predictive and generative tasks. Combining them into a single, end-to-end model with shared latent spaces addresses those failure modes more naturally than any pipeline can; representations are shared~\cite{liu_visual_2023}, gradient signals flow across modality boundaries during training~\cite{chen_janus-pro_2025}, and conditional generation can be steered through the same latent space the model uses to reason~\cite{li_blip2_2023}. 

\begin{figure}[t!]
  \centering
  \includegraphics[trim={13cm 4.25cm 5.5cm 4cm},clip,width=1\linewidth]{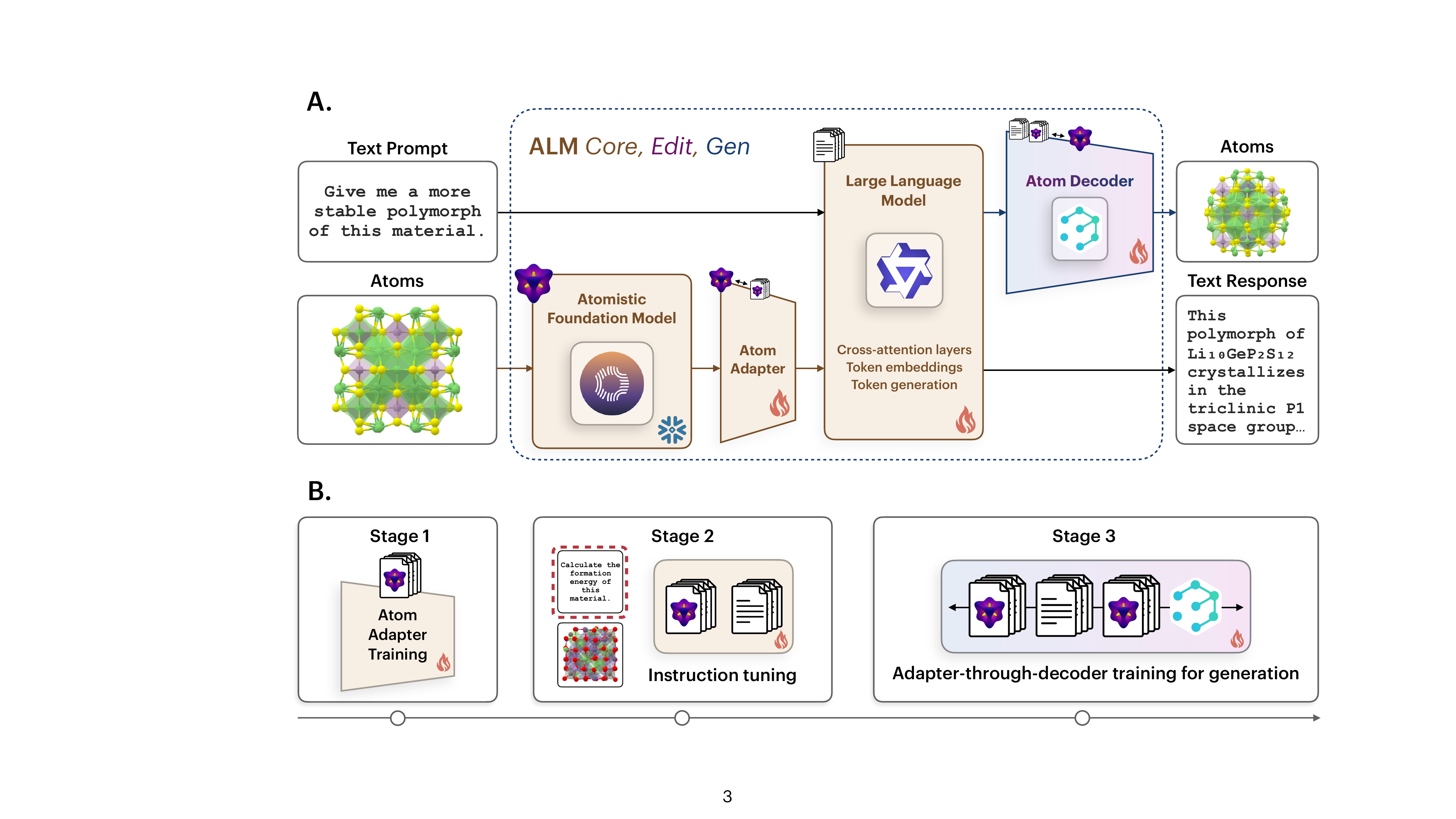}
  \caption{\textbf{Atomistic Language Models understand atoms as soft tokens from a machine learning interatomic potential and generate inorganic crystals by steering diffusion models with classifier-free guidance.} A. ALMs are comprised of an MLIP encoder, LLM, and diffusion decoder, unified by continuous projectors. B. Staged curriculum training which progressively unfreezes the model and instruction-tunes it, enabling property prediction and structure generation in the same model.}
  \label{fig:arch_and_training}
\end{figure}

The \textbf{Atomistic Language Models (ALMs)} realize this paradigm, unifying a pretrained atomistic encoder~\cite{rhodes_orb-v3_2025}, language model~\cite{yang_qwen3_2025}, and denoising diffusion decoder~\cite{zeni_generative_2025} through continuous, cross-modal projectors (Fig.~\ref{fig:arch_and_training}). ALMs are trained in three stages: \textit{alignment} of the modalities, \textit{multi-pattern instruction tuning} that instills structural reasoning over both text and structure, and \textit{guided generation} in which embeddings extracted from the language model steer the denoising trajectory. \textbf{ALM Core} is a language model specialized in characterizing materials structure, properties, and applications by natively taking in 3D atomic coordinates, lattice parameters, and element types. \textbf{ALM Edit} is finetuned on \textbf{Core}, learning to strongly steer~\cite{ho_classifier-free_2022} a denoising diffusion decoder built for crystal structure prediction, and \textit{is capable of optimizing a crystal as instructed by natural language}. \textbf{ALM Gen} replaces this decoder with a weakly steered diffusion model, enabling \textit{de novo} generation of inorganic crystals. To bridge the gap between the instruction-following abilities of \textbf{ALM Edit} and the competitive stability of \textbf{ALM Gen}'s generated materials, we introduce an inference-time steering method, \textbf{Text-to-Crystal Feynman--Kac (T2C-FK)}, to improve how closely \textbf{ALM Gen} follows stoichiometric targets when prompted. To evaluate the text-conditioned optimization capabilities of \textbf{ALM Edit}, we introduce \textbf{ALM Bench}, comprising over 7,000 natural language and inorganic crystal instruction pairs for evaluating language instruction-following in materials discovery.

\textbf{ALM Core} matches or outperforms prior language model approaches across measured property prediction tasks while matching the performance of finetuned machine learning interatomic potential (MLIPs)~\cite{rubungo_llm4mat-bench_2024,tang_multimodal_2026}. \textbf{ALM Edit} achieves state-of-the-art crystal structure prediction performance on MP-20~\cite{xie_crystal_2022} and MPTS-52~\cite{jiao_crystal_2023} while beating all frontier language model baselines on \textbf{ALM Bench}, and \textbf{ALM Gen} achieves competitive performance on \textit{de novo} generation, using MP-20 or LeMat-GenBench~\cite{betala_lemat-genbench_2026} hulls to determine stability, while beating previous language-based approaches.

% real architecture + training pipeline that enables this. 

%
%
%
%
% METHODS
%
%
%
%
%
\section{Methods}
\label{sec:methods}

Atomistic Language Modeling couples three components through trained, continuous projectors (Fig.~\ref{fig:arch_and_training}A): a machine-learning interatomic potential that encodes each atom of a crystal ($\mathcal{E}$, OrbV3~\cite{rhodes_orb-v3_2025}), a causal language model that serves as the central backbone ($\phi$, Qwen3-8B~\cite{yang_qwen3_2025}), and a denoising diffusion model that decodes the language model's latent instructions into 3D crystal structures ($\mathcal{D}$, MatterGen~\cite{zeni_generative_2025}). \textbf{ALM Core} (\secbadge{almCoreText}{sec:methods_core}) is the instruction-tuned core that can take in crystals as represented by continuous unit cell parameters $\mathbf{L}$, continuous fractional coordinates $\mathbf{X}$ of each of the atoms in the material, and a discrete atomic-number assignment $\mathbf{A}$ that corresponds to their element types (an exhaustive list of notation is available in Table~\ref{tab:app_gen_notation}). Two generative variants bridge this base to diffusion decoders: a strongly text-conditioned crystal structure prediction model (\textbf{ALM Edit}, \secbadge{almEditText}{sec:generator}) and a weakly text-conditioned \emph{de novo} generator (\textbf{ALM Gen}, \secbadge{almGenText}{sec:methods_dng}). 

Standard cross-modal interfaces discretize each modality~\cite{alayrac_flamingo_2022,liu_visual_2023,chen_janus-pro_2025} into codebooks generated autoregressively at inference time~\cite{oord_neural_2018}. Crystalline matter resists both ideas: TiO$_2$ polymorphs differ by small distances between atoms, yet have band gaps that are hundreds of meV apart, and single-atom defects in large supercells shift formation energy without moving many structural fingerprints~\cite{drautz_atomic_2019}. Therefore, all latent representations are kept continuous in ALMs, with no rounding or vector quantization (Appendix~\ref{app:codebook_comparison}). Aligning the encoder, backbone, and decoder in this single, continuous latent space also confines all domain-specific inductive biases to the encoder $\mathcal{E}$ and decoder $\mathcal{D}$. Therefore, although this work tackles crystalline inorganic materials, ALMs can extend to other types of matter by retraining on new structures or utilizing different encoders and decoders. 

\subsection{ALM Core: understanding materials using their native 3D structure}
\label{sec:methods_core}

\textbf{ALM Core} is an instruction-tuned multimodal language model that reads each material's atomic coordinates, lattice parameters, and element types, answering in natural language and serving as the foundation for our generative variants (\textbf{Edit} and \textbf{Gen}). The frozen encoder $\mathcal{E}$ maps each material ($\mathbf{L}$, $\mathbf{X}$, $\mathbf{A}$) to per-atom embeddings $\mathbf{H}\in\mathbb{R}^{N_p\times d_{\mathcal{E}}}$, which a two-layer GELU MLP $P_{\text{in}}$ projects into the LLM's token space as soft tokens, a lower-parameter method than the gated attention-based bridges of prior multimodal materials models~\cite{tang_multimodal_2026}. The model is trained in two stages (Fig.~\ref{fig:arch_and_training}B, 1-2). $P_{\text{in}}$ is first aligned alone on deterministic structural descriptions~\cite{ganose_robocrystallographer_2019} from LLM4Mat-Bench~\cite{rubungo_llm4mat-bench_2024}. The LLM is then instruction-tuned via LoRA~\cite{hu_lora_2021} on those descriptions, as well as narratives about applications, property prediction, and three text-only tasks that prevent catastrophic forgetting~\cite{liu_voxtral_2025}. Loss functions, bucket weights, optimizer settings, and ablations are in Appendix~\ref{app:training_objectives} and ~\ref{app:training_stage1}.

% perfect uptil here. not reviewed after. 
\subsection{ALM Edit: text-conditioned crystal generation}
\label{sec:generator}

\textbf{ALM Edit} finetunes \textbf{ALM Core} to enable inverse materials design, guided by prompts consisting of natural language \textit{and} materials. This unlocks text-conditioned structural optimization, in which an inputted material is edited according to textual instructions. MatterGen serves as the denoising diffusion model $\mathcal{D}$ for ALMs due to its extensibility to different conditioning heads via classifier-free guidance~\cite{zeni_generative_2025} (Appendix~\ref{app:generator_formal}). To build \textbf{ALM Edit}, $\mathcal{D}$ was trained from scratch with one architectural modification: denoising only over atom coordinates $\mathbf{X}$ and unit cell parameters $\mathbf{L}$.
\begin{wrapfigure}[31]{r}{0.46\linewidth}
  \centering
  \includegraphics[trim={29.5cm 11.5cm 18cm 1cm},clip,width=\linewidth]{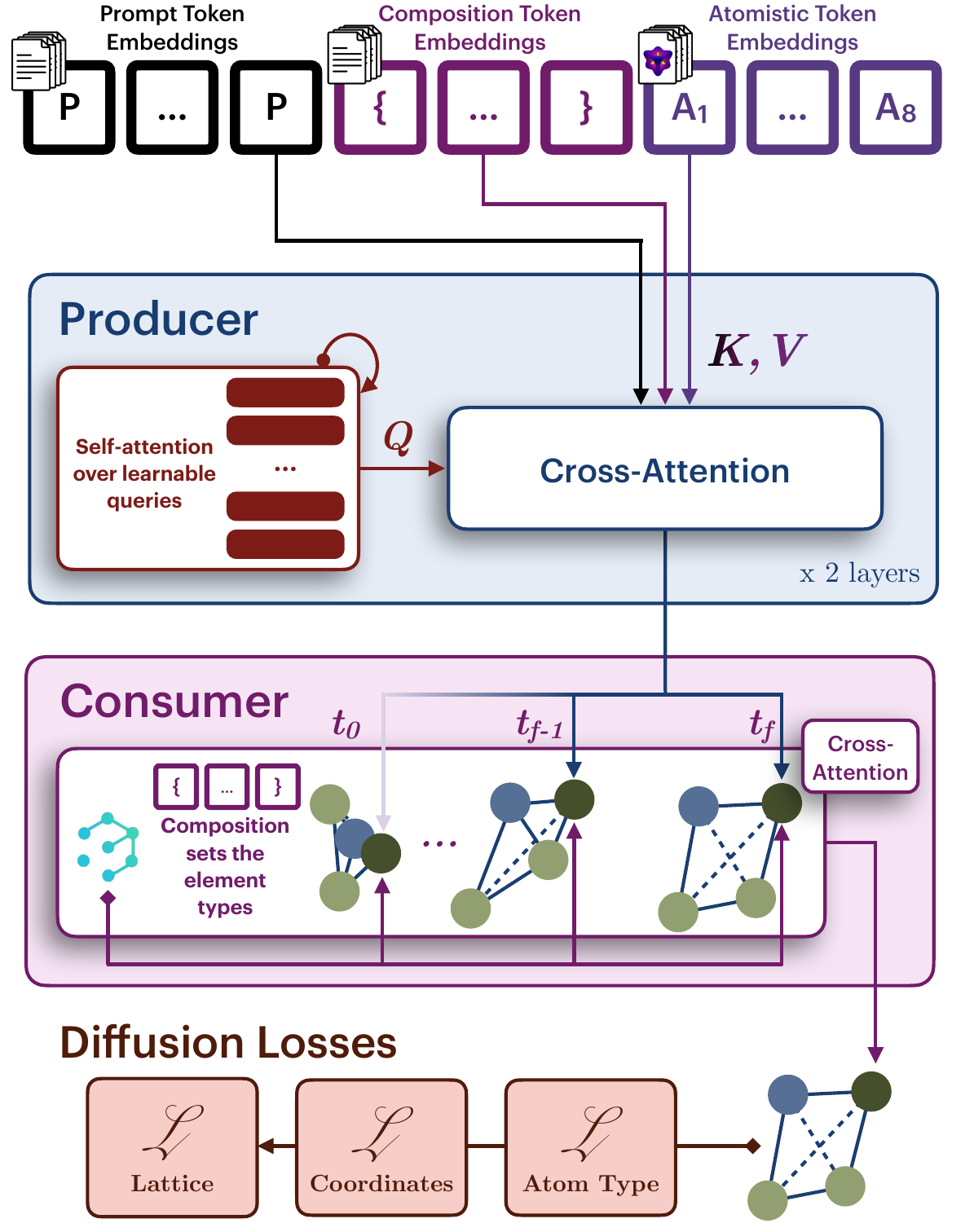}
  \caption{\textbf{A language-to-atomistic bridge enables the steering of crystal generation.} \textbf{ALM Edit} uses all components above. \textbf{ALM Gen} swaps the Q-Former-style~\cite{li_blip2_2023} producer for a lightweight per-token MLP (no learned queries or prompt context) feeding the same consumer, and does not emit composition embeddings.}
  \label{fig:gen_figure}
\end{wrapfigure}
% \vspace{2mm}

The decoder $\mathcal{D}$ \textit{observes} atomic-number assignment $\mathbf{A}$, initializing each node accordingly and never changing any atom's element. MatterGen cannot reliably reproduce desired stoichiometries and positions provided through CFG (discussed in \S\ref{sec:methods_dng} and Appendix~\ref{app:cfg_tension}), so this choice ensures \textbf{Edit} produces the structures it was prompted to generate. Therefore, the language model autoregressively generates the composition (a JSON of element types and counts) that $\mathcal{D}$ observes, and the model is pretrained on MP-20, MPTS-52, and all other structures that \textbf{Core} was trained to understand (Appendix~\ref{app:generation_data}). 

We introduce a two-piece, \emph{producer--consumer} bridge (Fig.~\ref{fig:gen_figure}, Eq.~\ref{eq:gen_short_combined}) to connect \textbf{Core} to this composition-observing decoder. To encode the textual task and structural information about the inputted material, $K=8$ "atomistic" tokens ($A_1$ through $A_8$ in Fig.~\ref{fig:gen_figure}) are teacher-forced onto the LLM's response and causally attend to the prompt and inputted material's soft tokens. Their final-layer hidden states $\mathbf{Z}\in\mathbb{R}^{K\times d_{\mathrm{LM}}}$ are extracted as continuous latents to guide $\mathcal{D}$ without passing through a discrete-vocabulary bottleneck~\cite{alayrac_flamingo_2022,liu_visual_2023,chen_janus-pro_2025}. Specifically, the \emph{producer} cross-attends over the $K$ atomistic, outputted composition, and text token embeddings, amplifying the information needed to generate the desired crystal. The \emph{consumer} then injects the fixed-size conditioning vector $\mathbf{C}$ into every block of $\mathcal{D}$'s score network through a cross-attention branch as part of CFG, leaving the base model frozen. 

During training (Stage 3 in Fig.~\ref{fig:arch_and_training}B), $\mathbf{\mathcal{D}}$'s \textbf{denoising diffusion losses backpropagate through the consumer, the producer, and the language model's atomistic-token hidden states} $\mathbf{Z}$. Additional losses, like per-element composition-count, anchor $\mathbf{Z}$ to a structurally meaningful direction; without this auxiliary, the atomistic token hidden states collapse to a near-constant direction across prompts (Appendix~\ref{app:aux_losses}). \textbf{ALM Edit} is trained on seven buckets of tasks, including crystal structure prediction and all tasks in \textbf{ALM Bench} (listed in Appendix~\ref{app:generation_data}). 

\subsection{ALM Gen: \textit{de novo} crystal generation}
\label{sec:methods_dng}

While conditional models like \textbf{ALM Edit} generate crystals as instructed by text, \textit{de novo} generative models that unconditionally generate large numbers of stable, novel crystals are essential for screening campaigns to discover new materials with desirable properties. \textbf{ALM Gen} unlocks \textit{de novo} generation by relaxing the strong conditioning achieved by \textbf{ALM Edit}. The producer (Fig.~\ref{fig:gen_figure}) is replaced by a lightweight per-token MLP. Each of the $K$ atomistic-token hidden states is projected independently into a $K$-token conditioning sequence, with no learned queries and no prompt-context window, feeding feeds the same consumer as \textbf{Edit}. In addition, the LLM no longer produces compositions when prompted to generate structures (Appendix~\ref{app:cfg_tension}). This is deliberately \emph{weak} conditioning, causing text prompts to bias, rather than dictate, sampled structures. \textbf{ALM Gen} is trained on the same data as \textbf{Edit} in different concentrations (Appendix~\ref{app:generation_data}), sampling input textual prompts at inference time from the respective eval partitions to generate crystals.

\subsection{T2C-FK: Steering generation with Feynman--Kac}
\label{sec:methods_steering}

\begin{wrapfigure}{r}{0.46\linewidth}
  \centering
  \includegraphics[width=\linewidth]{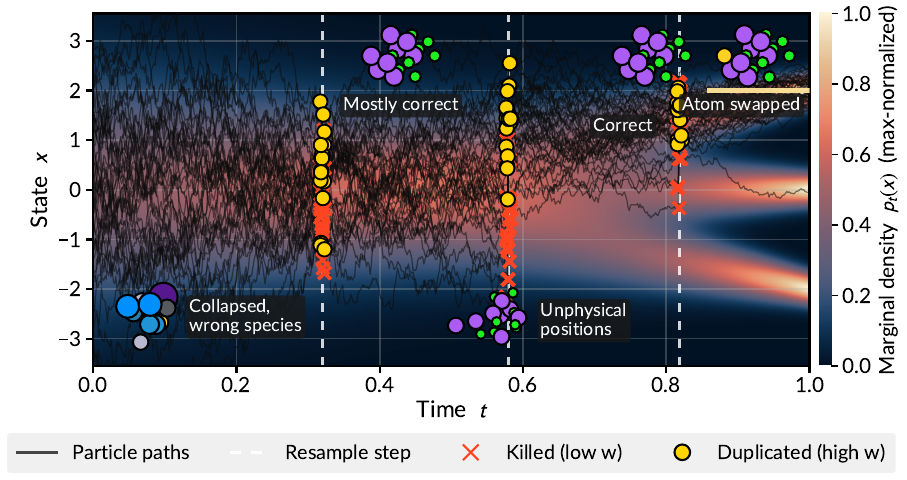}
  \caption{\textbf{Text-to-Crystal Feynman--Kac (T2C-FK) enables ALM Gen, a \textit{de novo} model, to generate structures with desired element sets and stoichiometry ratios.} A. Unphysical structures are removed throughout sampling, and any differences from the reference stoichiometry are fixed at the last step via Hungarian scoring.}
  \label{fig:t2c-fk}
\end{wrapfigure}
\textbf{ALM Edit} is designed to output a material with the desired element set and stoichiometry ratio. \textbf{ALM Gen}, on the other hand, is designed to produce more stable structures, but not necessarily with the exact stoichiometry used to prompt the model. This dichotomy arises from MatterGen's denoising, not the inherent ALM architecture (Appendix~\ref{app:reward_definitions}). We introduce \textbf{Text-to-Crystal Feynman--Kac} (\textbf{T2C-FK}) to close this gap at inference time, intercepting and reweighting denoising trajectories without retraining.

\textbf{T2C-FK} replaces MatterGen's single denoising trajectory with an $N$-particle bootstrap Sequential Monte Carlo sampler~\cite{wu_practical_2023, singhal_general_2025}: every $S$ steps, it reweights and resamples particles by a reward on the Tweedie-estimated clean structure $\hat{x}_0$, deferring scoring until the atomic-number distribution leaves its high-noise regime. The reward scores stoichiometric agreement of the score network's per-atom element distributions to the target multiset, and a final Hungarian override snaps each atom to its assigned target element, leaving lattice and coordinates untouched. A real example of \textbf{T2C-FK} is shown in Fig.~\ref{fig:t2c-fk}, and the full sampler (Algorithm~\ref{alg:t2c_fk}) with its posterior-correction guarantee, three reward components, potential function, and hyperparameter sweeps is given in Appendix~\ref{app:fk_details}.

\section{Results}

\begin{figure}[t!]
  \centering
  \includegraphics[width=1\linewidth]{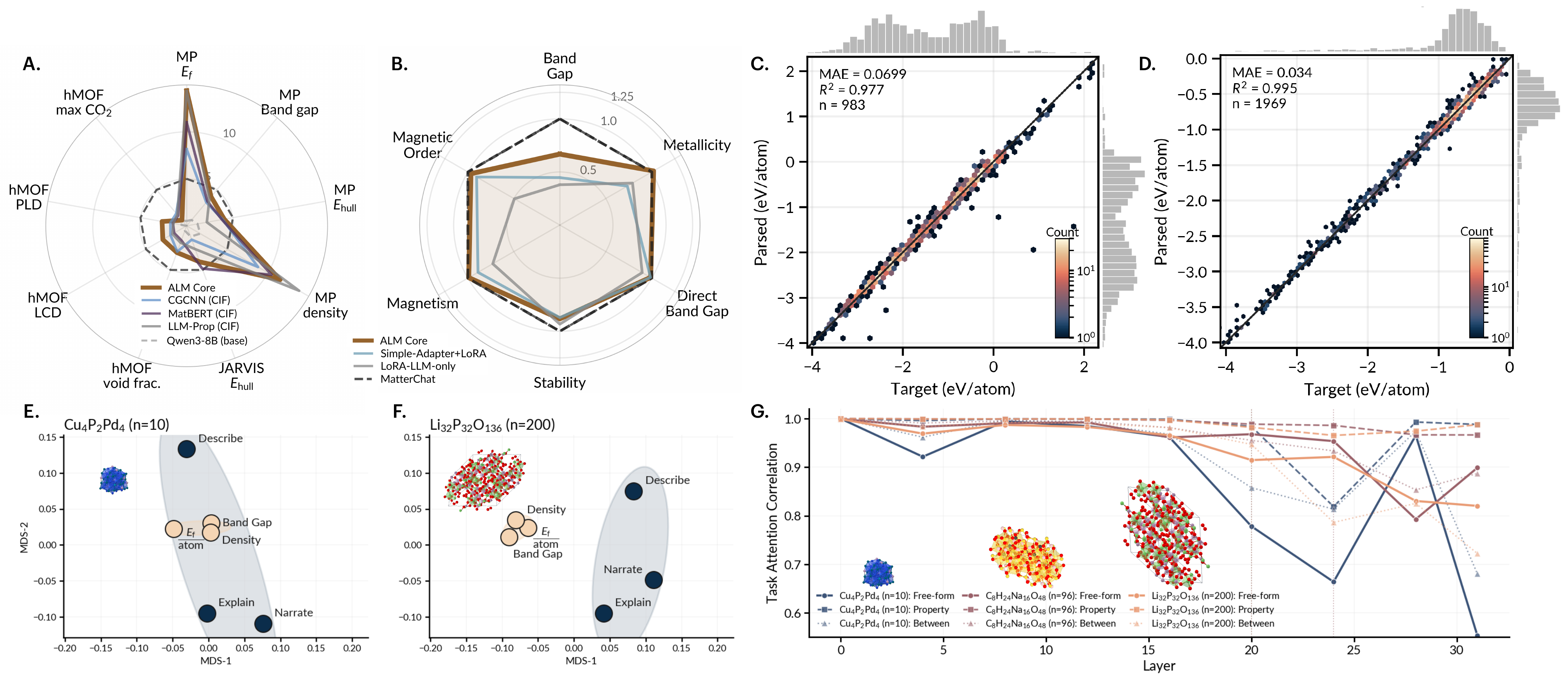}
  \caption{\textbf{Atomistic Language Models can accurately predict physical properties of materials.} Spider performance plots for selected materials property prediction tasks from A. LLM4Mat-Bench~\cite{rubungo_llm4mat-bench_2024} (MAD/MAE, with a performance threshold of $\geq 5$) and B. MatterChat (MAE, baseline from and normalized to \cite{tang_multimodal_2026}). Parity plots are shown for formation energy per atom ($E_f$) using C. Materials Project data \cite{Jain2013} and D. GNoME data \cite{merchant_scaling_2023}. We then show how similar activations are for property prediction (tan) and natural language (dark blue) tasks for E. small and F. large structures. G. visualizes how LLM attention weights differ across tasks and across materials input sizes through the LLM transformer layers.}
  \label{fig:understanding}
  \vspace{-3mm}
\end{figure}

By instruction tuning across several tasks, Atomistic Language Models both natively understand materials and can lift their knowledge into text-conditional materials generation. \textbf{ALM Core} matches or beats unimodal predictors on property prediction, \textbf{ALM Edit} achieves state-of-the-art at crystal structure prediction, and \textbf{ALM Gen} beats even atomistic models performance at \textit{de novo} generation. \textbf{ALM Edit} also beats all frontier model baselines on \textbf{ALM Bench}, designed to rigorously test text-conditioned materials optimization.

\subsection{Increased property prediction performance and training token efficiency}
\label{sec:llm4mat}
\label{sec:matterchat}
\label{sec:results_understanding}

% make way more lightweight. instead of "rows", just list the properties.
On LLM4Mat-Bench~\cite{rubungo_llm4mat-bench_2024}, an extensive crystal property prediction benchmark, \textbf{ALM Core} is one of the first natural language models to outperform previously published GNN baselines on Materials Project (MP) formation energy per atom, MP bandgap, MP density, and JARVIS-DFT energy above hull (Table~\ref{tab:stage2-llm4mat-mae}, Fig.~\ref{fig:understanding}A, C, D), breaking the so-called ``GNN--LLM wall''~\cite{rubungo_llm4mat-bench_2024}. The highest-performing language model baselines, in contrast to \textbf{Core}, are orders-of-magnitude smaller architectures finetuned on CIFs with poor reasoning and natural language skills~\cite{rubungo_llm4mat-bench_2024}. In addition, across several properties, \textbf{ALM Core} improves on previously published text-LLM property predictors by 5--100$\times$ in MAE. \textbf{ALM Core} is competitive with non-MLIP graph neural networks and stronger than previous language model property predictors on the Mat2Props~\cite{park_15_2024} benchmark as well (Appendix~\ref{tab:gen_vs_edit_understand}).

MatterChat~\cite{tang_multimodal_2026} is a recent method that also projects MLIP embeddings as soft tokens into LLMs for property prediction. When finetuned on $2$ epochs of the MatterChat training data (after training on roughly $20$ epochs' worth of other instruction tuning data, relative to MatterChat's reported training budget of $50$ epochs), \textbf{ALM Core} matches MatterChat on $5/9$ tasks and beats it on $2$ tasks (Fig.~\ref{fig:understanding}B). ALM underperformance on the other two tasks reflects the low LoRA rank chosen to preserve natural language skills and the limited underlying chemical expressivity of the atomistic encoder.

% can delete? check other papers. 
% \subsubsection{Attention scores are sensitive to chemistry and task}
% \label{sec:interpretability}\label{sec:appendix_interp} %????? should be in appendix, no?

The language model backbone of \textbf{ALM Core} learns attention scores that depend on the question \textit{and} chemical structure of the inputted material. For example, \textbf{ALM Core} attends to an inputted oxide structure's cations 1.5--2.5$\times$ more for a natural language output task compared to a property-prediction prompt. This asymmetry flips for intermetallic materials. Further, \textbf{Core} learns 
\begin{wraptable}[14]{r}{0.45\linewidth}
  \vspace{-\baselineskip}
  \centering
  \caption{\textbf{ALM retains zero-shot natural language capabilities and scientific knowledge despite multimodal finetuning.} The final, "Judge" column is an eval released with \textbf{ALM Bench}, measuring materials science knowledge retention as judged by GPT-4o (Appendix~\ref{app:stage2_eval}).}
  \label{tab:lang_retention_results}\label{tab:lang_retention}\label{tab:lang-retention}
  \scriptsize
  \setlength{\tabcolsep}{3pt}
  \begin{tabular}{lcccc}
  \toprule
  Model & MMLU & GSM8K & GPQA & Judge \\
  \midrule
  % \multicolumn{5}{l}{\emph{External crystal-LLM:}}\\
  CrystalReasoner~\cite{wu_crystalreasoner_2026} & $0.375$ & $0.020$ & $\mathbf{0.289}$ & $0.053$ \\
  \quad\emph{Qwen2.5-3B base} & \emph{0.550} & \emph{0.310} & \emph{0.221} & \emph{0.816} \\
  \midrule
  \rowcolor{almCoreBg}ALM Core & $\mathbf{0.595}$ & $\underline{0.775}$ & $0.247$ & $\mathbf{0.921}$ \\
  \rowcolor{almEditBg}ALM Edit & $\underline{0.485}$ & $\mathbf{0.780}$ & $0.228$ & $\mathbf{0.921}$ \\
  \quad \emph{Qwen3-8B base} & \emph{0.595} & \emph{0.705} & \emph{0.286} & \emph{0.895} \\
  \bottomrule
  \label{tab:knowledge-retention}
  \end{tabular}
  \end{wraptable}
  % \vspace{-1mm}
similar soft token activations among property prediction tasks and separately similar activations among free-form natural language tasks, like structure description and writing narratives about a material's properties and applications. This effect scales with input crystal size (Fig.~\ref{fig:understanding}E vs. F) and is exacerbated towards later transformer layers of the language model (Fig.~\ref{fig:understanding}G). 

% \subsubsection{Aligning structure does not displace natural language capabilities}
% \label{sec:no_forgetting}

A common failure mode of multimodal language models is catastrophically forgetting their post-trained natural language skills~\cite{liu_voxtral_2025}. Many systems avoid this by training small contrastive heads on frozen LMs, paying for it with reduced cross-modal control, or accepting language degradation as a cost of integration~\cite{liu_voxtral_2025}. \textbf{ALM Core} resists catastrophic forgetting and requires neither trade-off. \textbf{ALM Core} and \textbf{ALM Gen} improve multi-step reasoning and arithmetic multiple-choice question skills, as well as free-form materials science knowledge as evaluated by a frontier LLM judge, compared to their base language model (Table~\ref{tab:knowledge-retention}). This performance is driven by CAMEL~\cite{li2023camel} scientific question-answering and JARVIS~\cite{choudhary_joint_2020} materials science arXiv abstract buckets used for training \textbf{ALM Core}. In comparison, CrystalReasoner~\cite{wu_crystalreasoner_2026}, a language-based crystal generative model, regresses strongly on these tasks compared to its base model. Although it narrowly beats other models on GPQA~\cite{rein_gpqa_2023}, all models hover around the 25\% random baseline.

\subsection{State-of-the-art crystal structure prediction performance}
\label{sec:results_csp}

\begin{table*}[h]
\centering
\caption{\textbf{ALM Edit achieves SoTA performance at crystal structure prediction on MP-20 and MPTS-52.} Match rate MR (\%, $\uparrow$) and RMSE (\AA, $\downarrow$) to MP-20 and MPTS-52 test sets are scored at $K{=}1$ and best-of-$K{=}20$. Matcher, definitions, baselines, and conventions are available in Appendix~\ref{app:csp_metrics}. \textbf{Bold} and \underline{underlined} denote the best and second-best performance scores.}
\label{tab:body_csp_mp20}
\footnotesize
\setlength{\tabcolsep}{4pt}
\resizebox{\textwidth}{!}{%
\begin{tabular}{l cccc cccc}
\toprule
  & \multicolumn{4}{c}{MP-20} & \multicolumn{4}{c}{MPTS-52} \\
\cmidrule(lr){2-5}\cmidrule(lr){6-9}
Model & MR@1 (\%)$\uparrow$ & RMSE@1$\downarrow$ & MR@20 (\%)$\uparrow$ & RMSE@20$\downarrow$ & MR@1 (\%)$\uparrow$ & RMSE@1$\downarrow$ & MR@20 (\%)$\uparrow$ & RMSE@20$\downarrow$ \\
\midrule
CDVAE~\cite{xie_crystal_2022}        & 33.90 & 0.1045 & 66.95 & 0.1026 & 5.34 & 0.2106 & 20.79 & 0.2085 \\
DiffCSP~\cite{jiao_crystal_2023}     & 51.49 & 0.0631 & 77.93 & 0.0492 & 12.19 & 0.1786 & 34.02 & 0.1749 \\
FlowMM~\cite{miller_flowmm_2024}     & 61.39 & 0.0566 & --- & --- & 17.54 & 0.1726 & --- & --- \\
CrystaLLM-large~\cite{antunes_crystal_2024} & 58.70 & \underline{0.0408} & 73.97 & 0.0349 & 19.21 & \underline{0.1110} & 33.75 & \underline{0.1059} \\
CrystalFlow~\cite{luo_crystalflow_2025} & 62.02 & 0.0710 & \underline{78.34} & 0.0577 & 21.00 & 0.1613 & \underline{37.81} & 0.1584 \\
OMatG~\cite{hoellmer_omatg_2025}     & \underline{63.75} & 0.0720 & --- & --- & \underline{25.15} & 0.1931 & --- & --- \\
MCFlow-L~\cite{seong_mcflow_2026}    & \textbf{64.08} & 0.0561 & 76.08 & \underline{0.0383} & \textbf{27.16} & 0.1401 & 41.45 & 0.1296 \\
\midrule
\rowcolor{almEditBg}\textbf{ALM Edit}            & $45.6$ & $\mathbf{0.021}$ & $\mathbf{83.2}$ & $\mathbf{0.034}$ & $22.7$ & $\mathbf{0.022}$ & $\mathbf{45.7}$ & $\mathbf{0.038}$ \\
\rowcolor{almGenBg}\textbf{ALM Gen + T2C-FK}  & $22.3$ & $0.025$ & $41.0$ & $0.012$ & $6.0$ & $0.040$ & $10.0$ & $0.011$ \\
\bottomrule
\end{tabular}}
\label{tab:csp}
\end{table*}

Building on the performance and rich, aligned latent spaces of \textbf{ALM Core}, \textbf{ALM Edit} establishes a new state-of-the-art for unseen crystal structure prediction on MP-20, as well as MPTS-52, a significantly harder benchmark with structures over twice as large as MP-20 (Table~\ref{tab:csp}). Although many of the target polymorphs are relatively energetically stable compared to other geometries, the exact choice defined by each dataset is somewhat arbitrarily set~\cite{martirossyan_polymorph_2025}. Therefore, models are over-penalized for generating physically realistic unit-cell doublings, global rotations, or other stable polymorphs. However, \textbf{ALM Edit} not only achieves SoTA RMSE, outputting geometrically similar polymorphs to dataset targets, but also high best-of-$K=20$ match rates, indicating that it learns a valid distribution of polymorphs, one of which is likely to be the target polymorph set by the MP-20 and MPTS-52 evals. \textbf{ALM Edit} received the desired crystal's chemical composition \textit{and symmetry space group} during training, but only the chemical composition at inference, helping the model learn a relationship between crystal symmetry groups expressed as text tokens and generated symmetric polymorphs. Table~\ref{tab:csp} also reports \textbf{ALM Gen} with FK enabled, which trades crystal structure prediction performance for increased crystal stability and novelty, by design.

\subsection{Unlocking text-guided materials optimization and inverse design with ALM Bench}
\label{sec:results_generation}\label{sec:results_mode4}

% Specifications are paraphrased structural narratives drawn from GPT-Narratives parquets and LLM4Mat-Bench description fields. Each generated structure is geometry-validity-screened, MatterSim-relaxed (\texttt{mattersim-v1.0.0-1M}, fmax $0.05$), and scored against MP-2020 with standard \texttt{MaterialsProject2020Compatibility} corrections. M.S.U.N.\ is the fraction of generations simultaneously stable ($E_{\text{hull}} \le 0.1$ eV/atom), unique within the sampled batch, and novel against the reference.

% Most useful specification is looser than a formula --- "make this denser," "dope it with magnesium," "give me a thermoelectric" --- and no existing benchmark scores a model's ability to carry such instructions end-to-end into atomic coordinates. We release \textbf{ALM Bench} to fill that gap: held-out natural-language instructions spanning directional property editing, polymorph generation, doping and substitution, strain, application-conditioned generation, and text-to-structure recovery, each scored by \emph{realness-gated} metrics that a relabeling or lattice-rescaling shortcut cannot game, with the identical prompt set handed to every model. We benchmark \textbf{ALM Edit} head-to-head against frontier text-LLMs (gpt-4o, gpt-4.1, gpt-5.2) prompted to read and write CIFs directly (Tables~\ref{tab:body_alm_vs_frontier},~\ref{tab:body_almbench_gen}).

\textbf{ALM Edit} unlocks the ability to \textit{edit and generate materials as instructed by natural language}. We introduce \textbf{ALM Bench} to evaluate this capability, testing \textbf{Edit} and several frontier LLMs (with thinking mode enabled) on producing valid polymorphs of the inputted material with properties adjusted in a particular direction, e.g., "increase the formation energy of this crystal" (corresponding to ``$E_f\uparrow$'' below). \textbf{ALM Bench} also evaluates the model's one-shot ability to generate polymorphs of a given crystal with $E_\text{hull} < 0$ (``Polymorph''), as well as doping tasks, where models are prompted to dope a given crystal with a new element. The structural and compositional match of the generated and true doped crystal (``Doping''), as well as the strain of each (``Strain''), score success at this task (Appendix~\ref{app:almbench}). 

\begin{table*}[ht]
\centering
\caption{\textbf{ALM Bench evaluates models on atomistic editing tasks as guided by language.}
  Directional editing per property ($E_f$, $\rho$, $V$) and direction ($\uparrow$/$\downarrow$) are indicated ($N=7\times1000$). OpenAI models were prompted to generate CIFs.}
\label{tab:body_alm_vs_frontier}
\small
\setlength{\tabcolsep}{3pt}
\resizebox{\textwidth}{!}{%
\begin{tabular}{lccccccccc}
\toprule
Model & $E_f\uparrow$ & $E_f\downarrow$ & $\rho\uparrow$ & $\rho\downarrow$ & $V\uparrow$ & $V\downarrow$ & Polymorph & Doping & Strain \\
\midrule  
\rowcolor{almEditBg}\textbf{ALM Edit} & $\mathbf{0.613}\ci{0.062}$ & $\mathbf{0.624}\ci{0.021}$ & $\mathbf{0.353}\ci{0.067}$ & $\mathbf{0.367}\ci{0.032}$ & $\mathbf{0.451}\ci{0.059}$ & $\mathbf{0.355}\ci{0.033}$ & $\mathbf{0.224}\ci{0.022}$ & $\mathbf{0.879}\ci{0.012}$ & $\mathbf{0.151}\ci{0.028}$ \\
GPT-4o & $\underline{0.505}\ci{0.043}$ & $0.469\ci{0.057}$ & $0.024\ci{0.022}$ & $0.127\ci{0.013}$ & $0.081\ci{0.030}$ & $0.018\ci{0.015}$ & $0.040\ci{0.007}$ & $\underline{0.007}\ci{0.002}$ & $\underline{0.000}\ci{0.000}$ \\
GPT-4.1 & $0.465\ci{0.032}$ & $\underline{0.496}\ci{0.025}$ & $0.007\ci{0.009}$ & $0.239\ci{0.034}$ & $\underline{0.276}\ci{0.035}$ & $\underline{0.040}\ci{0.016}$ & $0.083\ci{0.013}$ & $0.003\ci{0.004}$ & $0.000\ci{0.000}$ \\
GPT-5.2 & $0.437\ci{0.073}$ & $0.414\ci{0.043}$ & $\underline{0.058}\ci{0.026}$ & $\underline{0.244}\ci{0.019}$ & $0.006\ci{0.005}$ & $0.032\ci{0.015}$ & $\underline{0.118}\ci{0.023}$ & $0.002\ci{0.002}$ & $0.000\ci{0.000}$ \\
\bottomrule
\end{tabular}}
\end{table*}

\begin{table*}[ht]
\centering
\caption{\textbf{ALM Bench also evaluates models on crystal generation tasks.}
  Consistency of generated materials ($N=7\times1000$) to requested application area (LLM Judge-assessed), materials description, and adversarial composition input (Appendix~\ref{app:almbench}).}
\label{tab:body_almbench_gen}
\footnotesize
\setlength{\tabcolsep}{3pt}
\begin{tabular}{lccccc}
\toprule
Model & Application & Describe (Comp.) & Describe (Struct.) & OOD (Comp.) & OOD (Struct.) \\
\midrule
\rowcolor{almEditBg}\textbf{ALM Edit} & $\mathbf{0.423}\ci{0.020}$ & $\mathbf{0.730}\ci{0.031}$ & $\mathbf{0.412}\ci{0.041}$ & $\mathbf{0.474}\ci{0.016}$ & $\mathbf{0.231}\ci{0.026}$ \\
GPT-4o & $0.131\ci{0.042}$ & $0.279\ci{0.019}$ & $0.121\ci{0.024}$ & $0.130\ci{0.020}$ & $0.025\ci{0.012}$ \\
GPT-4.1 & $0.224\ci{0.032}$ & $0.254\ci{0.019}$ & $0.090\ci{0.018}$ & $0.168\ci{0.018}$ & $0.035\ci{0.007}$ \\
GPT-5.2 & $\underline{0.252}\ci{0.051}$ & $\underline{0.356}\ci{0.014}$ & $\underline{0.162}\ci{0.013}$ & $\underline{0.263}\ci{0.019}$ & $\underline{0.075}\ci{0.010}$ \\
\bottomrule
\end{tabular}
\end{table*}

As shown in Table~\ref{tab:body_alm_vs_frontier}, \textbf{ALM Edit} produces valid edits to crystal geometry and structure, as each \textbf{ALM Bench} metric scores structurally invalid generations, trivial lattice rescalings, and unphysical atom relabelings as failures. When prompted to complete the same task using CIFs, frontier LLMs trail \textbf{ALM Edit} on every metric, also often failing to produce valid, nontrivial crystals.

\textbf{ALM Bench} also evaluates models on text-conditioned generation alone, including asking for a crystal that belongs to an ``Application'' area (e.g., ``generate a perovskite'') and difficult crystal structure prediction prompts, ranging from long narratives about properties and structure (``Describe'') to adversarially designed, terse prompts (``OOD'', e.g. "generate MgO structure at 3.58 g/cm$^3$"). Again, \textbf{ALM Edit} leads frontier LLM baselines by a sizeable margin (Table~\ref{tab:body_almbench_gen}).

\subsection{Competitive \textit{de novo} generation for stable crystal generation}
\label{sec:results_dng}%
To enable large-throughput screening campaigns of \textit{de novo} generated crystals, we deliberately relax \textbf{ALM Edit}'s strong conditioning into \textbf{ALM Gen} (\S\ref{sec:methods_dng}), in which prompts bias the distribution of sampled structures without dictating each sample. These prompts are drawn uniformly from the eval set of prompts for \textbf{Edit}. \textbf{ALM Gen} achieves state-of-the-art performance for simultaneously stable, unique, and novel (SUN) structures, where stability is measured by $E_{\text{hull}} < 0.016$ on the MP-20 hull~\cite{Jain2013}. Crucially, turning guidance on to $g=0.5$ and steering $\mathcal D$ using natural language \textit{improves} generation quality over the $g=0$ MatterGen base (Table~\ref{tab:dng_strict_mp20}), highlighting that learned language model priors improve upon unimodal, atomistic generation. On the harder, broader-chemistry LeMat-GenBench~\cite{betala_lemat-genbench_2026} protocol (Table~\ref{tab:body_dng_mpts52}), it tops the field on metastable ($E_{\text{hull}} < 0.1$) yield while being second to the \textit{de novo}-specialist flow models at strict SUN, where strict stability is defined as $E_{\text{hull}} < 0$. Metastable yield on MP-20 is reported in full in Appendix~\ref{app:metric_dng}.

\begin{table*}[ht]
\centering
\caption{\textbf{\textit{De novo} generation against the MP-20 hull}~\cite{wu_crystalreasoner_2026}, in which stability $S$ is defined by $E_\mathrm{hull}\le0.016$~eV/atom and structures are pre-relaxed. Evaluated on $N=10\times1000$ samples (Appendix~\ref{app:metric_dng}).}
\label{tab:dng_strict_mp20}
\footnotesize
\setlength{\tabcolsep}{3pt}
\begin{tabular}{lrrrrr}
\toprule
Method & $E_\mathrm{hull}$ (eV)\,$\downarrow$ & $U$ (\%)\,$\uparrow$ & $V_\mathrm{struct}$ (\%)\,$\uparrow$ & $V_\mathrm{chem}$ (\%)\,$\uparrow$ & \textbf{SUN} (\%)\,$\uparrow$ \\
\midrule
CrystalTextLLM~\cite{gruver_fine-tuned_2024}  & $0.61\ci{0.003}$ & $47.40\ci{0.30}$ & $90.01\ci{0.21}$ & $91.59\ci{0.05}$ & $0.38\ci{0.05}$ \\
PLAID++ Wyckoff~\cite{plaid_xu_2025}                                        & $0.57\ci{0.003}$ & $40.70\ci{0.30}$ & $89.06\ci{0.21}$ & $91.59\ci{0.04}$ & $0.50\ci{0.05}$ \\
CrysReas-Base (SFT only)                               & $0.58\ci{0.004}$ & $35.25\ci{0.31}$ & $84.03\ci{0.23}$ & $90.36\ci{0.10}$ & $0.57\ci{0.05}$ \\
CrysReas-Thinking (SFT+CoT)                            & $0.52\ci{0.003}$ & $38.64\ci{0.29}$ & $91.29\ci{0.19}$ & $\underline{91.72}\ci{0.04}$ & $0.59\ci{0.06}$ \\
CrysReas-RL (SFT+GRPO)                                 & $0.53\ci{0.003}$ & $82.49\ci{0.19}$ & $89.85\ci{0.20}$ & $91.10\ci{0.07}$ & $1.23\ci{0.07}$ \\
CrysReas~\cite{wu_crystalreasoner_2026}                                   & $0.45\ci{0.003}$ & $87.23\ci{0.14}$ & $\underline{94.92}\ci{0.15}$ & $\mathbf{91.78}\ci{0.03}$ & $1.70\ci{0.08}$ \\
\midrule
MatterGen (Base) & $\mathbf{0.079}\ci{0.002}$ & $\underline{93.50}\ci{0.90}$ & $\mathbf{100.00}\ci{0.00}$ & $86.50\ci{2.90}$ & $\underline{5.53}\ci{1.17}$ \\
\rowcolor{almGenBg}\textbf{ALM Gen}                & $\underline{0.085}\ci{0.003}$ & $\mathbf{98.90}\ci{0.70}$ & $\mathbf{100.00}\ci{0.00}$ & $83.20\ci{2.90}$ & $\mathbf{7.80}\ci{0.44}$ \\
\rowcolor{almGenBg}ALM Gen $+$ FK-stoich & $0.086\ci{0.005}$ & $73.80\ci{2.20}$ & $\mathbf{100.00}\ci{0.00}$ & $84.50\ci{1.00}$ & $5.21\ci{1.18}$ \\
\bottomrule
\end{tabular}
\end{table*}

% [metastable-$MS$ DNG table moved to appendix Metrics>DNG (tab:dng_meta_mp20) ---
%  metastable-MSUN is the one DNG view where we trail Crys-JEPA-full, so it lives there.]

\begin{table*}[ht]
\centering
\caption{\textbf{\textit{De novo} generation on LeMat-GenBench~\cite{betala_lemat-genbench_2026}}~\cite{seong_mcflow_2026} ($N=2500$), in which strict stability is defined by $\bar E_\mathrm{hull}<0$~eV/atom and structures are pre-relaxed. Validity is measured by charge neutrality, physical plausibility, and minimum distance checks, and metastability by $E_\mathrm{hull}<0.1$. $E_f$, $E_{\text{hull}}$, and RMSD are scored by 3 MLIPs (further details and conventions in Appendix~\ref{app:metric_dng}).}
\label{tab:body_dng_mpts52}
\scriptsize
\setlength{\tabcolsep}{3pt}
\resizebox{\textwidth}{!}{%
\begin{tabular}{lrrrrrrrrrr}
\toprule
 & & & & \multicolumn{3}{c}{Energy-based} & \multicolumn{2}{c}{Strict Stability} & \multicolumn{2}{c}{Metastab.} \\
\cmidrule(lr){5-7}\cmidrule(lr){8-9}\cmidrule(lr){10-11}
Model & Valid$\uparrow$ & Unique$\uparrow$ & Novel$\uparrow$ & $E_f\downarrow$ & $\bar E_\mathrm{hull}\downarrow$ & RMSD$\downarrow$ & Stable$\uparrow$ & SUN$\uparrow$ & Meta$\uparrow$ & MSUN$\uparrow$ \\
\midrule
MatterGen~\cite{zeni_generative_2025} & 95.7 & 95.1 & \textbf{70.5} & $-0.70\ci{0.79}$ & $0.18\ci{0.18}$ & $0.39\ci{0.50}$ & 2.0 & 0.2 & 33.4 & 15.0 \\
PLaID++~\cite{plaid_xu_2025}       & 96.0 & 77.8 & 24.2 & $-0.50\ci{0.44}$ & $0.09\ci{0.16}$ & $0.13\ci{0.29}$ & 12.4 & \textbf{1.0} & 60.7 & 7.6 \\
WyFormer~\cite{kazeev_wyformer_2025}      & 93.4 & 93.0 & \underline{66.4} & $-0.43\ci{0.95}$ & $0.50\ci{0.51}$ & $0.81\ci{0.98}$ & 0.5 & 0.1 & 15.7 & 1.9 \\
WyFormer-DFT  & 95.2 & 95.0 & 66.4 & $-0.67\ci{0.91}$ & $0.27\ci{0.36}$ & $0.42\ci{0.60}$ & 3.7 & 0.4 & 24.8 & 7.8 \\
MCFlow-S~\cite{seong_mcflow_2026}      & 97.2 & \textbf{96.3} & 52.2 & $-0.85\ci{0.84}$ & $0.10\ci{0.12}$ & $0.16\ci{0.27}$ & 11.7 & 0.7 & 49.5 & \underline{18.9} \\
MCFlow-B~\cite{seong_mcflow_2026}      & \underline{97.7} & \underline{95.5} & 25.4 & $\underline{-0.91}\ci{0.85}$ & $\underline{0.05}\ci{0.10}$ & $\underline{0.08}\ci{0.18}$ & \underline{17.6} & 0.7 & \underline{64.3} & 11.9 \\
MCFlow-L~\cite{seong_mcflow_2026}      & \textbf{98.6} & 95.2 & 18.6 & $\mathbf{-0.93}\ci{0.87}$ & $\mathbf{0.04}\ci{0.08}$ & $\mathbf{0.06}\ci{0.15}$ & \textbf{18.8} & 0.5 & \textbf{68.3} & 9.3 \\
\midrule
% \rowcolor{almGenBg}\textbf{ALM Gen} & 92.2$\ci{1.2}$ & 91.3$\ci{0.6}$ & 61.5$\ci{2.0}$ & $-0.44\ci{0.06}$ & $0.091\ci{0.003}$ & $0.199\ci{0.009}$ & 3.6$\ci{0.3}$ & \underline{0.8}$\ci{0.3}$ & 58.7$\ci{2.1}$ & \textbf{35.2}$\ci{2.7}$ \\
\rowcolor{almGenBg}\textbf{ALM Gen} & 92.2 & 91.3 & 61.5 & $-0.44\ci{0.06}$ & $0.09\ci{0.00}$ & $0.20\ci{0.01}$ & 3.6 & \underline{0.8} & 58.7 & \textbf{35.2} \\
\bottomrule
\end{tabular}}
\end{table*}

\section{Discussion}
\label{sec:discussion}

Atomistic Language Models are, to our knowledge, the first natively multimodal models that support atomistic understanding and text-instructed, materials-conditioned generation through continuous latent space bridges. ALMs directly address the three failure modes named in \S\ref{sec:intro}. The first, that components communicate only through lossy surface forms and discard continuous geometric information, is contradicted by breaking the ``GNN--LLM wall'' on LLM4Mat-Bench (\S\ref{sec:llm4mat}), where keeping the representation continuous recovers GNN-level property prediction through a language interface. The second, that independently trained components never share a latent space, is combated by continuous latent projectors, retained natural language abilities, and the late-layer task-mode classifier (\S\ref{sec:results_understanding}) in ALMs, which together show rich joint latent spaces. The third, that generation can be steered only through what a prompt can express and not through the latent space the model reasons in, is solved by the language-to-atomistic bridge architecture for steering materials generation (\S\ref{sec:results_generation}). Any single result might be matched by a stitched pipeline; the paradigm claim is that Atomistic Language Modeling delivers all three at once, reaching state-of-the-art performance on materials discovery while unlocking language-instructed inverse design. \textbf{ALM Bench} is released to evaluate this ability and invite new architectures.

\textbf{ALM Edit} demonstrates that language models can effectively guide materials generation via last-layer hidden state token embeddings. However, the architecture choices and training recipe that maximize this latent steering while keeping computational complexity low remain open. Although ablations were completed over several bridges (Appendix~\ref{app:diffusion_adapter}), only the proposed bridge could amplify task and structural signal enough to guide generation. This signal is sensitive even to the order in which the atomistic tokens are teacher-forced (Fig.~\ref{fig:steering_gsweep}). The bridge-architecture difference between \textbf{ALM Edit} and \textbf{Gen}, and the need for two separate diffusion decoders, would be mitigated by a different decoder model (Appendix~\ref{app:cfg_tension}). 

\textbf{T2C-FK}, an inference-time steering method, bridges the gap between the stability of \textbf{ALM Gen}'s outputs and the instruction-following capabilities of \textbf{ALM Edit}. As a decoder-agnostic layer, it can generalize far beyond stoichiometry for any constraint expressible as a per-step reward, including charge balance and unit cell symmetry. In particular, adapting \textbf{T2C-FK} to \textbf{ALM Edit}'s formation-energy directional editing tasks, using MatterSim energy evaluations as an inference-time reward, \textit{improved} $E_f\uparrow$ \textit{performance}, generating polymorphs with increased energy successfully $72.5$\% of the time, and lower energy polymorphs with $71.3$\% success. 

\begin{figure}[h!]
    \centering
    \includegraphics[trim={0cm 5cm 0cm 0cm},clip,width=1\linewidth]{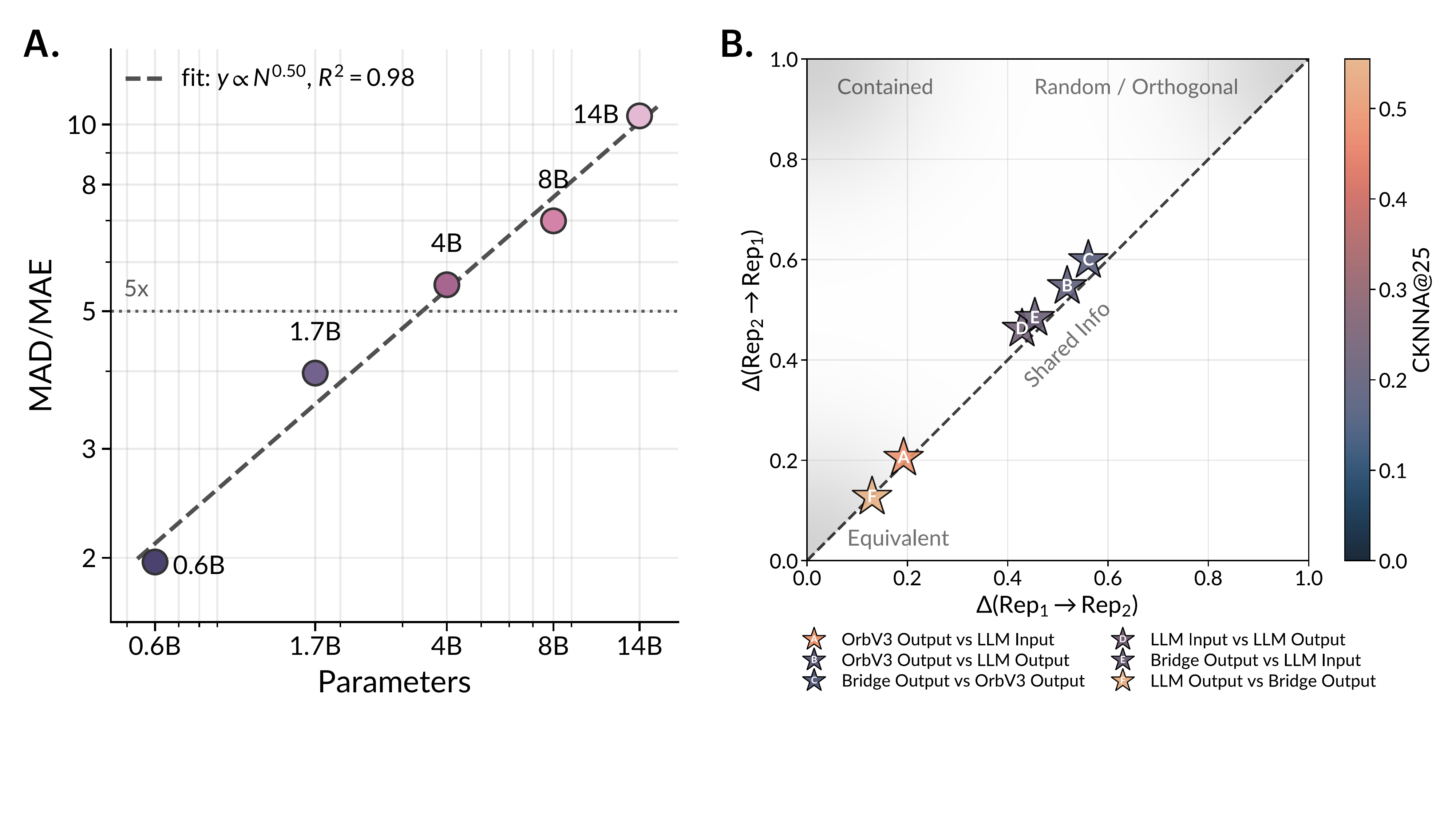}
    \vspace{-3mm}
    \caption{\textbf{Strong scaling laws emerge under fixed training and evaluation for several property prediction tasks.}
    A. For increasing Qwen3 model size, property prediction performance on several tasks, including JARVIS-QETB potential energy per atom above, improves monotonically in MAD/MAE on LLM4Mat-Bench. B. Representational analysis of embeddings extracted from each continuous latent space throughout \textbf{ALM Edit} for 2,000 prompts from \textbf{ALM Bench}. Information imbalance agrees with CKNNA, a local embedding neighborhood alignment metric~\cite{huh_platonic_2024}.}
    \label{fig:discussion_scaling_law}
\end{figure}

A crucial driver of multi-step reasoning and language model performance is the emergence of scaling laws with larger models and datasets. \textit{Accordingly, the property predictive performance of ALMs scales with language model size}. In particular, Fig.~\ref{fig:discussion_scaling_law}A shows how property prediction performance increases with parameter count (shown for several other properties in Fig.~\ref{fig:app_model_size_scaling}). This is a promising sign that further ALM capabilities may emerge with data and model scaling~\cite{wei2022emergentabilitieslargelanguage}. 

%%EDIT
By learning continuous latent spaces across all modalities within a single architecture, Atomistic Language Modeling makes representational alignment quantifiable \textit{across modalities} without resorting to an artificial contrastive objective. Information imbalance~\cite{glielmo_ranking_2022}, a global representational similarity metric over sets of embeddings, quantifies the difference in information content between representations of each modality. For the same prompts passed through \textbf{ALM Edit}, the atomistic-to-language MLP adapter and the language-to-atomistic bridge do not meaningfully change the information content of the latent spaces they translate between (Fig.~\ref{fig:discussion_scaling_law}B). The metric further reveals that the MLIP's predictive embeddings, the language model's atomistic-token embeddings, and the diffusion decoder's steering vectors share substantially more information with one another than independently trained language and atomistic models do~\cite{edamadaka_universally_2025}. However, no pair collapses into the \textit{informationally equivalent} corner: each latent space stays distinct enough to carry information the others lack. Atomistic Language Modeling thus overcomes the representational gap between language and 3D structure, learning latent spaces that grow increasingly aligned while remaining informationally distinct enough to meaningfully contribute. 

Several limitations frame \textbf{ALM Core, Edit,} and \textbf{Gen} as starting points for the Atomistic Language Modeling paradigm. ALMs learn in continuous latent spaces, but many multimodal models instead discretize latent spaces into a learned codebook, like in VQ-VAE~\cite{oord_neural_2018} or JANUS~\cite{chen_janus-pro_2025}. For simple inorganic crystals, quantizing a rich atomistic latent space and training the language model to emit corresponding tokens for decoding back into valid materials is an alternate design. The ALM is also trained only on inorganic, crystalline materials. However, it in principle extends to any system with an atomistic representation by swapping in a more foundational encoder, such as UMA or PET-MAD (Appendix~\ref{app:encoder_ablation}), or a less locally biased encoder~\cite{kreiman2025transformersdiscovermolecularstructure}. ALMs' continuous latent interface may be especially valuable for large, amorphous, or defect-ridden systems with large numbers of atoms, which yield more information-dense descriptions in language compared to all-atom representations. Lastly, thinking was disabled for all ALM configurations due to a lack of chain-of-thought reasoning data. Constructing datasets from tool-assisted frontier-model reasoning traces and RL-finetuning atomistic language models for improve performance is a promising future direction~\cite{wu_crystalreasoner_2026}.  % MLIPs as local.
\vspace{-2mm}
\vspace{-5mm}

\section{Conclusion}
\label{sec:conclusion}

Atomistic structure and natural language carry such different representational biases that prior work models them separately and stitches them together.
Atomistic Language Modeling instead unifies a pretrained atomistic encoder, language model, and denoising diffusion decoder into shared latent spaces through continuous projectors.
\textbf{ALM Core} predicts physical properties of crystals with the performance of atomistic graph neural networks, yet without losing its natural language abilities. 
\textbf{ALM Edit} unlocks the ability to optimize given crystals according to natural language prompts using a novel language-atomistic bridge architecture, also setting a new SoTA for crystal structure prediction Match@K=20 and RMSE on MP-20 and MPTS-52. 
\textbf{ALM Gen} achieves \textit{de novo} generation, producing SUN crystals at higher rates than prior atomistic and language-based models, with \textbf{T2C-FK} steering it toward stoichiometric targets at inference time. 
Atomistic Language Modeling is a promising paradigm that steers materials prediction, generation, and optimization with natural language, inheriting the strong scaling laws of underlying language model backbones.

%
%
% RELATED WORKS
%
%

\section{Related Work} % need to review. 

Unimodal atomistic models, multimodal materials models built by system-level composition, and inference-time control of diffusion denoising serve as the foundation for Atomistic Language Modeling. 

\subsection{Molecules}
\label{sec:rw_molecules}
Although ALMs are trained on inorganic crystals, multimodal models of molecules supply some working recipes, but lean on near-lossless string encodings (SMILES) that are faithful only to equilibrium conformations and for which crystalline materials have few analogs. SMILES~\cite{weininger_smiles_1988} and SELFIES~\cite{krenn_self-referencing_2020} encode molecular graphs as strings with essentially no information loss for an equilibrium geometry, making molecule-to-text conversion nearly trivial and yielding a long line of working systems. MoleculeSTM~\cite{liu_multi-modal_2024} contrastively aligns descriptions and atomistic structures for retrieval; LLM-Fusion~\cite{boyar_llm-fusion_2025} fuses SMILES, SELFIES, text, and learned embeddings for property prediction; 3D-MoLM~\cite{li_towards_2024} fine-tunes a language model on 3D molecular encodings, showing that atomistic embeddings meaningfully enhance a language model, an approach similar to our atomistic encoder-soft token architecture. SMILES-based translators~\cite{edwards_translation_2022, kim_merged_2021, qian_can_2023} move between language and molecular structure, and MMFRL~\cite{zhou_multimodal_2025} aligns SMILES with property data through spectra. Crystalline materials have no identical SMILES analog (e.g., CIF files preserve atomic coordinates but are not natural language). 

\subsection{Single-modality materials models}
\label{sec:rw_single_modality}

On the language side, LLM-Prop~\cite{niyongabo_rubungo_llm-prop_2025} predicts properties from Robocrystallographer text but discards numerical structural information; CrystalLLM~\cite{antunes_crystal_2024} trains autoregressively on CIF files for \textit{de novo} generation; MatSciBERT~\cite{gupta_matscibert_2021} captures domain language without structural input; and Crystal-Text-LLM~\cite{gruver_fine-tuned_2024} fine-tunes a pretrained LM on CIF-style generation for crystal structure prediction, reportedly with greater diversity and stability than contemporary specialized atomistic models, evidence that language model priors carry useful inductive biases for crystal generation. On the structure side, atomistic foundation models (machine learning interatomic potentials)~\cite{wood_uma_2025, batatia_foundation_2024} can be finetuned to predict properties from periodic graphs at state-of-the-art accuracy, and diffusion-based generators~\cite{xie_crystal_2022, jiao_crystal_2023, zeni_generative_2025} sample structures from learned distributions over crystal lattices. Neither side alone couples text-conditioned structure generation with property reasoning; ALM combines them in one backbone, breaking the ``GNN--LLM'' accuracy wall on property prediction while generating crystal structures competitively with natural language steering.

\subsection{Multimodal materials via system-level composition}
\label{sec:rw_system_composition}

Multimodal materials models to date compose pretrained components at the system level. MultiMat~\cite{moro_multimodal_2025} aligns atomistic embeddings, numerical property data, and Robocrystallographer text with a CLIP-style contrastive loss, but supports only property prediction. CLaSP~\cite{suzuki_bridging_2025} aligns language with atomistic structures from paper titles and abstracts and tests only retrieval. CLICS~\cite{ozawa_graph-text_2024} contrastively learns over atomistic embeddings and Robocrystallographer text but cannot consume free-form language. MatterChat \cite{tang_multimodal_2026} and L$^2$M$^3$OF \cite{cui_l2m3of_2025} consume free-form language and atomistic structure, but cannot generate atomistic structures. Against the failure modes of \S\ref{sec:intro}, none closes the full text-in, structure-out, structure-in, text-out loop; none places generation under cross-modal latent control to enable text-in, structure-in, structure-out generation; and their contrastive embedding spaces are aligned for retrieval, not steerable sampling. ALMs close that loop in a single model. In vision and language, analogous steps toward integrated multimodal understanding (LLaVA~\cite{liu_visual_2023}) and generation (Janus~\cite{chen_janus-pro_2025}) exist, yet do not transfer to atomistic understanding or scientific reasoning~\cite{rubungo_llm4mat-bench_2024}, nor to crystal generation. Their 2D image encoders carry no periodicity or atom-permutation symmetry, and the generator in Janus~\cite{chen_janus-pro_2025} discretizes outputs into a learned codebook that the continuous geometry and exact stoichiometry of crystals resist. LLaVA projects features for understanding only, while Janus generates within its own token space rather than bridging a language model to an external, physics-aware decoder, backpropagating the generative loss into the language model, or emitting a structured compositional target.

\subsection{Inference-time control of diffusion}
\label{sec:rw_diffusion_control}

Conditional diffusion typically injects guidance during sampling. Classifier guidance~\cite{dhariwal_diffusion_2021} and classifier-free guidance~\cite{ho_classifier-free_2022} steer the denoising trajectory toward conditioning information, but require differentiable score signals and do not support discrete compositional constraints. Particle-based Feynman--Kac methods~\cite{singhal_general_2025} reweight partial trajectories by a reward, softly enforcing arbitrary objectives without touching the diffusion model. None exploits a shared latent space between a language model and the diffusion decoder; T2C-FK does, scoring trajectories with a per-atom Hungarian assignment between the LM-conditioned predicted atomic distribution and the target stoichiometric multiset to enforce natural-language compositional targets at inference time.

% \subsection{Positioning}
% \label{sec:rw_positioning}

% ALM is, to our knowledge, the first natively multimodal model for materials that integrates a pretrained atomistic encoder, language model, and denoising diffusion decoder into a single shared latent space, supporting structure understanding, text-conditioned generation, and self-explanation in one autoregressive model. It departs from system-level composition in three ways that map directly to the failure modes of \S\ref{sec:intro}: components share continuous representations rather than communicating through surface forms, gradient signals flow across modality boundaries during training, and conditional generation is steered through continuous tokens within the model's own latent space. These choices pay off in the results: state-of-the-art crystal structure prediction and language-instructed materials optimization, leading property prediction, and competitive \textit{de novo} generation, all from one backbone and without surrendering language ability. T2C-FK is, to our knowledge, the first Feynman--Kac sampler that exploits a shared latent space between a language model and a diffusion decoder to enforce compositional targets specified in natural language.

\section*{Acknowledgements}
We would like to thank Ben Miller, Paul Liang, and Laura Ruis for their essential guidance on developing Atomistic Language Models and the framing of our work. We acknowledge the MIT Office of Research Computing and Data and Tata for providing high performance computing resources that have contributed to the research results reported within this paper. We are also grateful to Lyra Labs for providing compute for our work. We would also like to acknowledge \cite{bolya_perception_encoder_2025} for inspiring our overview figure.

\bibliographystyle{unsrt}
\bibliography{references}

\clearpage
\section*{Appendix}
\vspace{-3mm}
\medskip
% tocdepth=3 set globally BEFORE the appendix body so subsubsections are recorded
% into the partial ToC as each is typeset (titletoc filters at record time), then
% printed down to subsubsection.
\setcounter{tocdepth}{3}
\startcontents[appendix]
\printcontents[appendix]{}{1}{}

\clearpage \appendix

\section{Architecture design choices and ablations}
\label{app:architecture}

Atomistic Language Modeling (ALM) is designed to be the first paradigm that covers every direction of the materials structure--property--text map. Table~\ref{tab:capability_matrix} makes this concrete, contrasting ALM against the three model families that currently dominate materials science. Machine-learned interatomic potentials (MLIPs) regress properties from a fixed structure; atomistic generative models sample structures from a given composition, property, or an empty prompt; and large language models that serialize crystals as strings can read and write structures as text but carry no learned 3D geometric prior. Atomistic Language Modeling is the only paradigm that supports all seven.

\begin{table}[t]
\centering
\caption{\textbf{Positioning.} Capability coverage of materials science model families. Columns are seven task directions over the structure, property, and natural-language modalities; a \checkmark{} marks a direction each family supports natively with nontrivial performance. }
\label{tab:capability_matrix}
\footnotesize
\setlength{\tabcolsep}{4pt}
\begin{tabularx}{\linewidth}{@{}>{\raggedright\arraybackslash}X lllllll@{}}
\toprule
& Property & Text Desc. & Structure & \textit{De novo} & Structure+Text & Structure & Text \\
Model family & $\rightarrow$ Structure & $\rightarrow$ Structure & $\rightarrow$ Property & ($\varnothing\rightarrow$ Structure) & $\rightarrow$ Structure & $\rightarrow$ Text & $\rightarrow$ Text \\
\midrule
MLIPs~\cite{rhodes_orb-v3_2025,wood_uma_2025,batatia_foundation_2024}
  & $\times$ & $\times$ & \checkmark & $\times$ & $\times$ & $\times$ & $\times$ \\
Atomistic generative models~\cite{zeni_generative_2025,jiao_crystal_2023}
  & \checkmark & $\times$ & $\times$ & \checkmark & $\times$ & $\times$ & $\times$ \\
LLMs emitting string crystal reps.~\cite{antunes_crystal_2024,gruver_fine-tuned_2024}
  & $\times$ & \checkmark & \checkmark & \checkmark & $\times$ & \checkmark & $\times$ \\
\textbf{ALM (this work)}
  & \checkmark & \checkmark & \checkmark & \checkmark & \checkmark & \checkmark & \checkmark \\
\bottomrule
\end{tabularx}
\end{table}

\paragraph{Two conditional roles from one architecture.} The full coverage in Table~\ref{tab:capability_matrix} is realized by two checkpoints that instantiate the same language model-to-diffusion bridge on different decoder backbones. \textbf{ALM Edit} is the model with the full language-to-atomistic bridge architecture and composition-observing diffusion model, accomplishing both $\mathbf{T}\rightarrow\mathbf{S}$ and $(\mathbf{S},\mathbf{T})\rightarrow\mathbf{S}$, while \textbf{ALM Gen} covers the quasi-unconditional, de-novo generation direction ($\varnothing,\mathbf{T}\rightarrow\mathbf{S}$), with a diffusion model that also denoises over element types. The structure-to-property and structure-to-text directions are served by the shared atomistic encoder, projector, and LLM forming \textbf{ALM Core}. The remainder of this section documents the architecture and the design choices that make this coverage possible: the atomistic encoder (\S\ref{app:encoder_ablation}), the input-side adapter (\S\ref{app:codebook_comparison}), the language model adaptation (\S\ref{app:lora_config}), the generator and language-to-diffusion model bridge (\S\ref{app:generator_formal}), and the test-time T2C-FK steering method (\S\ref{app:fk_details}).

\begin{table}[ht]
\centering
\caption{\textbf{Training hyperparameters for the three released models.} Encoder $\mathcal{E}$ (OrbV3) is frozen throughout; ALM Edit/ALM Gen initialize the LM from the ALM Core understanding checkpoint. Data-mixture buckets and sampler are in Appendix~\ref{app:training} (ALM Core: Table~\ref{tab:s2-buckets}; generation fine-tuning: Section~\ref{app:training_stage3}). ``---'' marks an inapplicable entry.}
\label{tab:training_hparams}
\footnotesize
\setlength{\tabcolsep}{5pt}
\resizebox{\linewidth}{!}{%
\begin{tabular}{l l l l}
\toprule
& \textbf{ALM Core} & \textbf{ALM Edit} & \textbf{ALM Gen}  \\
\midrule
Objective              & $\mathcal{L}_{\mathrm{LM}}$ (Eq.~\ref{eq:loss_lm}) & $\mathcal{L}_{\mathrm{CSP}}$ (Eq.~\ref{eq:loss_csp}) & $\mathcal{L}_{\mathrm{DNG}}$ (Eq.~\ref{eq:loss_dng}) \\
LM adaptation          & LoRA $r{=}128$, $\alpha{=}256$        & full fine-tune        & LoRA $r{=}8$, $\alpha{=}16$ \\
Diffusion backbone     & ---                   & CSP-mode MatterGen & DNG-mode MatterGen-Base \\
Bridge                 & $P_{\text{in}}$ (2-layer MLP) & Q-Former ($M{=}16$) $+$ IP-Adapter & per-token MLP $+$ IP-Adapter \\
LM learning rate       & $2\mathrm{e}{-}4$ (LoRA) & $5\mathrm{e}{-}7$ & $2\mathrm{e}{-}4$ (LoRA) \\
Projector/bridge lr    & $2\mathrm{e}{-}5$ ($P_{\text{in}}$) & $3\mathrm{e}{-}4$ & $3\mathrm{e}{-}4$ \\
Optimizer              & AdamW                 & AdamW                 & AdamW \\
Batch / GPU            & $4$                   & $2$                   & $4$ \\
Max tokens             & $2048$                & $1536$                & $1536$ \\
Optimizer steps        & $12{,}000$            & $30{,}000$            & $10{,}000$ \\
Grad clip              & $1.0$                 & $1.0$                 & off \\
CFG dropout $p_{\mathrm{drop}}$ & ---          & $0.2$                 & $0.2$ \\
CFG Guidance strength $g$ & ---          & $0.5$                 & $1.0$ \\
Diffusion steps $T$    & ---                   & $100$                 & $1000$ \\
Guidance $g$ (op.\ pt.)& ---                   & $0.5$                 & $0.5$ \\
Data mixture           & 5-bucket (Tab.~\ref{tab:s2-buckets}) & 7-bucket (Tab.~\ref{tab:multi-bucket_buckets}) & 7-bucket (Tab.~\ref{tab:multi-bucket_buckets}) \\
\bottomrule
\end{tabular}}
\end{table}

\begin{table}[h]
\centering
\small
\caption{Symbol glossary for the generator formalism (Appendix~\ref{app:generator_formal}).}
\begin{tabular}{lll}
\toprule
Symbol & Definition & Shape \\
\midrule
$\mathbf{L},$ & Lattice vectors & $\mathbb{R}^{3\times3}$ \\
$\mathbf{X}$ & 3D fractional coordinates & $[0,1)^{N_p\times3}$ \\
$\mathbf{A}$ & Atomic numbers & $\{1,...,100\}^{N_p}$ \\
$\mathbf{u}_t=(\mathbf{L}_t,\mathbf{X}_t)$ & Continuous diffusion state at time $t$ & --- \\
$\hat{x}_0$ & Tweedie clean-structure estimate & --- \\
$\sigma(t)$ & SDE diffusion coefficient & scalar function of $t$ \\
$s_\theta(\mathbf{u}_t,\mathbf{A}_t,t\mid\mathbf{C})$ & Score network (GemNet-T) & --- \\
$N_p$ & Atoms per unit cell & integer \\
$T$ & Diffusion / PC iterations & $1000$ \\
$g \in \mathbb{R}_{\ge 0}$ & Classifier-free guidance scale & scalar \\
$p_{\mathrm{drop}}$ & CFG conditioning-dropout probability & $0.2$ \\
$K$ & Output-side atomistic tokens & $K{=}8$ \\
$N$ & Producer context window & $128$ \\
$d_{\mathrm{LM}}$ & LM hidden dim (Qwen3-8B) & $4096$ \\
$M$ & Learnable producer queries & $M{=}16$ \\
$d_{\mathrm{cond}}$ & MatterGen conditioning dim & $512$ \\
$d_h$ & Per-atom hidden dim inside GemNet-T & $512$ \\
$\mathbf{Z}$ & LM hidden states at the $K$ atomistic-token positions & $K \times d_{\mathrm{LM}}$ \\
$\mathbf{S}=[\mathbf{Z}_{\mathrm{ctx}};\mathbf{Z}]$ & Producer source ($N$ context $+$ $K$ atomistic states) & $(N{+}K) \times d_{\mathrm{LM}}$ \\
$\mathbf{Q}_{\mathrm{LQ}}$ & Learnable producer queries & $M \times d_{\mathrm{cond}}$ \\
$\mathbf{C}=f_{\mathrm{QF}}(\mathbf{Q}_{\mathrm{LQ}};\mathbf{S})$ & Producer output (conditioning sequence) & $M \times d_{\mathrm{cond}}$ \\
$\widetilde{\mathbf{C}}(t)$ & Timestep-fused conditioning & $M \times d_{\mathrm{cond}}$ \\
$\mathbf{h}_b$ & Per-atom hidden state at GemNet block $b$ & $N_p \times d_h$ \\
$\gamma_b \in \mathbb{R}$ & Learnable per-block bridge gate (init $1.0$) & scalar \\
\bottomrule
\end{tabular}
\label{tab:app_gen_notation}
\end{table}

\subsection{Teaching language models to natively understand materials through soft tokens}
\label{app:training_objectives}

\textbf{ALM Core} reads each material's 3D structure as continuous soft tokens (one per atom) and answers in natural language. It is trained with a causal language modeling loss over the assistant turn (below); the generation objectives for \textbf{ALM Edit} and \textbf{ALM Gen}, together with their auxiliary terms, are stated with the generator formalism in Appendix~\ref{app:generator_formal}. Notation for the whole appendix can be found above, in Table~\ref{tab:app_gen_notation}.

\subsubsection{Training objectives}

\paragraph{ALM Core.} Each training example is a ChatML-formatted token sequence $\mathbf{w}=(w_1,\dots,w_L)$ with a supervised (assistant turn) index set $\mathcal{S}\subseteq\{1,\dots,L\}$. System and user prompt tokens, as well as the input atomistic soft tokens, are label-masked out. Precisely, these soft tokens are node-wise embeddings of crystals outputted by the atomistic encoder, $\mathbf{H}\in\mathbb{R}^{N_p\times d_{\mathcal{E}}}$, projected into the LLM's input token space $P_{\text{in}}(\mathbf{H})$. The objective is the causal cross-entropy over the supervised positions,
\begin{equation}
\mathcal{L}_{\mathrm{LM}}
\;=\; -\,\mathbb{E}_{\mathbf{w}\sim\mathcal{D}_{\mathrm{S}}}
\!\left[\frac{1}{|\mathcal{S}|}\sum_{i\in\mathcal{S}}
\log p_{\phi}\!\big(w_i \mid w_{<i}\big)\right],
\label{eq:loss_lm}
\end{equation}
where $p_\phi(\cdot\mid w_{<i})$ is the language model's (LM's) next-token distribution evaluated on the $P_{\text{in}}$-spliced input embeddings and $\mathcal{D}_{\mathrm{S}}$ is the \textbf{ALM Core} training dataset (Appendix~\ref{app:training_stage1}). \textbf{Core} is trained with a warm start, in which the LLM is frozen, allowing only the projector $P_{\text{in}}$ to train, for 5 epochs. 

\subsubsection{Atomistic encoder ablations}
\label{app:encoder_ablation}

The choice of OrbV3 as the atomistic encoder $\mathcal{E}$ is justified here. Four architecturally distinct backbones are swept across: OrbV3 (Direct force prediction, 20 neighbor cutoff, trained on OMat, with 256-d per-atom features), UMA-S (Version 1.1 from \texttt{fairchem-core} 2.20.0 with \texttt{task=omat}, 128-d), PET-MAD XS (v1.5.0, 640-d), and PET-MAD S (v1.5.0, 1280-d). They were ablated over by holding the Qwen3-8B base, LoRA $r{=}64$ $\alpha{=}128$, data mixture, and 4,608,000 total samples seen during training. Each arm had its own atomistic-to-language MLP projector (warm started with a similar number of steps) sized to the encoder feature dimension. Three benchmarks supply the downstream metrics: LLM4Mat-Bench MAD/MAE (adjusted to penalize for samples in which the LLM would not output a parseable property), in which a "good model" that is useful to scientists achieves at least a 5; test set Mat2Props raw MAE on narratives, i.e. property prediction given free-form text with structure, property, and application descriptions; and GNoME formation-energy MAE.

\begin{table*}[ht]
\centering
\caption{\textbf{Atomistic encoder ablation.} LLM4Mat-Bench MAD/MAE is leak-adjusted as $\mathrm{RAW} \times (1 - \text{number of unparseable failures})$ ($\uparrow$ better; $\ge 5$ is the paper's ``good model'' threshold); the Mean is over the MP slice's four properties (formation energy, band gap, $E_{\text{hull}}$, density). Mat2Props MAEs are RAW on the MP held-out split ($\downarrow$ better); Validity (Valid.) is the fraction of outputs that are JSON-parseable to a number. GNoME-FE columns are RAW MAE and the rate at which the LLM outputted unparseable values on the GNoME slice.}
\label{tab:encoder-llm4mat}
\small
\setlength{\tabcolsep}{4pt}
\begin{tabular}{l r cc cccc cc c}
\toprule
        &
        & \multicolumn{2}{c}{LLM4Mat ($\uparrow$)}
        & \multicolumn{4}{c}{Mat2Props MAE ($\downarrow$)}
        & \multicolumn{2}{c}{GNoME-FE}
        & Mat2Props \\
\cmidrule(lr){3-4}\cmidrule(lr){5-8}\cmidrule(lr){9-10}
Encoder & dim
        & Mean & $\ge 5/9$
        & bg & $E_f$ & $E_h$ & $\rho$
        & MAE ($\downarrow$) & Leak ($\downarrow$)
        & Valid.\ ($\uparrow$) \\
\midrule
\textbf{ORB}     & $256$
        & $\mathbf{6.42}$  & $\mathbf{4/9}$
        & $\mathbf{0.244}$ & $0.085$          & $\mathbf{0.070}$ & $0.157$
        & $0.026$          & $\mathbf{2.6\%}$
        & $\mathbf{98.3\%}$ \\
UMA-S            & $128$
        & $2.34$           & $0/9$
        & $0.319$          & $0.101$          & $0.128$          & $0.541$
        & $\mathbf{0.015}$ & $52\%$
        & $47\%$ \\
PET-XS           & $640$
        & $0.70$           & $0/9$
        & $0.503$          & $0.774$          & $0.100$          & $0.295$
        & $0.125$          & $32\%$
        & $20\%$ \\
PET-S            & $1280$
        & $4.48$           & $2/9$
        & $0.371$          & $\mathbf{0.083}$ & $0.087$          & $\mathbf{0.160}$
        & $0.030$          & $35\%$
        & $71.5\%$ \\
\bottomrule
\end{tabular}
\end{table*}

% \begin{figure}[ht]
% \centering
% \includegraphics[width=0.78\linewidth]{figures/appendix/encoder_ablation_bars.pdf}
% \caption{\textbf{Atomistic encoder ablation in Table~\ref{tab:encoder-llm4mat}, visualized.} }
% \label{fig:encoder_ablation_bars}
% \end{figure}

OrbV3 is the only encoder that clears the LLM4Mat-Bench ``good model'' threshold on a non-trivial number of configs ($4/9$ at leak-adjusted MAD/MAE $\ge 5$; PET-S $2/9$, the others $0/9$), and the only arm whose checkpoint reaches the $\ge 98\%$ Mat2Props validity target at effective batch $256$. UMA-S has the lowest raw GNoME formation-energy MAE ($0.015$~eV/atom on the surviving $\sim 47\%$), but its $52\%$ URL-leak rate collapses the leak-adjusted comparison; PET-XS sits below both axes. PET-S is the only non-OrbV3 arm competitive on Mat2Props (winning formation energy and density at raw MAE), but below the validity $\ge 95\%$ floor.  Richer feature dimensions ($d \ge 640$) do not translate into understanding alignment within the $12$k-step budget, because the chat-template training has to relearn a projector geometry that OrbV3's $256$-d features happen to provide nearly for free. The validity problem, in which Qwen3-8B, when finetuned, starts to regress to its pretraining priors, outputting IMGUR URLs or mangled JSON instead of the valid JSON they are prompted to write (discussed further in Appendix~\ref{app:training_stage2}).

On the other hand, there is recent evidence that machine-learned interatomic potentials \emph{converge in representation space} as they improve. ~\cite{edamadaka_universally_2025} shows across nearly sixty scientific foundation models that, on inputs similar to those seen during training, high-performing models align closely in latent space, with weaker models diverging into local sub-optima. Our four encoders are all competent MLIPs trained on overlapping bulk-crystal distributions, so on the in-distribution LLM4Mat-Bench / GPT-Narratives materials we evaluate, they expose nearly the same structural signal to the projector; the gaps in Table~\ref{tab:encoder-llm4mat} are dominated not by representational content but by feature dimensionality (richer $d$ slows projector alignment within a fixed step budget) and by each arm's downstream calibration (the URL-leak and validity floors). OrbV3 was chosen, then, doubly because of its inference speed. In practice, node-wise embeddings of all train, validation, and test set structures were cached and retrieved during each phase.

\subsubsection{Continuous atomistic-to-language adapter}
\label{app:codebook_comparison}

On the input side, the frozen OrbV3 encoder emits a variable-length per-atom representation ($N$ tokens of $256$ dimensions for an $N$-atom cell) that must be mapped into the language model's $d_{\mathrm{LM}}=4096$ embedding space before being prepended to the prompt tokens. ALMs use the simplest possible \emph{adapter}: a two-layer GELU MLP, $\mathrm{Linear}(256\!\to\!4096)\to \mathrm{GELU}\to\mathrm{Linear}(4096\!\to\!4096)$ ($\approx\!21$M parameters, with full parameter counts in Appendix~\ref{app:lora_config}). This follows the LLaVA projector convention~\cite{liu_visual_2023,liu_improved_2024}, which uses the identical two-linear-with-GELU for vision language modeling. Crucially, the adapter is \emph{token-preserving}: it emits one LM token per atom and leaves the encoder's variable-resolution output intact, so the LM's expanded sequence length scales \emph{with} the structure rather than being compressed against a fixed budget.

Three families of cross-modal interface that have been successful in vision-language modeling each fail here because crystalline matter does not concentrate on a small set of recurring modes the way natural images do.

\paragraph{Learned-query bottlenecks (Q-Former, Perceiver IO).} Q-Former-style adapters~\cite{li_blip2_2023} and Perceiver~IO~\cite{jaegle_perceiver_2021} compress the foreign modality through a small fixed set of cross-attended learned queries (typically, $32$ to $64$). The construction assumes the encoder's output is well summarized by $L \ll N$ tokens. For a crystal of $N$ atoms this assumption fails in both directions: when $N$ is small ($N \le L$) the bottleneck wastes capacity on null queries; when $N$ is large it forces a fixed-budget compression on a representation whose information density grows with atom count. A $1000$-atom defect supercell carries information that does not exist in any smaller subset of its atoms, and a single fixed $L$ could struggle to represent both regimes at once. The two-layer MLP sidesteps the trade-off entirely by keeping the per-atom representation variable-length on the encoder side; a fixed-length learned-query producer is reserved for the \textbf{ALM Edit} and \textbf{Gen} \emph{decoder} side, where the diffusion sampler accepts a fixed-size conditioning tensor by construction (detailed in Appendix~\ref{app:gen_producer}).

\paragraph{Codebook quantization (VQ-VAE, FSQ, JANUS).} VQ-VAE~\cite{oord_neural_2018}, FSQ~\cite{mentzer_finite_2024}, and JANUS~\cite{chen_janus-pro_2025} round the latent representation to one of $V$ discrete entries. The resolution requirement here is severe. The $\mathrm{TiO}_2$ polymorph case is illustrative, as mentioned in the main text. A codebook small enough to be trainable may struggle to resolve both regimes, and growing the codebook to that resolution simply converges to leaving the representation continuous in the limit. ALMs therefore keep the latent continuous and let the LM's own attention scores, rather than a discrete code, decide atomic detail.

Overall, a richer encoding would add more training burden and computational cost than benefit, as the real representational learning is done elsewhere: the frozen OrbV3 encoder already produces a physically grounded per-atom embedding (Appendix~\ref{app:encoder_ablation}), and the LM attends over those tokens. The adapter's only job is a per-atom linear lift into the LM embedding space, for which two layers with a GELU nonlinearity are sufficient capacity. Ablations that pooled all embeddings into a single token or 3 tokens unnecessarily gated the amount of information flowing to the LM, decreasing its property prediction performance. Empirically, warm start (projector-only training) alignment loss drops sharply within the first few hundred steps and plateaus near $0.10$ (Appendix~\ref{app:training_stage1}), producing nearly perfect structural descriptions, confirming the dimensional lift is learned cleanly with this light, low-parameter map.

\subsection{Guiding crystal denoising with language model embeddings}
\label{app:generator_formal}

The ALM generator pairs Qwen3-8B with a pretrained crystal \emph{diffusion decoder}, MatterGen, and steers that decoder through a learned conditioning channel via CFG. This section formalizes the bridge architecture proposed in the main text, as well as validates the choice of a diffusion decoder and of MatterGen specifically.

\paragraph{Choice of diffusion models as decoders.} Conventional materials generation takes the form of \textit{de novo} discovery (\almGen{}) and crystal structure prediction (\almEdit{}). Score-based diffusion over the periodic-crystal manifold held the state of the art on both at the start of this work~\cite{jiao_crystal_2023, zeni_generative_2025}, jointly denoising lattice, fractional coordinates, and atomic numbers with periodic-translation and point-group symmetries often built in. Two properties make it the natural bridge target. First, the score network factors into an unconditional backbone and a conditioning branch, so an external producer attaches without retraining the backbone. Second, classifier-free guidance (CFG) exposes a single scalar $g$ that interpolates from the unconditional backbone ($g{=}0$) to the fully conditioned model ($g{>}1$ extrapolates) --- an inference-time dial on conditioning strength, decoupled from training, that underlies the stability tension analyzed below. Autoregressive string decoders (CrystaLLM~\cite{antunes_crystal_2024}, Crystal-Text-LLM~\cite{gruver_fine-tuned_2024}) expose no such dial, and VQ/codebook decoders quantize the latent we want kept continuous.

\paragraph{Choice of MatterGen as the decoder backbone.} We built on MatterGen~\cite{zeni_generative_2025} because of several reasons. Firstly, its GemNet-T score network accepts an external conditioning sequence through an adapter interface exposed in the released code, so a cross-attention bridge~\cite{ye_ip-adapter_2023} needs only new key/value/gate parameters and no backbone surgery. Second, its stability-filtered \texttt{mattergen\_base} checkpoint (Alex-MP-20) carries the metastability prior our training mixture lacks (quantified below), and serves as the \textbf{ALM Gen} backbone. Lastly, its from-scratch CSP-mode configuration, which observes atom types rather than denoising over them, was precisely what was retrained from scratch to serve as the base denoising model which \textbf{ALM Edit} used CFG to guide.
\label{app:gen_notation}
Table~\ref{tab:app_gen_notation} fixes the symbols used across all of Appendix~\ref{app:generator_formal}. A periodic crystal is factored into a continuous lattice $\mathbf{L}$, continuous fractional coordinates $\mathbf{X}$, and discrete atomic numbers $\mathbf{A}$; the continuous components are bundled into the diffusion state $\mathbf{u}_t=(\mathbf{L}_t,\mathbf{X}_t)$. We write the SDE diffusion coefficient as $\sigma(t)$ and the GemNet consumer block index is $b$.

\subsubsection{Denoising diffusion model training and objectives}
\label{app:gen_diff}

\begin{figure}[ht]
\centering
\includegraphics[width=0.46\linewidth]{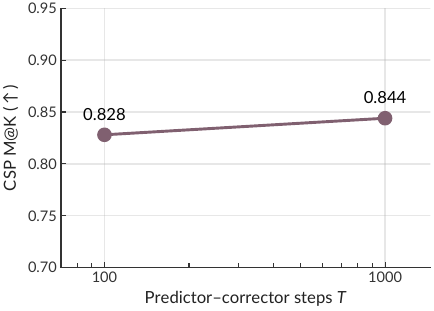}
\caption{\textbf{CSP M@K=64 is flat in denoising timesteps $T$.} (\textbf{ALM Edit} , MP-20).}
\label{fig:csp_step_sweep}
\end{figure}

MatterGen factorizes a periodic crystal into a continuous lattice $\mathbf{L}\in\mathbb{R}^{3\times 3}$, continuous fractional coordinates $\mathbf{X}\in[0,1)^{N_p\times 3}$, and a discrete atomic-number assignment $\mathbf{A}\in\{1,\dots,100\}^{N_p}$. The continuous components $\mathbf{u}=(\mathbf{L},\mathbf{X})$ evolve under a score-based forward SDE,
\begin{equation}
\mathrm{d}\mathbf{u}_t \;=\; \mathbf{f}(\mathbf{u}_t, t)\,\mathrm{d}t \;+\;
  \sigma(t)\,\mathrm{d}\mathbf{w}_t,
  \qquad \mathbf{u} = (\mathbf{L},\mathbf{X}),
\label{eq:app_gen_sde_forward}
\end{equation}
while the discrete atomic numbers diffuse under an absorbing-state D3PM~\cite{austin_structured_2021} whose forward kernel mixes each clean type toward an absorbing \textsc{mask} state with cumulative probability $\bar\beta_t$,
\begin{equation}
q(\mathbf{A}_t \mid \mathbf{A}_0) \;=\;
  \mathrm{Cat}\big(\mathbf{A}_t \,;\,
  (1-\bar\beta_t)\,\delta_{\mathbf{A}_0} \,+\, \bar\beta_t\,\delta_{\mathrm{MASK}}\big).
\label{eq:app_gen_d3pm}
\end{equation}
The score network $s_\theta(\mathbf{u}_t, \mathbf{A}_t, t \mid \mathbf{C})$ is parameterized by GemNet-T and depends on the conditioning sequence $\mathbf{C}$ exactly through the consumer cross-attention branch of Eq.~\eqref{eq:app_gen_combined}. Reverse-time generation integrates
\begin{equation}
\mathrm{d}\mathbf{u}_t \;=\; \big[
  \mathbf{f}(\mathbf{u}_t, t) - \sigma(t)^2\, s_\theta(\mathbf{u}_t, \mathbf{A}_t, t \mid \mathbf{C})
\big]\,\mathrm{d}t \;+\; \sigma(t)\,\mathrm{d}\bar{\mathbf{w}}_t,
\label{eq:app_gen_sde_reverse}
\end{equation}
paired with the D3PM reverse step for $\mathbf{A}_t$. The number of predictor-corrector iterations, or diffusion timesteps, is 1000 for all models; \almEdit{} CSP match-rate is flat in this count (Fig.~\ref{fig:csp_step_sweep}).

The bridge is trained with classifier-free-guidance dropout: with probability $p_{\mathrm{drop}}{=}0.2$ per step the \texttt{alm\_embedding} conditioning is replaced by a learned zeros vector, so the network jointly learns the conditional and unconditional scores. At inference we apply the standard CFG extrapolation,
\begin{equation}
\widetilde{s}_\theta(\mathbf{u}_t, \mathbf{A}_t, t \mid \mathbf{C}, g)
\;=\; s_\theta(\cdot \mid \varnothing) + g \cdot \big(
  s_\theta(\cdot \mid \mathbf{C}) - s_\theta(\cdot \mid \varnothing)\big),
\label{eq:app_gen_cfg}
\end{equation}
where $g{=}0$ recovers the unconditional MatterGen distribution exactly.

\paragraph{Shared diffusion objective.} Both generative models train the GemNet-T score network $s_\theta$ against the MatterGen denoising loss. For a conditioning signal $C$, with the corrupted state $(\mathbf{u}_t,\mathbf{A}_t)$ from the forward SDE (Eq.~\ref{eq:app_gen_sde_forward}) and the absorbing-state D3PM (Eq.~\ref{eq:app_gen_d3pm}),
\begin{multline}
\mathcal{L}_{\mathrm{diff}}(C)
=\mathbb{E}_{\substack{(\mathbf{u}_0,\mathbf{A}_0)\sim\mathcal{D}_3\\ t\sim\mathcal{U}\{1,\dots,T\}}}
\!\bigg[
\underbrace{\omega(t)\,\big\|\, s_\theta(\mathbf{u}_t,\mathbf{A}_t,t\mid C)
-\nabla_{\mathbf{u}_t}\log p_t(\mathbf{u}_t\mid\mathbf{u}_0)\,\big\|^2}_{\text{lattice }\mathbf{L}+\text{ coords }\mathbf{X}\ \text{(score matching)}} \\
\;-\;
\underbrace{\log p_\theta\!\big(\mathbf{A}_0 \mid \mathbf{A}_t, t, C\big)}_{\text{atom types }\mathbf{A}\ \text{(absorbing D3PM)}}
\bigg],
\label{eq:loss_diff}
\end{multline}
with $p_t(\mathbf{u}_t\mid\mathbf{u}_0)$ the per-field forward kernel ($\mathbf{X}$ uses the periodic wrapped-normal kernel) and $\omega(t)$ the standard denoising weight (we reserve $\lambda$ for the Feynman-Kac log-weight and $g$ for the CFG scale). The conditioning $C$ is produced by LLM encoder $\phi$ from the text prompt paired with $(\mathbf{u}_0,\mathbf{A}_0)$.

\paragraph{ALM Edit.} \textbf{Edit} conditions on the Q-Former producer output $\mathbf{C}=f_{\mathrm{QF}}(\mathbf{Q}_{\mathrm{LQ}};\mathbf{S})$, the $M{=}16$-token conditioning sequence (timestep-fused to $\widetilde{\mathbf{C}}(t)$ inside $s_\theta$). It is trained with CFG conditioning dropout at rate $p_{\mathrm{drop}}$:
\begin{equation}
\mathcal{L}_{\mathrm{CSP}}
=\mathbb{E}_{\xi\sim\mathrm{Bern}(p_{\mathrm{drop}})}
\!\Big[\,\mathcal{L}_{\mathrm{diff}}\big((1-\xi)\,\mathbf{C}+\xi\,\varnothing\big)\Big],
\label{eq:loss_csp}
\end{equation}
i.e.\ with probability $p_{\mathrm{drop}}{=}0.2$, the producer sequence $\mathbf{C}$ is replaced by the learned null $\varnothing$, so $s_\theta$ learns the conditional and unconditional scores that the CFG mixing of Eq.~\ref{eq:app_gen_cfg} extrapolates at $g{=}0.5$. The score network is trained from scratch and LLM $\phi$ is fully fine-tuned.

\paragraph{ALM Gen.} \textbf{Gen} replaces the Q-Former producer with a lightweight per-token projector $P_{\text{out}}$, which maps each of the $K$ atomistic-token hidden states \emph{independently} into the conditioning sequence $\mathbf{C}=P_{\text{out}}(\mathbf{Z})\in\mathbb{R}^{K\times d_{\mathrm{cond}}}$ with $[\,P_{\text{out}}(\mathbf{Z})\,]_k=\mathrm{MLP}(\mathbf{Z}_k)$ --- no learned queries, no context window, and no pooling --- and feeds the resulting $K$-token sequence to the \emph{same} IP-Adapter cross-attention consumer as \textbf{Edit}. Training uses an identical CFG-dropout objective:
\begin{equation}
\mathcal{L}_{\mathrm{DNG}}
=\mathbb{E}_{\xi\sim\mathrm{Bern}(p_{\mathrm{drop}})}
\!\Big[\,\mathcal{L}_{\mathrm{diff}}\big((1-\xi)\,\mathbf{C}+\xi\,\varnothing\big)\Big].
\label{eq:loss_dng}
\end{equation}
The objective is identical to Eq.~\ref{eq:loss_csp}. The differences from \textbf{Edit} are the producer (a per-token MLP emitting the $K$-token sequence $\mathbf{C}$, versus\ the $M{=}16$-query Q-Former --- both consumed by the \emph{same} per-block IP-Adapter cross-attention), the LLM adaptation (LoRA $r{=}8$, versus\ full fine-tuning), and the backbone (MatterGen-Base in DNG mode, versus\ the from-scratch CSP-mode score network).

\label{app:aux_losses}

On top of the headline objectives (Eqs.~\ref{eq:loss_lm}--\ref{eq:loss_dng}), both \textbf{ALM Edit} and \textbf{Core} generation models add a per-element composition-count term ($\lambda_{\text{aux}}{=}1.0$) and a contrastive term ($\lambda_{\text{contr}}{=}0.02$); \almEdit{} adds a third, directional term ($\lambda_{\text{dir}}{=}0.1$, with $\lambda_{\text{dir}}{=}0$ for the de-novo \textbf{Core} model). The full generation loss is
\begin{equation}
\begin{aligned}
\mathcal{L}_{\text{gen}}
={}& \mathcal{L}_{\text{diff}}\!\bigl(s_\theta(\mathbf{u}_t,\mathbf{A}_t,t\mid\mathbf{C}),\,\mathbf{u}_0,\mathbf{A}_0\bigr) \\[2pt]
&+ \lambda_{\text{aux}}\, \mathcal{L}_{\text{count}}\!\bigl(g_{\text{aux}}(\mathbf{C}),\, c(x)\bigr)
+ \lambda_{\text{contr}}\, \mathcal{L}_{\text{contr}}
+ \lambda_{\text{dir}}\, \mathcal{L}_{\text{dir}},
\end{aligned}
\label{eq:loss_gen_aux}
\end{equation}
with the per-element composition presence loss a class-balanced binary cross-entropy ---
\begin{equation}
\mathcal{L}_{\text{count}}
= -\frac{1}{N_Z}\sum_{z=1}^{N_Z}\Bigl[\, w_{+}\, c_z(x)\,\log\sigma(s_z)
+ \bigl(1-c_z(x)\bigr)\log\bigl(1-\sigma(s_z)\bigr)\Bigr],
\label{eq:loss_count}
\end{equation}
the contrastive (decorrelation) term
\begin{equation}
\mathcal{L}_{\text{contr}}
= \frac{1}{B(B-1)}\sum_{i\neq j}\!\left(\frac{\bar{\mathbf{c}}_i^{\top}\bar{\mathbf{c}}_j}{\lVert\bar{\mathbf{c}}_i\rVert\,\lVert\bar{\mathbf{c}}_j\rVert}\right)^{\!2},
\label{eq:loss_contr}
\end{equation}
and the directional term (\almEdit{} only)
\begin{equation}
\mathcal{L}_{\text{dir}} = \frac{1}{|\mathcal{F}|}\sum_{i\in\mathcal{F}} \mathrm{CE}\!\bigl(W_{\text{dir}}\,\bar{\mathbf{c}}_i,\; y_i\bigr),
\qquad y_i = \mathbb{1}\!\left[\text{prompt } i \text{ raises the target property}\right].
\label{eq:loss_dir}
\end{equation}
Here, $\mathbf{Z}\in\mathbb{R}^{K\times d_{\mathrm{LM}}}$ are the atomistic-token hidden states extracted from the language model, and $\mathbf{C}$ is the producer output that conditions the score network $s_\theta$ (Section~\ref{app:gen_producer}). Auxiliary heads like $g_{\text{aux}}$ (a per-element atom-count regression head trained with a BCE presence objective) operate on $\mathbf{C}$ directly, allowing gradients to reach the producer without passing through the score network. In $\mathcal{L}_{\text{count}}$, $s = g_{\text{aux}}(\mathbf{C})\in\mathbb{R}^{N_Z}$ are per-element presence logits, $c(x)\in\{0,1\}^{N_Z}$ is the multi-hot composition of the target structure ($c_z(x){=}1$ iff element $z$ is present), $N_Z{=}100$ spans $Z\in\{1,\dots,100\}$, and $w_{+}{=}32$ up-weights the rare present-element class. In $\mathcal{L}_{\text{contr}}$, $\bar{\mathbf{c}}_i$ is the producer output for prompt $i$ averaged over its $M$ conditioning tokens and $B$ is the batch size, so $\mathcal{L}_{\text{contr}}$ is the mean squared off-diagonal cosine similarity across the batch. Intuitively, this is a decorrelation penalty pushing distinct prompts toward distinct conditioning vectors. In $\mathcal{L}_{\text{dir}}$, $W_{\text{dir}}\in\mathbb{R}^{2\times d_{\text{cond}}}$ is a learned linear head, $\bar{\mathbf{c}}_i$ is the same $M$-token-pooled producer output, and $\mathcal{F}$ is the subset of \emph{directional} prompts, rows carrying an explicit raise/lower instruction (non-directional rows are masked out). The direction label is \emph{not} a MatterGen cond\_field: it never reaches the diffusion decoder and trains only the producer, forcing $\mathbf{C}$ to be linearly separable by direction, the targeted fix for the near-collinear ``raise'' and ``lower'' conditioning vectors of Appendix~\ref{app:gen_producer}. Finally, $\mathcal{L}_{\text{diff}}$ is the standard SDE/D3PM loss on the lattice, fractional coordinates, and atomic numbers $(\mathbf{L},\mathbf{X},\mathbf{A})$ (Section~\ref{app:gen_diff}). Equation~\eqref{eq:loss_gen_aux} is the loss actually optimized during generation training, but only $\mathcal{L}_{\text{diff}}$ flows through the score network via CFG. The auxiliary losses are purely regularizers added to shape the latent steering vectors (producer output $\mathbf{C}$, and thus the atomistic-token states). \textit{We present systematic ablations to validate the presence of each term below.}

\begin{figure}[ht]
  \centering
  \includegraphics{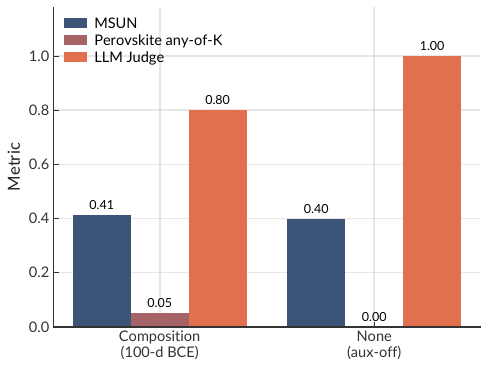}
  \caption{\textbf{Auxiliary supervision target comparison.} Composition BCE (the optimum) vs aux-off across MSUN, perovskite any-of-$K$, and the LM judge; aux-off keeps MSUN but loses prompt-following entirely.}
  \label{fig:aux_target}
\end{figure}

\paragraph{Ablation results for auxiliary losses.} To start, the composition auxiliary head is essential. Lowering the $\lambda_{\text{aux}}{=}1.0$ to $0.0$ degrades de-novo metastable-SUN (MSUN) by 2\%, but increasing it to $3.0$ decreased MSUN by 49\%. This effect is not only present in latent space, but also in observable statistics on generated samples. Figure~\ref{fig:aux_target} shows how turning the compositional auxiliary loss off loses prompt conditioning on difficult tasks, like asking \textbf{ALM Edit} to ``generate a perovskite'' and checking if any of $K=20$ generated samples are perovskites (middle bar in each group). However, as \textbf{ALM Gen} is weakly conditioned on the atomistic tokens, the MSUN of its generated structures doesn't change drastically. With the atomistic token embeddings collapsing to a single, low-information vector, the model does not need to shift away from its language model priors, allowing it to stay performant on LLM judge evaluations of materials science knowledge. 

The contrastive loss over atomistic token hidden states $\mathbf Z$ is also essential; without it, the $\mathbf{Z}$ collapses to a cosine distance of 0.12 between different prompts. With $\lambda_{\text{aux}}{=}1$, the average cosine distance between $\mathbf Z$ across all prompts in the evaluation dataset is $0.85$.

%%% ===== frag_I_A.tex =====

\subsubsection{Separating crystal structure prediction from de-novo generation}
\label{app:cfg_tension}

\begin{table}[ht]
  \centering
  \caption{\textbf{MatterGen architectures trade off between \textit{de novo} generation and crystal structure prediction performance.} Metastability of generated samples versus crystal structure prediction (CSP) performance. The two denoising diffusion models are architecturally identical at $g=0$, but differ only in that MatterGen CSP does not denoise over element types, instead taking them in as input.}
  \label{tab:backbone_tradeoff}
  \footnotesize
  \begin{tabular}{lcc}
  \toprule
  Backbone & Metastability ($E_{\text{hull}} \leq 0.1$) & CSP M@K=128 (MP-20) \\
  \midrule
  \textbf{ALM Gen @} $g=0$ (MatterGen Base) & $\mathbf{0.750}$ & $0.370$ \\
  \textbf{ALM Edit @} $g=0$ (MatterGen CSP)   & $0.167$          & $\mathbf{0.777}$ \\
  \bottomrule
  \end{tabular}
  \end{table}

The architectural tension that leads to \textbf{ALM Edit} and \textbf{ALM Gen} being separate models is solely dependent on the denoising diffusion model $\mathcal{D}$. A fundamental limitation of MatterGen is that it trades composition and stoichiometry obeyance with stability of generated structures. When provided with the exact element count of a desired structure through CFG, Mattergen Base tends to generate crystals with similar compositions, but that are more stable (\textbf{T2C-FK} leverages the fact that denoised compositions don't stray too far to enforce composition and stoichiometry following). Mattergen CSP, which has the same architecture as Base but does not denoise over element types (instead taking them as input to directly initialize node embeddings), produces structures with far lower energy. There are two additional factors beyond architecture that help explain the disparity between each architecture in metastability and CSP: the underlying training data and guidance scale $g$, which also have strong interplay, as discussed below. 

\begin{figure}[ht]
  \centering
  \includegraphics[width=\linewidth]{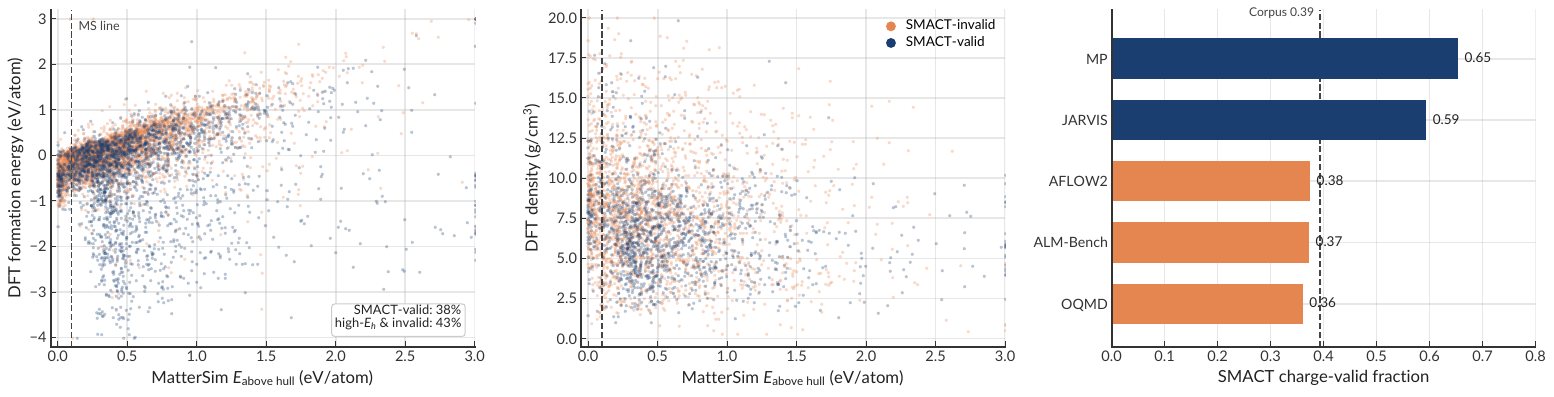}
  \caption{\textbf{SMACT charge-validity of the training compositions.} \emph{Left, middle:} MatterSim $E_h$ versus\ DFT formation energy and density, coloured by SMACT charge-validity. Most training compositions are charge-invalid and high-$E_h$ (lower-right quadrant). \emph{Right:} SMACT charge-valid fraction by training source.}
  \label{fig:smact_parity}
\end{figure}

\paragraph{Underlying data carry stability and validity biases.} The polymorphs in MP-20 don't always have the lowest energy out of other geometries and are not guaranteed to be metastable. MatterGen Base, which generates structures with stabilities at similar rates to its training data, would suffer from lower stability and SUN performance after training on MP-20 CSP. MatterGen CSP, which closely learns how to predict polymorphs from given compositions in MP-20, thus also learns to produce the labeled polymorphs, thus generating structures with lower stability than a \textit{de novo} model that only trains on metastable structures. Another example of this is SMACT validity, a common metric for realistic structures including charge neutrality, electronegativity, and mixed-valence checks. \textbf{ALM Gen} and \textbf{ALM Edit} don't produce a high proportion of SMACT-valid structures. A large, contributing factor to this is our training data, which, as shown in Fig.~\ref{fig:smact_parity} to the right, are only 39\% SMACT-valid on average. In addition, many SMACT-valid structures are high energy (Fig.~\ref{fig:smact_parity} left and middle), while many example materials with performant band gaps and properties of interest are SMACT-invalid (OQMD Fig.~\ref{fig:smact_parity} right).

\begin{figure}[ht]
  \centering
  \includegraphics[width=0.5\linewidth]{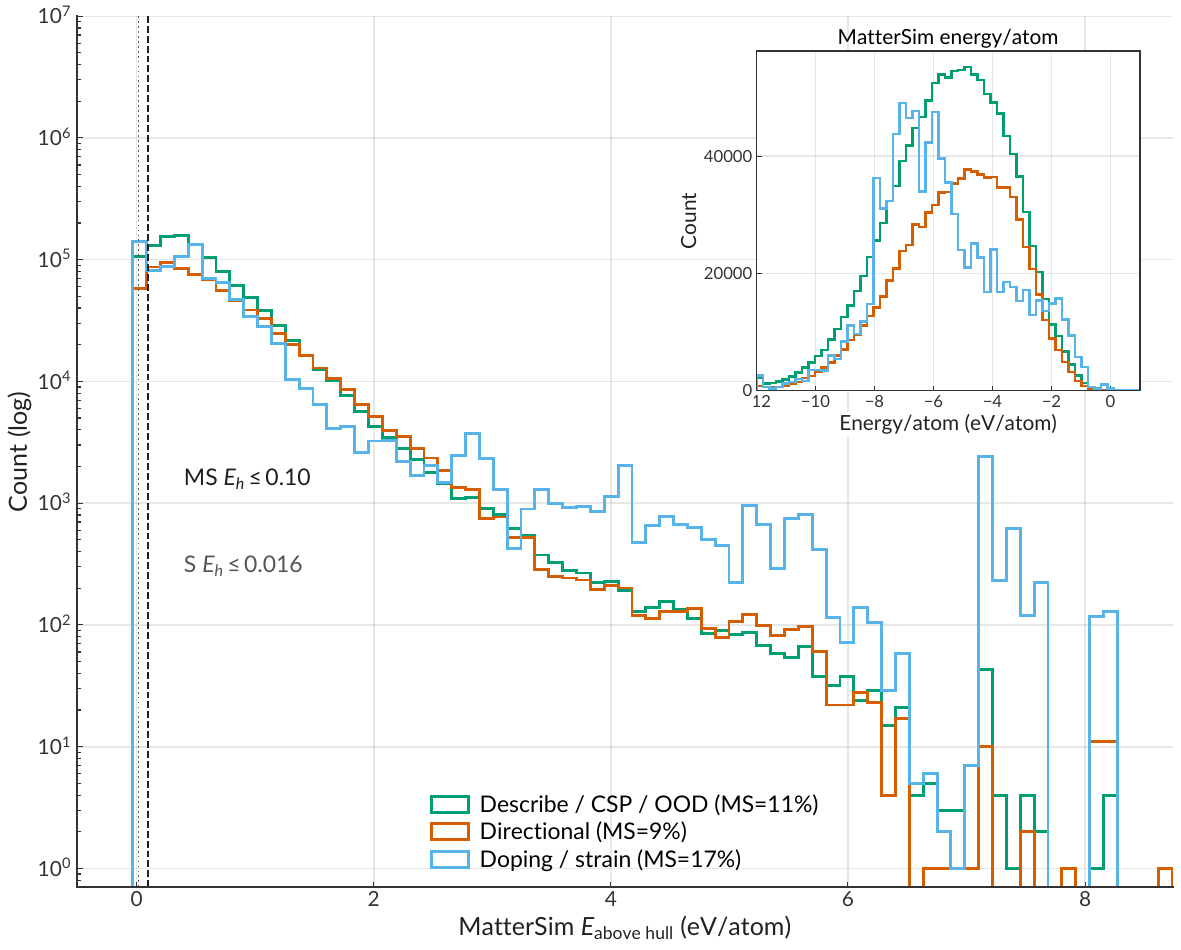}
  \caption{\textbf{Right-tailed energy distribution of the generation training corpus.} Main shows per-bucket MatterSim energy above hull ($E_h$, eV/atom) on a log-count axis over the full range. The mass of the distribution sits far above the hull with a tail extending to ${\sim}8$~eV/atom (the $92$ percentile are OQMD/AFLOW structures). The metastable ($E_h{\le}0.10$, dashed) and stable ($E_h{\le}0.016$, dotted) thresholds are marked, and the inset shows MatterSim-evaluated energy-per-atom for the same buckets. A conditioned generator would reproduce this unstable distribution.}
  \label{fig:dist_e_above_hull}
\end{figure}

In addition, the aggregate distribution of materials that we post-train both models on has a large amount of volume away from the energetic hull ($E_{\text{h}}$ or $E_{\text{above hull}} \leq 0$), as seen in Fig.~\ref{fig:dist_e_above_hull}.

Further, when the MatterGen CSP mode model was pretrained on our dataset, its Match@$K=20$ performance for MP-20 was 4\% higher than the same architecture trained on MP-20-only, the same scale of difference as between \textbf{ALM Edit} and the second best model at CSP. Therefore, the ceiling for the stability is bounded by the pre-trained MatterGen diffusion model for \textit{de novo} generation and the data that it was trained on.

\begin{figure}[ht]
  \centering
  \includegraphics[width=\linewidth]{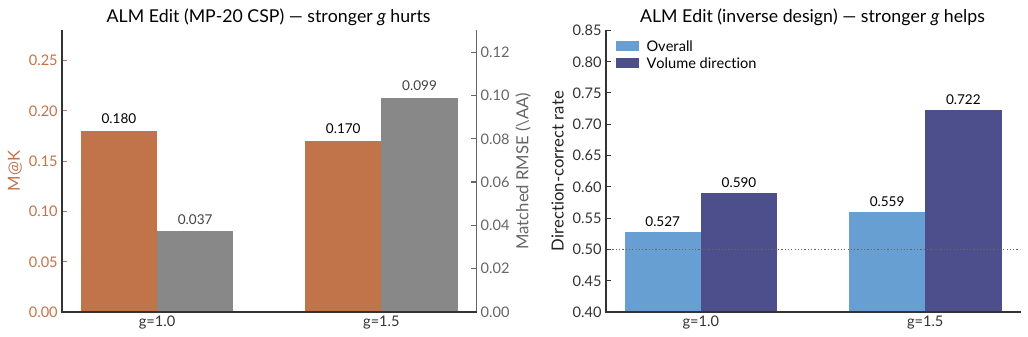}
  \caption{\textbf{ALM Edit CFG guidance scale tradeoff between CSP and inverse design.} Stronger $g$ hurts crystal structure prediction (left, Match@K and RMSE for matches) but helps \textbf{ALM Bench} inverse design performance.}
  \label{fig:cfg_inversion}
\end{figure}

\paragraph {The effect of guidance scale $g$ on generation stability and crystal structure prediction.} The guidance parameter $g$ controls how much base denoising diffusion models obey the conditional priors instilled by the additional data they were finetuned on, by CFG's design. A model trained to reproduce an unstable distribution will, when conditioned, reproduce its instability. However, $g$ offers the possibility of a tradeoff between instruction-following and generation quality (as measured by metrics like stability). Fig.~\ref{fig:cfg_inversion} shows how $g$ has different effects for a variant of the backbone used for \textbf{ALM Edit}. Raising $g$ hurts CSP performance while helping editing tasks, as it controls the strength of the task- and input composition-encoding conditional signal from the language-to-atomistic bridge. This information is crucial for improving performance on \textbf{ALM Bench} tasks, but as \textbf{Edit} autoregressively generates the composition for the MatterGen CSP backbone when doing crystal structure prediction, any additional conditioning may pull the model away from its already strong performance. 

\begin{figure}[ht]
  \centering
  \includegraphics{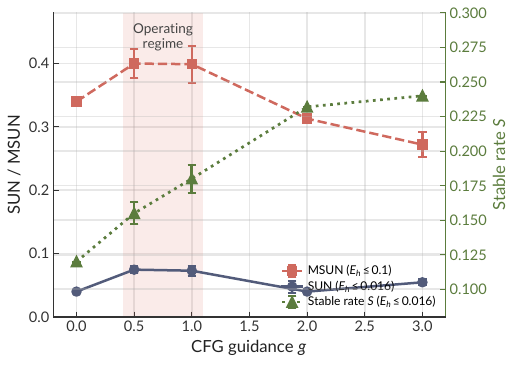}
  \caption{\textbf{CFG guidance scale sweep for ALM Gen.} SUN ($E_h \leq 0.016$, left axis, circles) and MSUN ($E_h \le 0.1$, right axis, squares) depend heavily on $g$.}
  \label{fig:g_sweep_8724}
\end{figure}

The guidance score produces very different behavior for \textbf{ALM Gen}. Here, $g$ controls how much a global conditioning vector steers Mattergen Base away from its strongly performing frozen base. There is a positive operating range for $g$, leading to the choice of $g=1.0$, as shown in Fig.~\ref{fig:g_sweep_8724}. 
  
\subsubsection{The producer--consumer bridge}
\label{app:gen_overview}

Our language-to-atomistic bridge architecture enables the language model $\phi$ to guide the crystal diffusion decoder $\mathcal{D}$'s generation. Let $\mathbf{Z}\in\mathbb{R}^{K\times d_{\mathrm{LM}}}$ be the final-layer hidden states of the $K{=}8$ atomistic-token positions emitted by $\phi$ (Table~\ref{tab:app_gen_notation}). These $K$ tokens are randomly initialized and added to the model's vocabulary before training (we ablated over $K$ values of 4, 8, and 16, and found that $K=8$ balanced performance and computational cost). The producer reads these $K$ states together with the window of $N{=}128$ language-model hidden states immediately preceding them, forming the source sequence $\mathbf{S}=[\mathbf{Z}_{\mathrm{ctx}};\,\mathbf{Z}]\in\mathbb{R}^{(N+K)\times d_{\mathrm{LM}}}$ (length $N{+}K{=}136$), with a learned type embedding marking the $N$ context states apart from the $K$ atomistic states; the selected window width $N{=}128$ is ablated in Figure~\ref{fig:qformer_window}.

The \emph{producer} is a shallow learnable-query transformer~\cite{koh_gill_2023,li_blip2_2023} that compresses $\mathbf{S}$ into a fixed-shape conditioning sequence
\begin{equation}
\mathbf{C} \,=\, f_{\mathrm{P}}\!\big(\mathbf{Q}_{\mathrm{LQ}};\,\mathbf{S}\big)
\;\in\;\mathbb{R}^{M\times d_{\mathrm{cond}}},
\qquad M{=}16,\;d_{\mathrm{cond}}{=}512,
\label{eq:gen_short_producer}
\end{equation}
where $\mathbf{Q}_{\mathrm{LQ}}\in\mathbb{R}^{M\times d_{\mathrm{cond}}}$ is a set of continuous, learnable queries that cross-attend into the LM-side source $\mathbf{S}$.

The \emph{consumer} injects $\mathbf{C}$ into every block of $\mathcal{D}$'s GemNet-T score network as an IP-Adapter--style cross-attention head~\cite{ye_ip-adapter_2023,wang_msdiffusion_2024} --- the only cross-attention in the network. MatterGen's native property conditioning does not enter through cross-attention: it is concatenated into a per-crystal latent broadcast into the atom embedding, and is therefore already carried by the frozen backbone block. Writing $\Psi_b$ for that frozen block update and $\mathbf{h}_b$ for the per-atom hidden state at GemNet block $b$, the bridge adds a single gated read-out:
\begin{equation}
\boxed{\;
\begin{aligned}
\mathbf{h}_b \,\leftarrow\,\;& \underbrace{\Psi_b(\mathbf{h}_b)}_{\text{frozen backbone (native conditioning)}} \\
&\,+\, \gamma_b\cdot \underbrace{W^{\mathrm{mix}}_b\,\mathrm{Attn}\big(\Psi_b(\mathbf{h}_b)\, W^Q_b,\, \widetilde{\mathbf{C}}(t) W^{K,\mathrm{alm}}_b,\, \widetilde{\mathbf{C}}(t) W^{V,\mathrm{alm}}_b\big)}_{\text{new (trained): the network's only cross-attention}}.
\end{aligned}
\;}
\label{eq:gen_short_combined}
\end{equation}
The bridge contributes only $\{W^Q_b, W^{K,\mathrm{alm}}_b, W^{V,\mathrm{alm}}_b, W^{\mathrm{mix}}_b, \gamma_b\}$: the query reads the block's atom features and the keys and values are linear projections of the timestep-fused conditioning $\widetilde{\mathbf{C}}(t)$ (the producer output $\mathbf{C}$ fused with the noise level; Section~\ref{app:gen_tenc}), with the mixin $W^{\mathrm{mix}}_b\in\mathbb{R}^{d_h\times d_h}$ projecting the read-out back into the GemNet stream. $\gamma_b\in\mathbb{R}$ is a learnable per-block scale (init $1.0$); $g\in\mathbb{R}_{\ge 0}$ is the classifier-free guidance scale applied at sampler time. Two ControlNet-style zero-initializations~\cite{zhang_controlnet_2023} make the bridge a no-op at training step zero: $W^{\mathrm{mix}}_b\equiv\mathbf{0}$ and the final layer of the timestep-fusion MLP (Section~\ref{app:gen_tenc}) is zero-initialized in weight and bias. 

\paragraph{Producer: cross-modal block stack.}
\label{app:gen_producer}

The producer maps the source sequence $\mathbf{S}\in\mathbb{R}^{(N+K)\times d_{\mathrm{LM}}}$ --- the $N{=}128$ context states followed by the $K{=}8$ atomistic-token states (extracted as in Appendix~\ref{app:diffusion_adapter}) --- to the conditioning sequence $\mathbf{C}=f_{\mathrm{QF}}(\mathbf{Q}_{\mathrm{LQ}};\mathbf{S}) \in\mathbb{R}^{M\times d_{\mathrm{cond}}}$ ($M{=}16$, $d_{\mathrm{cond}}{=}512$), inspired by Q-Former~\cite{li_blip2_2023}. Expanded, $f_{\mathrm{QF}}$ is a stack of $L_{\mathrm{QF}}{=}2$ transformer blocks ($8$ attention heads, $M{=}16$ learned queries), indexed by $j$ (the consumer block index $b$ is reserved for the GemNet stack below):
\begin{align}
\mathbf{Q}^{(0)} &= \mathbf{Q}_{\mathrm{LQ}},
  \qquad \mathbf{S}^{\mathrm{cond}} = W^{\mathrm{down}} \mathbf{S} + \mathbf{E}_{\mathrm{type}}
  \quad (W^{\mathrm{down}} \in \mathbb{R}^{d_{\mathrm{cond}} \times d_{\mathrm{LM}}}),
  \label{eq:app_gen_z_proj}
\\
\widetilde{\mathbf{Q}}^{(j)} &= \mathbf{Q}^{(j)} + \mathrm{MHA}\big(
  \mathbf{Q}^{(j)},\, \mathbf{Q}^{(j)},\, \mathbf{Q}^{(j)} \big),
  \quad j = 0,\dots,L_{\mathrm{QF}}{-}1, \label{eq:app_gen_qf_self}
\\
\widehat{\mathbf{Q}}^{(j)} &= \widetilde{\mathbf{Q}}^{(j)} + \mathrm{MHA}\big(
  \widetilde{\mathbf{Q}}^{(j)},\, \mathbf{S}^{\mathrm{cond}},\,
  \mathbf{S}^{\mathrm{cond}} \big),
  \label{eq:app_gen_qf_cross}
\\
\mathbf{Q}^{(j+1)} &= \widehat{\mathbf{Q}}^{(j)} +
  \mathrm{FFN}\big(\widehat{\mathbf{Q}}^{(j)}\big),
  \label{eq:app_gen_qf_ffn}
\\
\mathbf{C} &= \mathrm{LayerNorm}\!\big(\mathbf{Q}^{(L_{\mathrm{QF}})}\big),
  \label{eq:app_gen_qf_out}
\end{align}
with $\mathrm{MHA}(\mathbf{Q},\mathbf{K},\mathbf{V}) = \mathrm{Softmax}(\mathbf{Q}W^Q (\mathbf{K}W^K)^\top/\sqrt{d})\,\mathbf{V}W^V$. Equation~\eqref{eq:app_gen_z_proj} is the only place dimensionality changes from $d_{\mathrm{LM}}{=}4096$ to $d_{\mathrm{cond}}{=}512$; the rest of the Q-Former operates in the diffusion decoder's native conditioning space. The additive type embedding $\mathbf{E}_{\mathrm{type}}$ (one learned vector for the $N$ context rows, another for the $K$ atomistic rows), $W^{\mathrm{down}}$, the per-block attention and FFN weights, and the learnable queries $\mathbf{Q}_{\mathrm{LQ}}$ are the only trainable parameters in the producer. This Q-Former-style encoder generalizes a Perceiver-style resampler~\cite{jaegle_perceiver_2021}: it compresses the variable-content $(N{+}K)$-token source into a fixed-length $M$-token conditioning set decoupled from the source length.

\paragraph{Timestep-aware conditioning fusion.}
\label{app:gen_tenc}

The producer output $\mathbf{C}$ is timestep-independent, but the diffusion trajectory passes through wildly different noise regimes governed by $\sigma(t)$ (Appendix~\ref{app:gen_diff}). We fuse a noise-level encoding $\boldsymbol{\tau}(t) \in \mathbb{R}^{d_{\mathrm{cond}}}$ (the same NoiseLevelEncoding used by MatterGen) into $\mathbf{C}$ via a zero-initialized residual MLP:
\begin{equation}
\widetilde{\mathbf{C}}(t) \;=\; \mathrm{LayerNorm}\Big(
  \mathbf{C} \,+\, \mathrm{MLP}\big([\mathbf{C}\,;\,\boldsymbol{\tau}(t)]\big)
\Big),
\label{eq:app_gen_tenc_fuse}
\end{equation}
whose final linear is zero-initialized, so at training step zero $\widetilde{\mathbf{C}}(t) = \mathrm{LayerNorm}(\mathbf{C})$ and the MLP is the only path through which the bridge becomes noise-aware. $\widetilde{\mathbf{C}}(t)$ replaces $\mathbf{C}$ in Eq.~\eqref{eq:app_gen_attn_alm}.

\paragraph{Consumer: cross-attention injection.}
\label{app:gen_consumer}

The bridge amplifies the LLM's conditioning signal with the Q-Former producer and injects it, at every denoising block, through a single decoupled cross-attention head. Let $\Psi_b(\mathbf{h}_b)\in\mathbb{R}^{N_p\times d_h}$ be the per-atom hidden state at the output of GemNet-T consumer block $b\in\{1,\dots,L_{\mathrm{D}}\}$'s message passing ($N_p$ atoms per cell, $d_h{=}512$). Expanding Section~\ref{sec:generator}'s boxed Eq.~\eqref{eq:gen_short_combined}, the injected read-out and the resulting block update are:
\begin{align}
\Delta\mathbf{h}_b^{\mathrm{alm}}
  &= W^{\mathrm{mix}}_b \cdot
     \mathrm{Softmax}\!\big(
        \Psi_b(\mathbf{h}_b)\, W^Q_b\,
        (\widetilde{\mathbf{C}}(t) W^{K,\mathrm{alm}}_b)^\top
        /\sqrt{d_h}\big)\,
     \widetilde{\mathbf{C}}(t) W^{V,\mathrm{alm}}_b,
  \label{eq:app_gen_attn_alm}
\\
\mathbf{h}_b &\leftarrow
\Psi_b(\mathbf{h}_b) \,+\, \gamma_b \cdot \Delta\mathbf{h}_b^{\mathrm{alm}},
  \label{eq:app_gen_combined}
\end{align}
where $\Psi_b$ is the frozen GemNet block (message passing plus MatterGen's additive native-property conditioning) and $\widetilde{\mathbf{C}}(t)$ is the timestep-fused conditioning of Appendix~\ref{app:gen_tenc}. The bridge introduces a single cross-attention head --- the only cross-attention in the score network --- whose query reads the block's neighborhood-aggregated atom features and whose keys and values are independent linear projections of the $M{=}16$ producer tokens $\widetilde{\mathbf{C}}(t)$; the mixin $W^{\mathrm{mix}}_b\in\mathbb{R}^{d_h\times d_h}$ projects the read-out back into the GemNet stream. Native property conditioning is \emph{not} a parallel cross-attention --- it is folded into $\Psi_b$ via the concatenated per-crystal latent --- so the score $s_\theta$ of Appendix~\ref{app:gen_diff} depends on the producer output $\mathbf{C}$ \emph{only} through this single gated branch. The per-block gate $\gamma_b$ is learned; classifier-free guidance $g$ is applied at sampler time through the CFG extrapolation of Eq.~\eqref{eq:app_gen_cfg}. The entire language-to-atomistic bridge has roughly 19M parameters. 

Equation~\eqref{eq:app_gen_combined} shares MS-Diffusion's decoupled-K/V cross-attention topology~\cite{wang_msdiffusion_2024} but differs in three deliberate ways --- the source signal is LoRA-adapted Qwen3-8B hidden states (not CLIP image features), the queries $\mathbf{Q}_{\mathrm{LQ}}$ are purely learnable (not grounding-token-initialized), and the consumer keys/values derive only from $\widetilde{\mathbf{C}}(t)$ (not concatenated with the text stream). In addition, unlike canonical IP-Adapter~\cite{ye_ip-adapter_2023}, it keeps random K/V initialization, achieving zero contribution at the beginning of training from zero-initialized $W^{\mathrm{mix}}_b$ and timestep-fusion MLP.

\subsubsection{Language model backbone finetuning for steering generation}
\label{app:lora_config}

\paragraph{LoRA and Full finetuning ablations.} Full finetuning was necessary for the language model $\phi$ (Qwen3-8B) to produce rich enough atomistic token embeddings to steer generation in \textbf{ALM Edit}. LoRA, although at different ranks and $\alpha$ values, sufficed for \textbf{ALM Core} and \textbf{Gen}. We present systematic ablations to validate these choices, as well as to explain why only \textbf{Edit} performs well on the LLM-judged materials-knowledge retention task in \textbf{ALM Bench}. 

We found that different LoRA parameters worked best for \textbf{ALM Core} versus \textbf{Gen}. \textbf{Core} uses rank $r{=}128$, $\alpha{=}256$, dropout $0.05$, at effective batch size $256$, with a sweep shown in Table~\ref{tab:s2-lora-rank} over $r$.
\begin{table}[ht]
\centering
\caption{\textbf{LoRA rank ablation on the LLM4Mat-Bench MP slice.}}
\label{tab:s2-lora-rank}
\small
\begin{tabular}{lrr}
\toprule
Config & LLM4Mat MP MAD/MAE & Leak Rate \\
\midrule
$r{=}8$            & $2.5$ & $\mathbf{<1\%}$ \\
$r{=}32$           & $4.0$ & $\mathbf{<1\%}$ \\
$r{=}64$           & $5.23$     & $\mathbf{<1\%}$ \\
\textbf{$r{=}128$} & $\mathbf{6.42}$ & $\mathbf{<1\%}$ \\
\bottomrule
\end{tabular}
\end{table}

\begin{figure}[ht]
  \centering
  \includegraphics{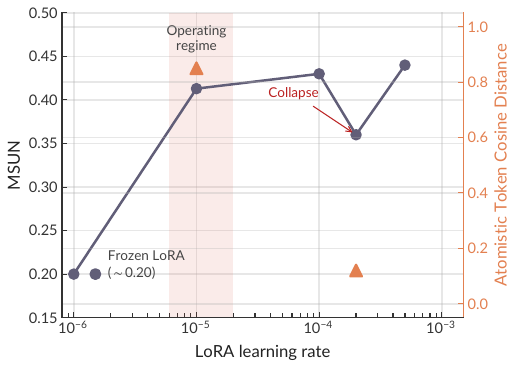}
  \caption{\textbf{generation LoRA learning-rate sweep.} MSUN as a function of fresh-rank-8 LoRA learning rate (log scale). $\text{lr}{=}0$ leaves the atomistic tokens out-of-vocabulary; $\text{lr}{=}2\text{e-}4$ collapses cross-prompt cosine distance to $0.12$ (triangles on the right axis). $\text{lr}{=}1\text{e-}5$ is the selected setting.}
  \label{fig:lora_lr_sweep}
\end{figure}

To train \textbf{ALM Gen}, the 224 LoRA matrices from training \textbf{Core} were merged into the base weights at the start of finetuning, and a new LoRA was attached at $r{=}8$, $\alpha{=}16$, and a learning rate of $1\text{e-}5$, decoupling generation adaptation from \textbf{Core} materials understanding. Raising learning rate to $2\text{e-}4$ collapses the cross-prompt hidden-state geometry (Fig.~\ref{fig:lora_lr_sweep}). Counter to the intuition that more adapter capacity helps, the final configuration outperforms an unmerged understanding LoRA at $r{=}64$, $\alpha{=}128$, which regressed the mixed de-novo MSUN to $0.094$. The inductive bias of the merged understanding LoRA already carries the LM-side adaptation, so the small fresh LoRA only has to translate the $K{=}8$ atomistic-token hidden states into useful conditioning.

An ablation over full- versus LoRA-finetuning \textbf{ALM Core} to develop \textbf{Gen} is shown in Table~\ref{tab:fullft-vs-lora}. Full finetuning achieves better metrics across all measured evaluations than LoRA $r=8$. Crucially, it also prevents the degradation of instruction following observed with LoRA, in which the model starts to output IMGUR links or loop continuously, described further in Appendix~\ref{app:training_stage2}. 

\begin{table}[ht]
\centering
\caption{\textbf{Full finetuning versus LoRA on \almEdit{}.} Full finetuning is done with PEFT. Retention judge is the LLM-judge score ($0$--$2$).}
\label{tab:fullft-vs-lora}
\small
\begin{tabular}{lccc}
\toprule
Metric & LoRA $r{=}8$ & Full Finetuning \\
\midrule
Retention judge ($0$--$2$)      & $1.053$ & $\mathbf{1.842}$ \\
\quad loop-rate                 & $0.316$ & $\mathbf{0.0}$ \\
\quad keyword-pass              & $0.462$ & $\mathbf{0.923}$ \\
App-consistency                 & $0.0275$ & $\mathbf{0.326}$ \\
Direction-correct               & $0.608$ & $\mathbf{0.621}$ \\
CSP Match@1                     & $0.428$ & $\mathbf{0.472}$ \\
CSP Match@$K=64$               & $0.881$ & $\mathbf{0.896}$ \\
\bottomrule
\end{tabular}
\end{table}

As a result of the further training on instruction tuning, \textbf{ALM Edit} and \textbf{Gen} are better at predicting certain properties than \textbf{ALM Core}. The URL leak and loop rates are also near-zero for \textbf{ALM Gen}, as shown in Table~\ref{tab:gen_vs_edit_understand}.

\begin{table*}[ht]
\centering
\caption{\textbf{ALM Variants evaluated on atomistic understanding and materials science knowledge tasks}.}
\label{tab:gen_vs_edit_understand}
\footnotesize
\setlength{\tabcolsep}{4pt}
\resizebox{\textwidth}{!}{%
\begin{tabular}{l rrrr rr rrrr}
\toprule
 & \multicolumn{4}{c}{LLM4Mat MAD:MAE $\uparrow$} & \multicolumn{2}{c}{Mat2Props MAE $\downarrow$} & \multicolumn{4}{c}{Accuracy $\uparrow$} \\
\cmidrule(lr){2-5}\cmidrule(lr){6-7}\cmidrule(lr){8-11}
Model & MP $E_f$ & MP gap & MP $\rho$ & OQMD $E_f$ & $E_f$ & gap & MaScQA & GPQA-chem & Mat2MCQ & GSM8K \\
\midrule
\rowcolor{almCoreBg}{ALM Core} & $14.12$ & $3.88$ & $10.10$ & $13.23$ & $0.087$ & $0.275$ & $0.643$ & $0.247$ & $0.417$ & $0.775$ \\
\rowcolor{almGenBg}ALM Gen & $3.94$ & $2.24$ & $3.07$ & $0.64$ & $0.397$ & $0.584$ & $0.381$ & ${0.333}$ & ${0.564}$ & $0.720$ \\
\rowcolor{almEditBg}ALM Edit & ${15.79}$ & ${4.37}$ & ${12.33}$ & ${15.65}$ & ${0.074}$ & ${0.244}$ & ${1.000}$ & $0.228$ & $0.508$ & ${0.790}$ \\
\bottomrule
\end{tabular}}
\end{table*}

%%% ===== frag_I_D_iii_2.tex =====
\subsubsection{Architectural ablations over generative variants}
\label{app:diffusion_adapter}

\begin{figure}[ht]
  \centering
  \includegraphics[width=\linewidth]{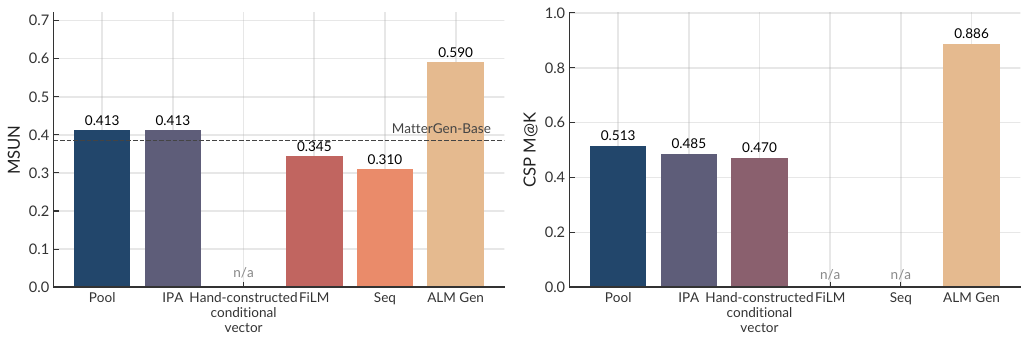}
  \caption{\textbf{Alternate bridge architectures on DNG SUN and CSP M@K.} Left: MSUN ($E_{\text{hull}} < 0.1$) at each bridge's optimal $g$. Right: CSP M@K at $K=64$ on the CSP-mode backbone. $N=500$ rows were drawn for each task from the test set, explaining the higher-than-reported CSP and MSUN than ALMs achieve in the main text.}
  \label{fig:bridge_comparison}
\end{figure}

\paragraph{Language-to-atomistic bridge architecture ablations.} Several ablations over simpler architectures of the language-to-atomistic bridge were conducted, validating the final design, as shown in Fig.~\ref{fig:bridge_comparison}. These architectures were a simple MLP across the $K=8$ token embeddings concatenated together; IP-Adapter~\cite{ye_ip-adapter_2023}; a hand-constructed guidance vector consisting of a multi-hot vector to encode composition and additional bits to account for directional tasks in \textbf{ALM Bench} (e.g., 1 for raise, 0 for lower); FiLM~\cite{perez2018film}; and lastly, one MLP per atomistic token, preserving the \textit{Seq}uence information without flattening all of the embeddings upon input. 

Several of the architectures did not produce any valid structures for certain tasks, outputting generations that exploded upon relaxation or degenerate crystals with a single atom. However, no architecture performed as well as the consumer--producer bridge formulation used to build \textbf{ALM Edit} and \textbf{Gen}. One reason for this is that several of these architectures don't support growing the bridge contribution from a literal zero at step~0 (the ControlNet-style zero-init of Appendix~\ref{app:gen_consumer},~\cite{zhang_controlnet_2023}). In practice, this stabilizes training across a wide range of CFG guidance scale $g$ values. FiLM, Seq, and the hand-constructed conditional vector all did not support this training dynamic; as an example, FiLM-style feature-wise linear modulation~\cite{perez2018film} is multiplicative ($\gamma(\mathbf{C})\odot\mathbf{h}_b + \beta(\mathbf{C})$) and initializes $\gamma\!\approx\!1$, so it already perturbs every block at step~0, never establishes a conditioning gradient, and every learning-rate, warmup, and data-mix intervention drives it further toward a no-op. Seq also crashes under strong conditioning, e.g. producing degenerate cells on $20/20$ when prompted for perovskites.

\begin{figure}[ht]
  \centering
  \includegraphics[width=0.75\textwidth]{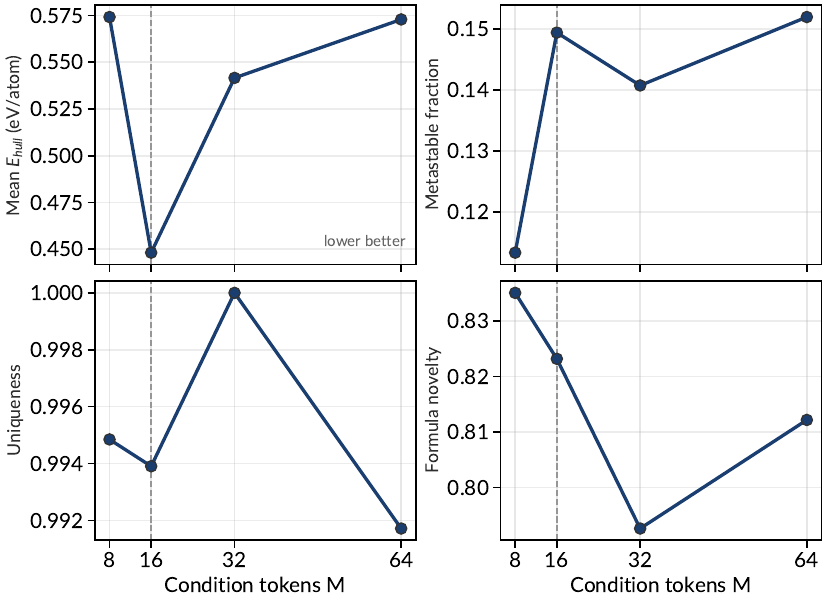}
  \caption{\textbf{De-novo generative metrics across condition-token count $M\in\{8,16,32,64\}$.}}
  \label{fig:dng_m_sweep}
\end{figure}

\paragraph{Fixed-length conditioning output from the producer.} The producer emits a length-$M$ condition sequence $\mathbf{C}\in\mathbb{R}^{M\times d_{\mathrm{cond}}}$. $M$ therefore sets the bandwidth of the signal the consumer can cross-attend to at each denoising step. The effect of sweeping over values of $M$ on \textit{de novo} generation metrics is shown in Fig.~\ref{fig:dng_m_sweep}. $M=16$ was chosen as the operating point due to strong performance across each metric. 

\begin{figure}[ht]
\centering
\includegraphics{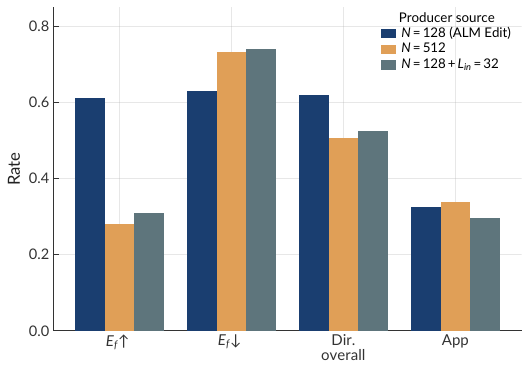}
\caption{\textbf{Q-Former producer source-length (context-window) ablation} ($g{=}0.5$, honest denominator over all $768$ atomtxt attempts; rest of the final recipe held fixed). \emph{Both} ways of feeding the producer more source --- widening the window ($N{=}512$) or prepending an explicit $L_{\mathrm{in}}{=}32$ input-\texttt{<atoms>} segment --- collapse raise-$E_f$ direction-correctness below chance (the dashed line at $0.5$) and lift lower-$E_f$, landing at overall direction-correct $\sim\!0.51$--$0.53$ versus the final $N{=}128$ recipe's symmetric $0.62$. App-consistency is roughly unchanged across all three. The small directional residual the consumer reads is diluted by any extra source content, but coarse text$\to$structure conditioning is not.}
\label{fig:qformer_window}
\end{figure}

\begin{figure}[h]
  \centering
  \includegraphics[width=0.62\linewidth]{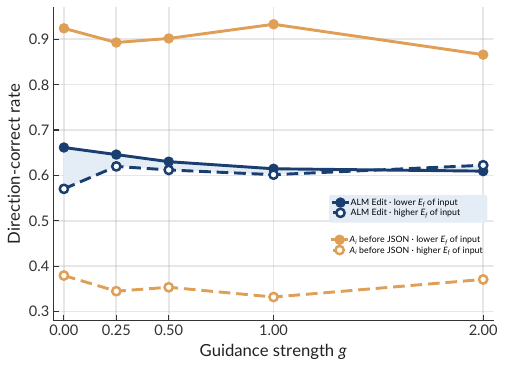}
  \caption{\textbf{Output-token ordering determines whether ALM Edit follows directional instructions.} Direction-correct rate (fraction of generations that moved formation energy $E_f$ the requested way relative to the input) versus\ classifier-free-guidance strength $g$, for two output-token orderings: the \textbf{ALM Edit} ordering (composition JSON before the $A_i$ atom tokens, blue, highlighted) and an ablation that teacher-forces the $A_i$ tokens before the composition ($A_i$ before JSON, orange).}
  \label{fig:steering_gsweep}
\end{figure}

\paragraph{Cross-attention context window for the producer.} The second producer-bandwidth hyperparameter is the length $S$, the total number of token embeddings cross-attended to by the $M{=}16$ queries. Here, $S=N+K$ is chosen such that $N{=}128$ recent LLM context states and $K{=}8$ atomistic token states. Widening $N{=}128$ to $512$ does not help directional editing, although it marginally improves the rate at which \textbf{ALM Edit} generates structures that obey the prompted application area, as shown in Figure~\ref{fig:qformer_window}. We observe the collapse of direction following when the $E_f\uparrow$ rate decreases, as the model is trained on several tasks to output polymorphs with $E_{\text{h}}$ lower than the inputted material, and thus is regressing to the distribution of structures marginalized over stability. This collapse is also observed when removing the composition auxiliary head, as well as swapping the order of the atomistic token teacher forcing to before the composition JSON is outputted (Fig.~\ref{fig:steering_gsweep}).

\begin{figure}[ht]
  \centering
  \includegraphics[width=\linewidth]{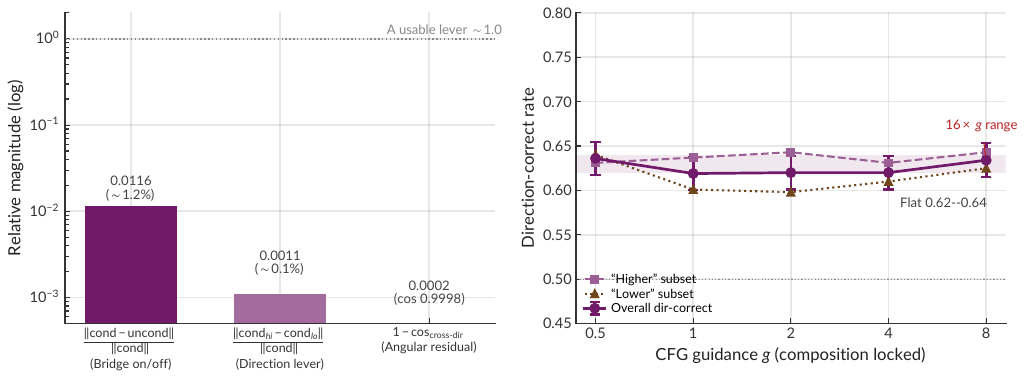}
  \caption{\textbf{Unconditional and conditional outputs are similar for ALM Edit.} \emph{Left:} relative magnitudes (log scale) of setting $g$ to be $0.5$ (conditional) or $0$ (unconditional). \emph{Right:} direction-correctness is flat at $0.62$--$0.64$ over a $16\times$ range of $g$, indicating that the ceiling is a representation-quality limit, not a magnitude limit.}
  \label{fig:cosine_vs_metric}
\end{figure}

We also find that the conditional signal in \textbf{ALM Edit} is small (Fig.~\ref{fig:cosine_vs_metric}), learning a conditional head through CFG that is very close in cosine distance to the unconditional head output. However, the signal does produce slight $g$-dependence in the overall performance on directional tasks in \textbf{ALM Bench}.

%%% ===== frag_I_E.tex =====
\subsection{Text-to-Crystal Feynman-Kac algorithm details}
\label{app:fk_details}

Text-to-Crystal Feynman-Kac steering (T2C-FK; Section~\ref{sec:methods_steering}) is an inference-time mechanism that makes \almGen{} generate crystals with a requested element set and stoichiometry. \almGen{} is a strong but deliberately weakly-conditioned de-novo sampler: its backbone produces stable cells, but the language prompt only biases the composition towards CFG-provided element counts. \textbf{T2C-FK} closes that gap by wrapping the unconditional reverse diffusion in an $N$-particle sequential Monte Carlo (SMC) sampler that reweights toward a stoichiometry reward, with no retraining or change to the score network. The same machinery ports to \almEdit{} for composition-exact CSP decoding and for directional editing; those uses are collected as a side application in Appendix~\ref{app:fk_alm_edit}. All diffusion notation follows the glossary of Appendix~\ref{app:gen_notation}.

\subsubsection{Bootstrap-SMC sampler}
\label{app:fk_smc_derivation}

T2C-FK replaces MatterGen's single denoising trajectory with an $N$-particle bootstrap SMC sampler over the reverse diffusion~\cite{wu_practical_2023, singhal_general_2025}. All $N$ particles are propagated in lockstep by the unconditional predictor-corrector step (shared score network $s_\theta$, condition $\mathbf{C}$); every $S$ steps the population is reweighted by the reward on the Tweedie clean estimate $\hat{x}_0$ and resampled when its effective sample size falls below $\rho N$. The per-particle log-weight updates as
\begin{equation}
\ell^{(i)}_t \;\gets\; \mathrm{clip}\!\big(\ell^{(i)}_t + \lambda\, r(\hat{x}_0(x_t^{(i)},t)),\ \pm L_{\mathrm{clip}}\big),
\end{equation}
where $\hat{x}_0(x_t^{(i)},t)$ is the Tweedie estimator and $\lambda$ the steering scale. Because FK reweights and resamples whole particles outside the score evaluation, it composes additively with the language-to-atomistic bridge architecture through CFG. In addition, the absorbing-state D3PM over $\mathbf{A}_t$ is MASK-dominated for $t > T/2$, where the reward on $\hat{x}_0$ is uninformative, so scoring begins only once $t < T(1-\tau_{\mathrm{start}})$; with $\tau_{\mathrm{start}}{=}0.5$ this halves the scoring compute at no measurable cost in match rate. Algorithm~\ref{alg:t2c_fk} gives the full procedure.

\begin{algorithm}[ht]
\caption{T2C-FK: Feynman-Kac steered structure generation.}
\label{alg:t2c_fk}
\begin{algorithmic}[1]
\Require text prompt $u$, target multiset $\mathcal{T}$, particles $N$, steps $T$, scoring period $S$, deferred start $\tau_{\mathrm{start}}$, scale $\lambda$, ESS threshold $\rho$, clip $L_{\mathrm{clip}}$.
\State $\mathbf{S} \gets$ context $+$ $[\mathtt{atoms\_*}]$ hidden states from $\phi(\mathrm{tokenize}(u))$;\; $\mathbf{C} \gets f_{\mathrm{QF}}(\mathbf{Q}_{\mathrm{LQ}};\mathbf{S})$.
\State $\{x_T^{(i)}\}_{i=1}^N \sim p_{\mathrm{prior}}$;\; $\ell^{(i)} \gets 0$.
\For{$t = T-1, \dots, 0$}
  \State $\{x_t^{(i)}\} \gets \mathrm{PCStep}(\{x_{t+1}^{(i)}\}, s_\theta, \mathbf{C})$.
  \If{$t < T(1 - \tau_{\mathrm{start}})$}
    \State $\hat{x}_0^{(i)} \gets \mathrm{Tweedie}(x_t^{(i)}, t)$;\; $\ell^{(i)} \gets \mathrm{clip}(\ell^{(i)} + \lambda r(\hat{x}_0^{(i)}), \pm L_{\mathrm{clip}})$.
    \If{$t \bmod S = 0$ \textbf{and} $\mathrm{ESS}(\ell) < \rho N$}
      \State $\{x_t^{(i)}\} \gets \mathrm{Multinomial}(\{x_t^{(i)}\}, \mathrm{softmax}(\ell), N)$;\; $\ell^{(i)} \gets 0$.
    \EndIf
  \EndIf
\EndFor
\State Apply Hungarian $Z$-override on $\{x_0^{(i)}\}_{i=1}^N$.
\State \Return $\{x_0^{(i)}\}_{i=1}^N$.
\end{algorithmic}
\end{algorithm}

\label{app:fk_smc_formal}
\paragraph{Posterior-correction guarantee.} \textbf{T2C-FK} accumulates the Feynman-Kac weight with the \texttt{sum} rule, $G_t^{(i)} = \exp\!\big(\lambda\, r(\hat{x}_0(x_t^{(i)},t))\big)$ (alternate accumulation rules are ablated in Appendix~\ref{app:fk_hyperparams}). Let $p^*(x_0) \propto p(x_0)\exp(r(x_0)/\tau)$ be the target posterior at $t{=}0$, with $p(x_0)$ the unconditional MatterGen marginal and $\tau$ an effective temperature set by $\lambda$. Bootstrap-SMC with proposal $p$ and potential $G_t$ recovers $p^*$ as $N \to \infty$ under bounded reward and standard SMC regularity~\cite{wu_practical_2023, singhal_general_2025}; the three conditions hold here: the reward is bounded above (after clipping), the proposal is the unconditional MatterGen sampler at every step (FK never alters $s_\theta$), and multinomial resampling on \texttt{softmax}-normalized log-weights is a valid SMC move. Clipping at $\pm L_{\mathrm{clip}}$ introduces a controlled bias ($<\!5\%$ clip-saturation at $\varepsilon{=}10^{-4}$).

\subsubsection{Stoichiometry reward and Hungarian \texorpdfstring{$Z$}{Z}-override}
\label{app:reward_definitions}\label{app:smear_eq}

The reward is needed because providing element counts through CFG effectively enforces the element set, but MatterGen's score network loosens the stoichiometric ratios during denoising. With the auxiliary composition loss, decoding the atomistic-token hidden states $\mathbf{Z}$ through the generation auxiliary head recovers the target composition at $>\!95\%$ top-1, yet the unsteered diffusion trajectory drifts off the exact elemental ratio. The per-particle reward scores that ratio as a sum of three components,
\begin{equation}
r(\hat{x}_0) = r_{\mathrm{stoich}}(\hat{x}_0) + r_{\mathrm{count\_L1}}(\hat{x}_0) + r_{\mathrm{ratio\_JS}}(\hat{x}_0).
\end{equation}
For predicted per-atom element distributions $\{\mathbf{p}_a\}_{a=1}^{N_p}$ ($N_p$ atoms in the cell) and target multiset $\mathcal{T}$:
\begin{align}
r_{\mathrm{stoich}} &= -\frac{1}{N_p}\sum_a -\log\!\big(\mathbf{p}_a[z_{\pi^*(a)}] + \varepsilon\big), \label{eq:fk_r_stoich}\\
r_{\mathrm{count\_L1}} &= -\frac{1}{|\mathcal{S}|} \sum_{e \in \mathcal{S}} \big|\, n_e^{\mathrm{argmax}} - n_e^{\mathrm{target}} \,\big|, \\
r_{\mathrm{ratio\_JS}} &= -\mathrm{JS}\!\left(\hat{\mathbf{q}}^{\mathrm{argmax}} \,\big\|\, \mathbf{q}^{\mathrm{target}}\right),
\end{align}
where $\pi^*$ is the Hungarian (linear-sum) assignment of atoms to target elements, $\mathcal{S}$ the set of distinct target elements, $n_e^{\mathrm{argmax}}$ the count of element $e$ under per-atom argmax, and $\mathrm{JS}$ the symmetric, $\log 2$-normalized Jensen-Shannon divergence. $r_{\mathrm{stoich}}$ is the negated mean Hungarian-assigned per-atom NLL (cost $C_{aj}{=}-\log(\mathbf{p}_a[z_j]+\varepsilon)$, solved at $\sim\!10\,\mu$s per particle, capped at $\log(1/\varepsilon)\approx9.2$); $r_{\mathrm{count\_L1}}$ and $r_{\mathrm{ratio\_JS}}$ are each bounded in $[-1,0]$. The two hard-argmax components are necessary: a uniform soft distribution that averages to the target stoichiometry earns a perfect soft score even though no individual particle is valid, so $r_{\mathrm{count\_L1}}$ and $r_{\mathrm{ratio\_JS}}$ commit each atom to a single element \emph{before} scoring, breaking it.

\paragraph{Hungarian $Z$-override.} After the final denoising step, the linear-assignment problem is solved on the terminal atomic-number probabilities and each atom is set to its assigned element $z_{\pi^*(a)}$, leaving the lattice $\mathbf{L}$ and fractional coordinates $\mathbf{X}$ untouched. SMC has already concentrated mass on target-consistent particles, so this almost always agrees with the argmax it replaces; its role is to \emph{guarantee} the composition match on the residual.

The per-atom reward is informative only when $N_p$ is a multiple of the number of atoms in the requested material: otherwise, every Hungarian assignment is forced to place wrong-element atoms, the soft NLL of Eq.~\ref{eq:fk_r_stoich} is fixed at its $\log(1/\varepsilon)$ cap, and the cumulative log-weight saturates $L_{\mathrm{clip}}$ on every particle, causing SMC to degenerate to uniform resampling. \textbf{T2C-FK} therefore makes the number of particles $N_p$ a multiple of the total number of atoms. This fixed aspect of particles is what makes the \texttt{sum} potential well-behaved: with the right atom count, each per-atom NLL stays bounded away from its cap, so every step contributes a finite, discriminative increment.

\subsubsection{Additional hyperparameter sweeps and ablations}
\label{app:fk_hyperparams}

\almGen{}'s T2C-FK configuration is $N{=}8$, $T{=}1000$, $S{=}10$, $\tau_{\mathrm{start}}{=}0.5$, $\lambda{=}0.5$, $\rho{=}0.5$, $L_{\mathrm{clip}}{=}50$, $\varepsilon{=}10^{-4}$, the \texttt{sum} potential, and the equal-weight three-component reward over the last half of a $T{=}1000$ trajectory. The DNG-FK row of Table~\ref{tab:dng_strict_mp20} uses these settings. A value of $\lambda{=}0.5$ is selected for \almGen{} stoichiometry and $\lambda{=}3$ for directional editing (Appendix~\ref{app:fk_alm_edit}), where the stronger signal is needed to overcome MatterSim's relaxation bias toward lower energy. \textbf{T2C-FK} with $8$ particles costs $\sim\!8.1\times$ unsteered MatterGen sampling. As for reward components: $r_{\mathrm{stoich}}$ gives a calibrated continuous signal early in the second half of denoising while the element distributions are still soft; $r_{\mathrm{count\_L1}}$ supplies discrete-count alignment as the absorbing-state D3PM commits to specific elements; $r_{\mathrm{ratio\_JS}}$ is the bounded shape regularizer that keeps the SMC from collapsing onto a single dominant particle.

Tolerance $\varepsilon$ was swept over. Values of $\varepsilon{=}10^{-8}$ (per-atom NLL cap $\sim\!18.4$) saturated $L_{\mathrm{clip}}{=}50$ on every wrong-element atom and degenerated the SMC into uniform resampling because every particle's cumulative log-weight hit the clip at once. The final version for \textbf{T2C-FK} uses $\varepsilon{=}10^{-4}$ (cap $\sim\!9.2$), the smallest value keeping clip-saturation below $5\%$ in our diagnostic logs. Lastly, the \texttt{sum} rule (Appendix~\ref{app:fk_smc_formal}) accumulates $\lambda\,r(\hat{x}_0)$ each step. Other formulations of the rule, including a difference rule $G_t^{(i)} = \lambda\big(r(\hat{x}_0(x_t^{(i)},t)) - r(\hat{x}_0(x_{t+1}^{(i)},t{+}1))\big)$ do not reach the same performance as the sum formulation. 

\subsubsection{Porting T2C-FK to enable ALM Edit}
\label{app:fk_alm_edit}

The same T2C-FK serves \almEdit{} by allowing for rewards prescribed by \textbf{ALM Bench} directional editing tasks to steer structure generation during denoising. Specifically \textbf{ALM Bench} task direction is parsed from the prompt into the reward, a MatterSim energy reward (one MatterSim forward pass yielding potential energy $E$/atom). On \textbf{ALM Edit} at $g{=}0.5$, this formulation of FK pushed the \textbf{ALM Bench} direction-correctness rise to $0.719$. Inference-time FK reward-steering recovers both directions whenever the requested direction can be parsed from the prompt into an explicit reward.

%%% ===== frag_I_F.tex =====
\subsection{Scaling laws}
\label{app:scaling}

\begin{figure*}[t]
    \centering
    \includegraphics[width=\linewidth]{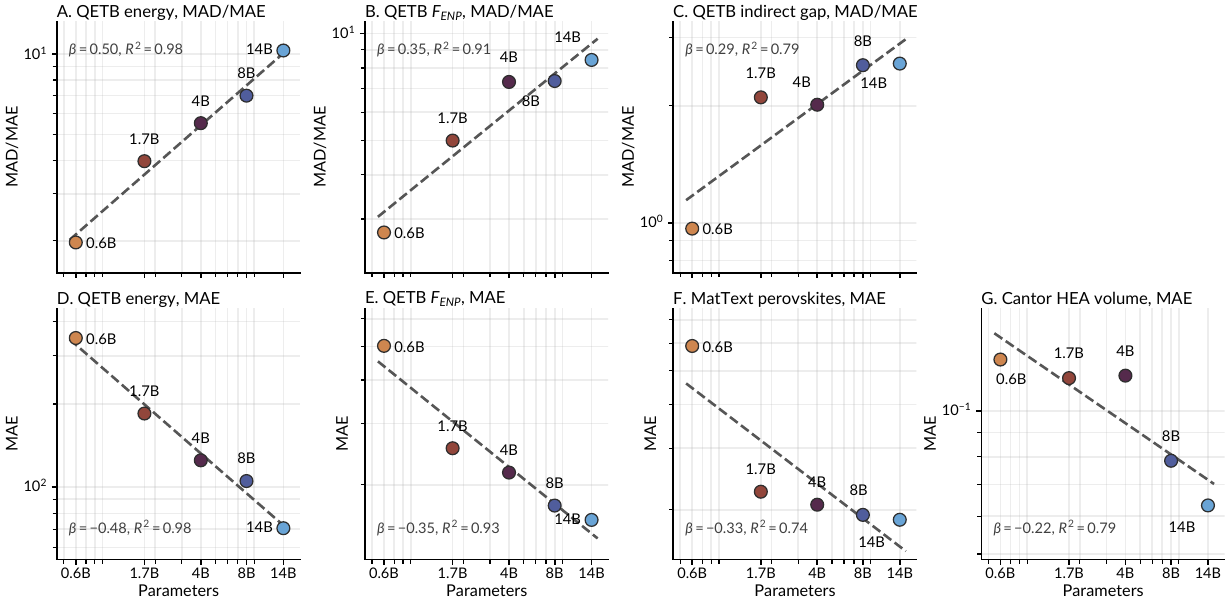}
\caption{\textbf{Model-size scaling across seven property-prediction metrics.} Each panel uses the available Qwen3 \textbf{ALM Core} evaluation suite for $0.6$B, $1.7$B, $4$B, $8$B, and $14$B-sized LMs. The top row reports MAD/MAE skill ratios on LLM4Mat-Bench~\cite{rubungo_llm4mat-bench_2024} JARVIS-QETB properties (higher is better). The bottom reports raw MAE on JARVIS-QETB, MatText, and Cantor-HEA properties (lower is better).}
    \label{fig:app_model_size_scaling}
\end{figure*}

Across several regression metrics, strong scaling laws emerge for property prediction tasks, as shown in Fig.~\ref{fig:app_model_size_scaling}; as the underlying language model, Qwen3, grows in parameter count, property prediction performance improves. The cleanest series is JARVIS-QETB energy, where both the MAD/MAE skill ratio and the raw MAE follow a near-linear trend in log--log space. The trend is not universal: indirect gap and MatText perovskites improve with scale but retain task-specific curvature and noise. Cantor-HEA shows emergent scaling, where additional size does not help until a sudden jump from 4B to 8B parameters. We hypothesize that further scaling of the ALM could produce similar emergent effects. 

%%% ===== frag_II_1.tex =====
\section{Training data}
\label{app:training}\label{sec:methods_training}

This appendix documents the data ALM is trained on, in the same order the model acquires its capabilities: first \emph{understanding} (structure~$\to$~text and structure$+$instruction~$\to$~text/value, Section~\ref{sec:methods_training} below), then \emph{generation and editing} (Section~\ref{app:training_stage3}). For each phase we give the exact bucket mixture, the per-bucket source datasets and sample counts, the pairing procedure, and the balance ablations that fix the mixture weights. Generation/editing buckets are deferred to Appendix~\ref{app:training_stage3}; the precise definitions of the evaluation metrics quoted here live in Appendix~\ref{app:stage2_metric_defs}.

\subsection{Data used to teach Atomistic Language Models to understand atoms}
\label{app:training_stage1}

\begin{figure}[ht]
  \centering
  \includegraphics{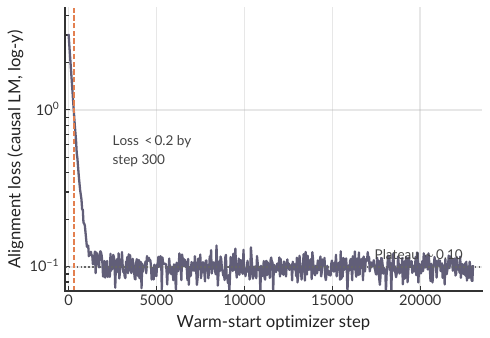}
  \caption{\textbf{Warm-start alignment loss trajectory ($K{=}8$).} Causal LM loss on $\sim\!1.35$M structure-description pairs drops below $0.2$ in $\sim\!300$ optimizer steps and plateaus near $0.10$.}
  \label{fig:stage1_loss}
  \end{figure}

ALM is taught to understand atoms in two phases. \emph{Alignment} fits the input-side projector $P_{\text{in}}$ so the frozen LM can attend to atomistic features at all; \emph{instruction tuning} then leverages this warm start to attach a LoRA adapter and finetune the language model on the full distribution of structure-conditioned tasks it must answer. Both phases keep the OrbV3 encoder frozen.

\subsubsection{Five-bucket training mixture}
\paragraph{Stage 1: Alignment data and optimizer.} Only $P_{\text{in}}$ is trained, under the standard causal language modeling loss on nearly $1.35$M structure--description pairs drawn from LLM4Mat-Bench~\cite{rubungo_llm4mat-bench_2024} and the four GPT-Narratives parquets (\texttt{dft\_3d}, \texttt{mp\_3d\_2020}, \texttt{aflow2}, \texttt{oqmd}). AdamW~\cite{loshchilov_decoupled_2019} is used at learning rate $1\mathrm{e}{-}3$, as well as weight decay $0$, betas $(0.9, 0.999)$, a $3\%$ cosine warmup capped at $2000$ steps then cosine decay, gradient clip $1.0$, and an effective batch of $32$. The alignment loss falls below $0.2$ within $\sim\!300$ steps and plateaus near $0.10$ across the full $23$k-step run (Figure~\ref{fig:stage1_loss}). The two-linear-with-GELU projector was chosen over a single-linear map (insufficient capacity) and a three-linear map (marginal gain at extra parameters); LLaVA-1.5~\cite{liu_improved_2024} uses the same two-linear projector at smaller scale. The encoder-adapter design space is analyzed in Appendix~\ref{app:codebook_comparison}.

\begin{figure}[ht]
  \centering
  \includegraphics[width=\linewidth]{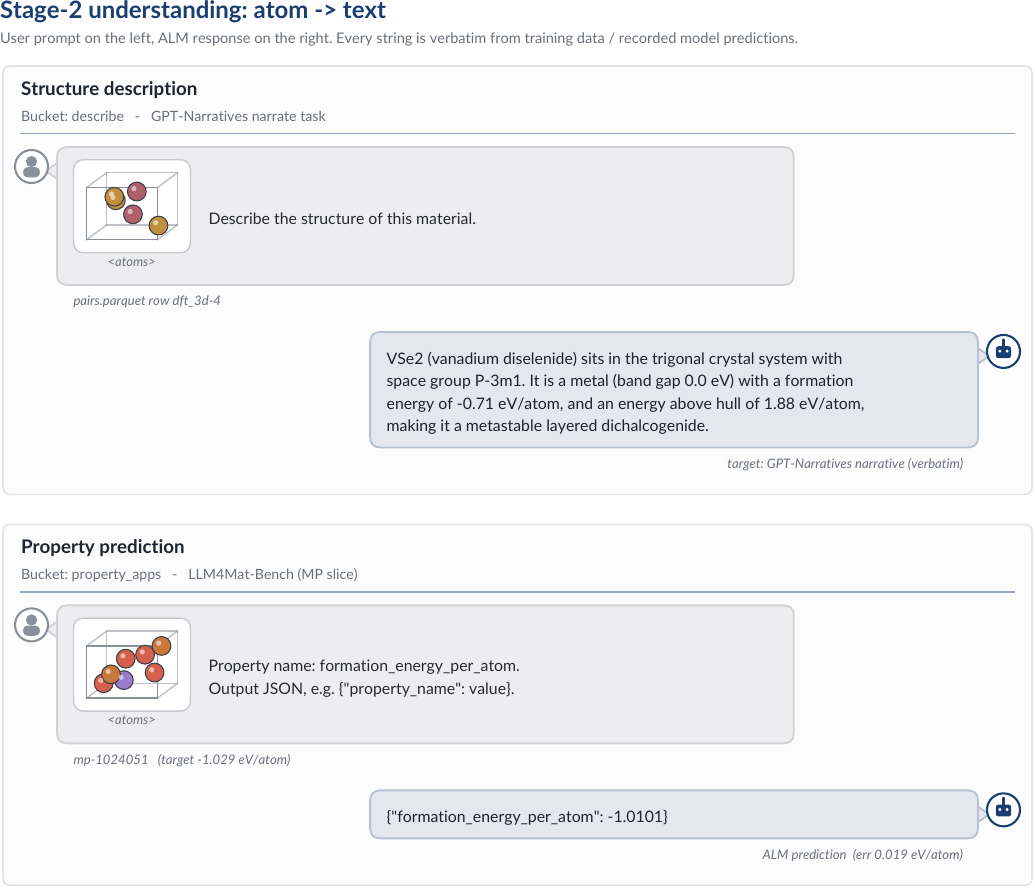}
  \caption{\textbf{Representative understanding-phase interactions}, one per bucket type (user prompt on the left, ALM response on the right): captioning (\texttt{describe}), property and applications VQA (\texttt{property\_apps}), and the text-only science Q\&A buckets (\texttt{arxiv}/\texttt{camel}/\texttt{mascqa}). The \texttt{<atoms>} placeholder is expanded inline to OrbV3 node features for the structure-conditioned turns.}
  \label{fig:chat_examples_understanding}
  \end{figure}

\begin{figure}[ht]
  \centering
  \includegraphics[width=\linewidth]{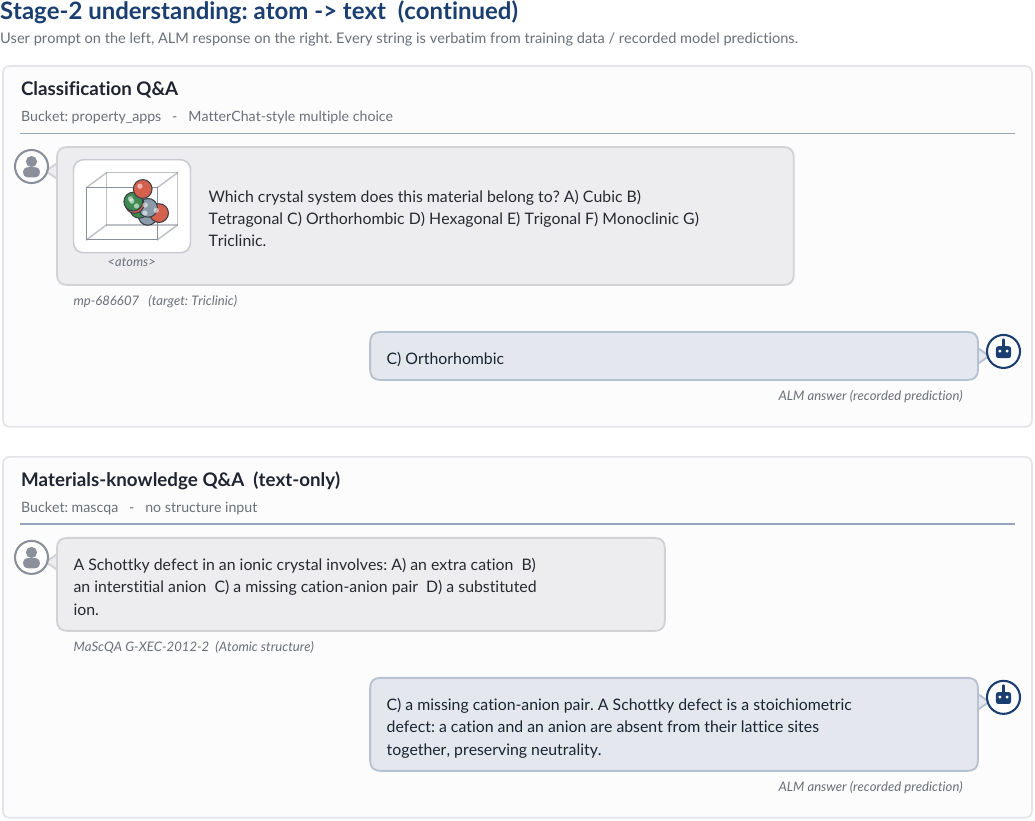}
  \caption{\textbf{Representative understanding-phase interactions (continued).}}
  \label{fig:chat_examples_understanding_cont}
\end{figure}

\paragraph{Stage 2: Instruction-tuning mixture.} With $P_{\text{in}}$ aligned, a LoRA adapter is attached to the LM, then instruction-tune on a fixed budget of $12$k optimizer steps (following the LLaVA-1.5 convention~\cite{liu_improved_2024} of total optimizer steps, \emph{not} epochs over any single bucket). Training draws from a \textbf{five-bucket} mixture (Table~\ref{tab:s2-buckets}) spanning two structure-conditioned tasks and three text-only tasks; the per-step bucket selection is a categorical draw with probability vector $\pi$. The two structural buckets (\texttt{describe}, consisting of tasks prompting the model to describe the structure of a material or to generate a free-form narrative about its device applications, \texttt{property\_apps}, consisting of property prediction tasks which request a JSON with only the given property as output) share the $\sim\!0.80$ structural budget at a $0.40/0.40$ split, while the \texttt{arxiv} (given a title and keywords of materials science preprints, generate a realistic abstract) supplies scientific fluency, and \texttt{camel}/\texttt{mascqa} (both scientific QA datasets) text buckets supply scientific fluency and Q\&A instruction following to prevent catastrophic forgetting. Over the full run, this mixes a total of $2.82$M training samples. Figures~\ref{fig:chat_examples_understanding} and~\ref{fig:chat_examples_understanding_cont} visualize representative rows of the dataset and how \textbf{ALM Core} responds compared to the ground truth labels. 

\begin{table*}[ht]
\centering
\caption{\textbf{Understanding-phase five-bucket mixture.} $\pi$ is the categorical probability used by the per-rank bucket sampler (Appendix~\ref{app:training_stage2})}
\label{tab:s2-buckets}
\resizebox{\textwidth}{!}{%
\begin{tabular}{lllrc}
\toprule
Bucket & Task type & Source & Rows & $\pi$ \\
\midrule
\texttt{describe} & captioning (structure $\to$ prose) & LLM4Mat \texttt{description} + GPT-Narratives & $\sim\!700$k & $0.40$ \\
\texttt{property\_apps} & VQA (structure $+$ instruction $\to$ value/text) & LLM4Mat property + GPT-Narratives EXPLAIN & $1.45$M & $0.40$ \\
\texttt{arxiv} & ChatML IT (title $+$ categories $\to$ abstract) & JARVIS arXiv abstracts & $375{,}571$ & $0.14$ \\
\texttt{camel} & text Q\&A & CAMEL-AI chemistry + physics role-plays & $37$k & $0.04$ \\
\texttt{mascqa} & MCQ benchmark & MaScQA ($80\%$ train split) & $519$ & $0.02$ \\
\bottomrule
\end{tabular}}
\end{table*}

\begin{figure}[ht]
  \centering
  \includegraphics[width=\linewidth]{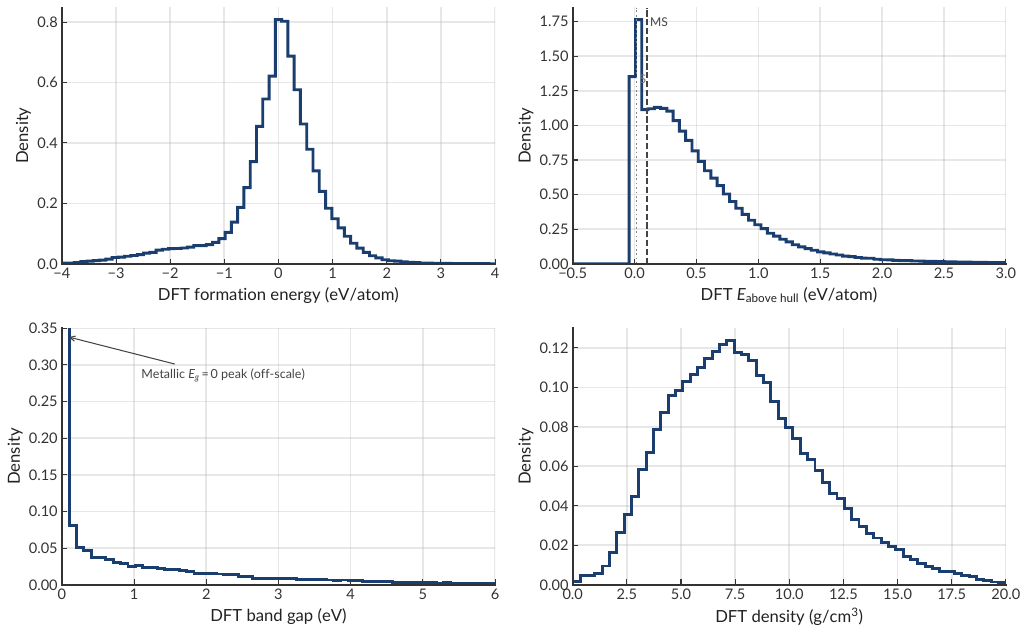}
  \caption{\textbf{Target-property distributions} over the training data (formation energy, energy above hull, band gap, density). These set the support of the values ALM is asked to read off and, in the editing phase, to move.}
  \label{fig:target_property_hist}
  \end{figure}

Figure~\ref{fig:target_property_hist} visualizes the distributions of properties for inputted materials across the data buckets.

During this stage of training, AdamW is used to optimize two groups of parameters with different learning rates: LoRA at learning rate $2\mathrm{e}{-}4$, projector at $2\mathrm{e}{-}5$, and both with weight decay $0.01$, betas $(0.9, 0.95)$, and a cosine schedule with warmup $\min(2000, 0.03\,T_{\max})$. The effective batch size is $256$. Validation buckets are partitioned with the same seed via \texttt{split\_seed=42}: arXiv and CAMEL hold out $500$ rows each, MaScQA holds out $20\%$ stratified by topic ($131/650$).

\subsubsection{The URL leak failure mode and mitigations}
\label{app:training_stage2}

\paragraph{The effect of batch size on URL leak rate.} The understanding phase (Stage 2 of training) requires an effective batch size of at least $64$. Below this threshold (Table~\ref{tab:s2-batchsize-app}) the LoRA receives too few gradient updates to suppress Qwen3-8B's base-language model web prior at the given LoRA learning rate, which re-surfaces on uncertain tasks: at an effective batch size of $16$, $98.3\%$ of Mat2Props outputs return URL placeholders (\texttt{i.imgur.com} / \texttt{materialsproject.org} links inside a markdown image embed) rather than structured numeric answers in JSON. Learning rate is not the cause (matching the same learning rate at low effective batch reproduces the failure), token-wise suppression at inference time did not help, and a smaller Qwen3-0.6B for an effective batch size of $16$ fails identically. At an effective batch size of $256$, the URL leak rate falls below $1\%$ and Mat2Props validity rises to $98.3\%$.

\begin{table}[ht]
\centering
\caption{\textbf{Effective batch size ablation.} Mat2Props validity and URL-leak rate (fraction of outputs that are URL placeholders or contain a leaked URL prior to property recovery). Leak rate is defined in Appendix~\ref{app:stage2_metric_defs}.}
\label{tab:s2-batchsize-app}
\small
\begin{tabular}{lrrl}
\toprule
Config & Mat2Props validity & URL leak \\
\midrule
\textbf{ALM Core} ($B_{\text{eff}}{=}256$, $r{=}128$)            & $\mathbf{98.3\%}$ & $\mathbf{<1\%}$  \\
$B_{\text{eff}}{=}16$, $\text{lr}{=}1\text{e-}4/1\text{e-}5$    & $4$--$7\%$        & $95\%$  \\
$B_{\text{eff}}{=}16$, $\text{lr}{=}2\text{e-}4/2\text{e-}5$    & $4\%$             & $96\%$ \\
Qwen3-0.6B at $B_{\text{eff}}{=}16$                             & $3.4\%$           & $97\%$\\
\bottomrule
\end{tabular}
\end{table}

\begin{figure}[ht]
\centering
\includegraphics{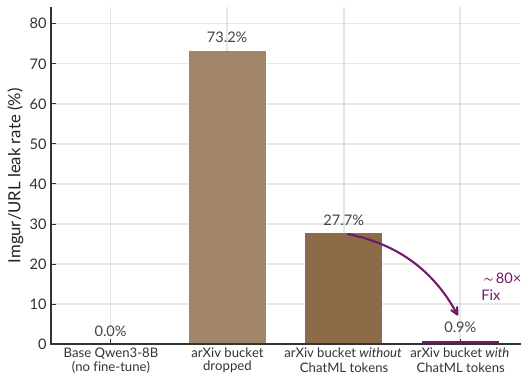}
\caption{\textbf{URL-leak rate is set by how the arXiv bucket is formatted.}}
\label{fig:leak_vs_weight}
\end{figure}

\paragraph{The leak is a base model prior, not data contamination.} Auditing the supervised text confirms it is essentially URL-free: the \texttt{describe}, \texttt{property\_apps}, \texttt{arxiv}, and \texttt{mascqa} buckets contain zero \texttt{imgur} occurrences and \texttt{camel} contains only two, across millions of rows. The leak is therefore a Qwen3-base pretraining prior that surfaces under fine-tuning-induced uncertainty, not a property of the training. Filtering $28$k LLM4Mat predictions per checkpoint for \texttt{i.imgur.com} confirms this (Figure~\ref{fig:leak_vs_weight}): the base model never leaks ($0.0\%$), including the arXiv text bucket in training data but without ChatML tokens leaks $27.7\%$, and dropping the arXiv bucket entirely makes the leak strictly worse ($73.2\%$). This is because removing the scientific-text exposure removes the text-only task contributions during training that keep the prior suppressed when learning on soft tokens. By including the dataset but formatting it with ChatML tokens, the leak rate issue is fixed.

%%% ===== frag_II_2_i.tex =====
\subsection{Data used to teach Atomistic Language Models to generate atoms}
\label{app:training_stage3}

The final training phase seeks to train the language-to-atomistic bridge to enable \textbf{ALM Edit} and \textbf{Gen} to steer the denoising of crystal structures.

\subsubsection{Seven-bucket training mixture}
\label{app:generation_data}

The generation mixture is drawn from seven task buckets at the weights of Table~\ref{tab:multi-bucket_buckets}: three unpaired text$\to$structure buckets (\texttt{describe}, a long GPT-Narrative$\to$structure caption; \texttt{csp}, a (formula, space-group) crystal-structure-prediction prompt; \texttt{ood}, an LLM-rephrased noisy-template paraphrase of the same structures) plus an application-class bucket (\texttt{app}), and three paired structure$+$text$\to$structure editing buckets (\texttt{atomtxt}, directional property edits; \texttt{polymorph}, polymorph$\to$lower-hull edits; \texttt{doping}, single-element substitution and strain). The \texttt{describe}, \texttt{csp}, and \texttt{ood} buckets share the same underlying GPT-Narrative~\cite{park_15_2024} structures ($\leq 20$ atoms, $K{=}8$ tokens) and differ only in how the prompt is templated; the editing buckets are constructed as explicit input/output structure pairs. The \texttt{atomtxt} or "directional" bucket consists of \textbf{ALM Bench} tasks and was built as described in Appendix~\ref{app:atomtxt_direction}.

\begin{table*}[ht]
\centering
\caption{\textbf{Seven-bucket generation training mixture.} Different bucket weightings $\pi$ were found to yield the best performance for \textbf{ALM Edit} and \textbf{Gen}. The weights indicate the likelihood for a row from each bucket to be drawn during training.}
\label{tab:multi-bucket_buckets}
\begin{tabular}{llrcc}
\toprule
Bucket & Task & Rows & $\pi$ (ALM Edit) & $\pi$ (ALM Gen) \\
\midrule
\texttt{describe}  & long narrative $\to$ structure                     & $1{,}352{,}176$ & $0.08$ & $0.10$ \\
\texttt{csp}       & (formula, sg) $\to$ structure                      & $1{,}352{,}176$ & $0.15$ & $0.20$ \\
\texttt{ood}       & noisy template $\to$ structure                     & $1{,}350{,}711$ & $0.08$ & $0.10$ \\
\texttt{app}       & application class $\to$ structure                  & $20{,}000$      & $0.04$ & $0.05$ \\
\texttt{atomtxt}   & (struct., prop.\ dir.) $\to$ struct. & $882{,}499$ & $0.40$ & $0.05^{\ddagger}$ \\
\texttt{polymorph} & (struct., polymorph) $\to$ struct.   & $545{,}145$ & $0.15$ & $0.25$ \\
\texttt{doping}    & (struct., edit) $\to$ structure      & $1{,}000{,}000$ & $0.10$ & $0.25$ \\
\bottomrule
\end{tabular}
\end{table*}

\subsubsection{Bucket-mixture ablations}
\label{sec:results_gen_ablations}\label{app:full_ablations}

\paragraph{Multi-bucket versus\ single-bucket.} Spreading weight across the seven buckets does slightly reduce the performance of certain individual tasks. In particular, adding the directional \textbf{ALM Bench} editing buckets at the cost of \texttt{csp} weight drops CSP match-rate by 44\%. We nonetheless use the multi-bucket mixture throughout, because it is the mixture that enables the directional-editing capability. On the other hand, increasing the weight on the \textbf{ALM Bench} (\texttt{atomtxt}) bucket from \textbf{ALM Gen}'s $5\%$ to $40\%$ lifts directional-editing performance significantly. The directional ceiling on the pooled bridge is therefore \emph{structural} rather than data-scale-bound; a different bridge architecture and finetuning method would lead to higher performance.

%%% ===== frag_III.tex =====
\section{ALM Bench}
\label{app:almbench}

\begin{figure}[ht]
  \centering
  \includegraphics[width=\linewidth]{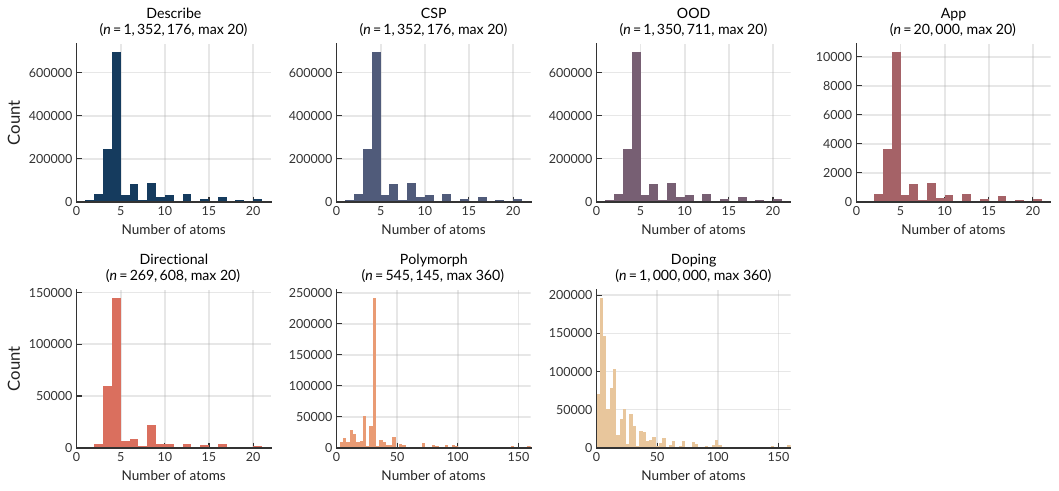}
  \caption{\textbf{Per-dataset atom-count distributions} for the structural training data. The structural buckets follow the LLM4Mat/GPT-Narratives distribution; the generation/editing pairs of Appendix~\ref{app:training_stage3} are capped at $\le\!20$ atoms per cell.}
  \label{fig:atomcount_hist}
\end{figure}

We introduce \textbf{ALM Bench}, a benchmark for \emph{conditional} crystal generation in which the conditioning carries both a structure and a natural-language instruction. It is the first benchmark to score the \textbf{atom${+}$text${\rightarrow}$atom} and \textbf{atom${+}$text${\rightarrow}$text} tasks. Here, \textbf{input(s)${\rightarrow}$output(s)} indicates a model capable of taking in a prompt of \textbf{input} modalities and generating responses in \textbf{output} modalities, a notation that will be used frequently in this section. In particular, \textbf{ALM Bench} asks a model to generate a new polymorph of the given crystal that \emph{satisfies a described intent}, scoring the result against a physical predictor or an LLM judge instead of against a single reference. \textbf{ALM Edit} is evaluated on each task here. The benchmark comprises seven scored tasks: \emph{directional editing}, CSP, \emph{application-consistency}, \emph{polymorph}, \emph{doping/substitution}, \emph{strain}, and \emph{text${\rightarrow}$structure recovery} (similar to the \emph{describe} bucket in \textbf{ALM Core}'s instruction tuning dataset) whose metric formulas are given in \secbadge{almCoreText}{app:stage2_metric_defs}. Every scored rate is reported as the \textbf{mean $\pm$ 95\% CI over $5\times N{=}200$ held-out prompts}. Fig.~\ref{fig:atomcount_hist} visualizes the distributions of structure sizes across the 7 buckets. Generated structures for all tasks (besides the structure recovery task) are briefly relaxed, as detailed in Appendix~\ref{app:atomtxt_direction}.

\subsection{Task~1: directional editing (atom${+}$text${\rightarrow}$atom)}
\label{app:atomtxt_direction}

The directional editing task requires \emph{paired} structures $(A, B)$ that share a composition but differ along one property in a known direction. Our final pairing technique builds these pairs entirely \emph{within a single parent dataset} so that DFT-calculated property labels have consistent levels of theory: it (i) clusters candidate structures by reduced formula, deduplicating near-identical polymorphs; (ii) attaches a MatterSim single-point energy-per-atom (and density / volume) label to every member of a cluster; and (iii) for each source structure emits \emph{both} a higher-$E$ and a lower-$E$ target drawn from the same cluster. Emitting both directions per source auto-balances the dataset across \texttt{higher}/\texttt{lower} prompts. The text prompt ends with ``\dots with a higher / lower $\langle$property$\rangle$,'' and the input and target materials are the paired structures, accordingly.

\paragraph{Scoring.} We MatterSim-relax both the input and the generated structure (full-cell relaxation, $f_{\max}{=}0.05$, $500$ steps, as it de-games a pure lattice rescale, which would otherwise relax straight back to the input volume), then score whether or not the MatterSim-calculated properties have the direction prompted for in language. For density and volume requests, a correct edit must \emph{additionally} be valid, composition-preserving, and structurally distinct from the input under \texttt{StructureMatcher} ($\texttt{ltol}{=}0.3$, $\texttt{stol}{=}0.5$, $\texttt{angle\_tol}{=}10$). The headline metric is $\texttt{direction\_correct\_rate} = (\#\text{moved the requested way}) / (\text{all scored candidates})$, with degenerate / NaN-property gens kept in the denominator. 

\paragraph{Sub-categories.} There are three types of directional editing tasks (within which are ``higher'' and ``lower'' subtasks):
\begin{itemize}
\item \textbf{formation energy} $E_f$: MatterSim total energy per atom, valid at fixed composition;
\item \textbf{density} $\rho$: relaxed mass / volume;
\item \textbf{volume} $V$: relaxed cell volume per atom;
\end{itemize}

The $E_f$ ``higher'' prompts, in particular, push \emph{against} the energy-lowering relaxation prior of the diffusion decoder, whereas a ``lower'' request takes advantage of the lower-energy distribution of unconditional generations.

\subsection{Task~2: application-consistency (text${\rightarrow}$atom)}
\label{app:app_consistency}

This task evaluates whether models can take in a generic application prompt that contains \emph{no} formula and \emph{no} space group (e.g.\ ``a porous metal hydride for lightweight hydrogen storage'') and generate crystals matching the requested application class. This is a text-conditional crystal generation task. 

\paragraph{Scoring.} A GPT-4o-mini or GPT-4o judge (which were ablated across to reveal roughly similar scoring preferences) reads a (prompt, formula, space group, $N_p$, elements, density, volume per atom, formation energy) tuple of the MatterSim-relaxed generation and returns a verdict on the $\{0,1,2\}$ scale. An invalid raw structure is forced to score $0$. The headline metric is the per-prompt mean consistency $\in [0,2]$, reported alongside the fraction scoring $2$ and the fraction judged inconsistent. The judge was independently calibrated against $150$ ground-truth (positive, negative-control) pairs across $8$ application classes, producing TP $90/90$ and TN $60/60$, confirming that a high inconsistent-verdict rate reflects the model, not a judge artifact. The dominant failure mode across application classes is that the model omits the required element entirely (a metal-hydride prompt returns a hydrogen-free structure, or a Li-ion-cathode prompt returns a lithium-free one).

\subsection{Task~3: polymorph (atom${+}$text${\rightarrow}$atom)}
\label{app:almbench_polymorph}

This task consists purely of inputting crystals and instructing the model to ``Generate a lower-energy polymorph.'' The generated crystal must have the \emph{same composition} as the input but a different geometry and a \emph{lower} energy. The \texttt{StructureMatcher} tolerances used to verify that the input and output structures are not identical are ($\texttt{ltol}{=}0.3$, $\texttt{stol}{=}0.5$, $\texttt{angle\_tol}{=}10$).

\paragraph{Scoring.} The primary metric is the fraction of generations whose MatterSim-relaxed total energy per atom is below the relaxed input's, \emph{gated} on the generation being valid, composition-preserving (reduced formula equal to the input), and structurally distinct from the input (\texttt{StructureMatcher.fit(input, gen)} is false). The pairs are MP-derived polymorph$\rightarrow$lower-$E_{\text{hull}}$ mappings (the \texttt{polymorph} training bucket of Table~\ref{tab:multi-bucket_buckets}), held out for evaluation.

\subsection{Task~4: doping substitution (atom${+}$text${\rightarrow}$atom)}
\label{app:almbench_doping}

This task consists of ``Substitute element $X$ with element $Y$'' prompts, wherein the model must perform a clean single-species substitution in a known input crystal.

\paragraph{Scoring.} A generation counts as successful only when the dopant is present, the donor is removed, the per-element ratio matches, \textit{the structure is valid}, and it is distinct from a naive relabel (the same positions up to numerical precision with element types swapped). Here, the per-element ratio of the target and generated structures can differ by up to $10\%$. 

\subsection{Task~5: strain (atom${+}$text${\rightarrow}$atom)}
\label{app:almbench_strain}

Extending the section above (Appendix~\ref{app:almbench_doping}), another metric to determine if a generated, doped crystal is realistic is calculating its strain compared to the original crystal. The doping edit alone (Task 4) ignores this lattice-deformation axis, so a dedicated strain task is added that shares the same parquet and prompt (``replace $X$ with $Y$'') but additionally scores whether the generated cell adjusts its volume by the correct amount.

\paragraph{Scoring.} In addition to the validity and element ratio metrics mentioned in the section above, the strain on the generated doped crystal should be similar to that of the labeled doped crystal. After an initial, short relaxation, the equilibrium strain of the generated crystal is compared to that of the reference, achieving success if they differ by less than 5\%. 

\subsection{Task~6: text${\rightarrow}$structure recovery (text${\rightarrow}$atom)}
\label{app:almbench_text2struct}

This task is to generate crystals from their structure descriptions, the inverse of the understanding \emph{describe} (structure$\rightarrow$text) task. We evaluate two prompt styles over the same held-out materials: \textbf{describe} (the verbose generation narrative from~\cite{rubungo_llm4mat-bench_2024} and~\cite{park_15_2024}, e.g.\ ``\dots\ has a tetragonal crystal system with space group P4/nmm'') and \textbf{OOD} (a terse, out-of-training-style spec, e.g.\ ``Show me As, Cu, Si, Ti in 119.56~\AA$^3$ and 8 [atoms]'').

\paragraph{Scoring.} Each generation is scored against the ground-truth structure on how well the elemental compositions match and an ordered \texttt{StructureMatcher} with $\texttt{ltol}{=}0.3$, $\texttt{stol}{=}0.5$, $\texttt{angle\_tol}{=}10$ to match the generation to the GT. Intuitively, this is a crystal structure prediction task with a much richer and more free-flowing language prompt. No relaxation is applied.

\begin{figure}[ht]
\centering
\includegraphics[width=\linewidth]{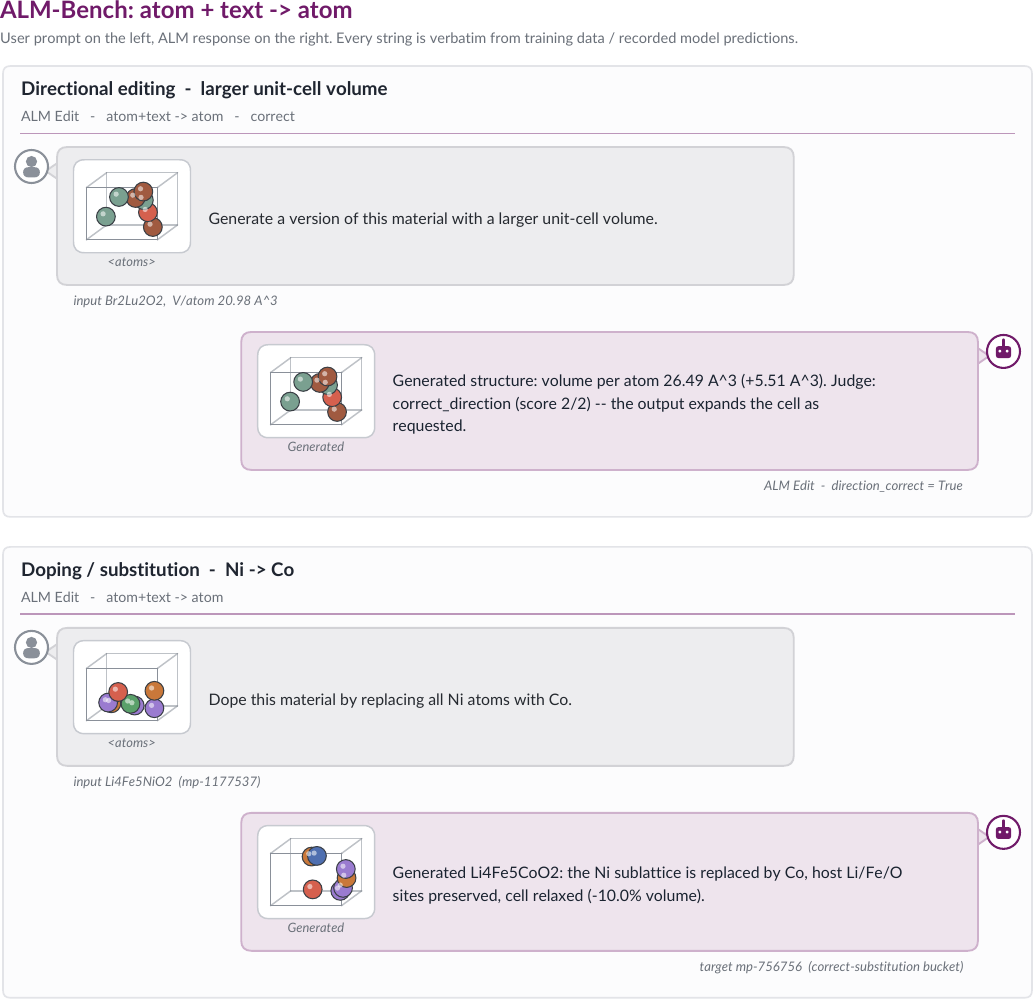}
\caption{\textbf{ALM-Bench chat examples.}}
\label{fig:chat_examples_almbench}
\end{figure}

\paragraph{Qualitative examples.} Figure~\ref{fig:chat_examples_almbench} shows verbatim \textbf{ALM Edit} transcripts (\textbf{ALM Bench} prompt on the left, model response on the right) across four representative ALM-Bench task types, including one deliberate failure case (a ``higher formation energy'' request, the arm that pushes against the relaxation prior) to make the scoring concrete.

\begin{figure}[ht]
\centering
\includegraphics[width=\linewidth]{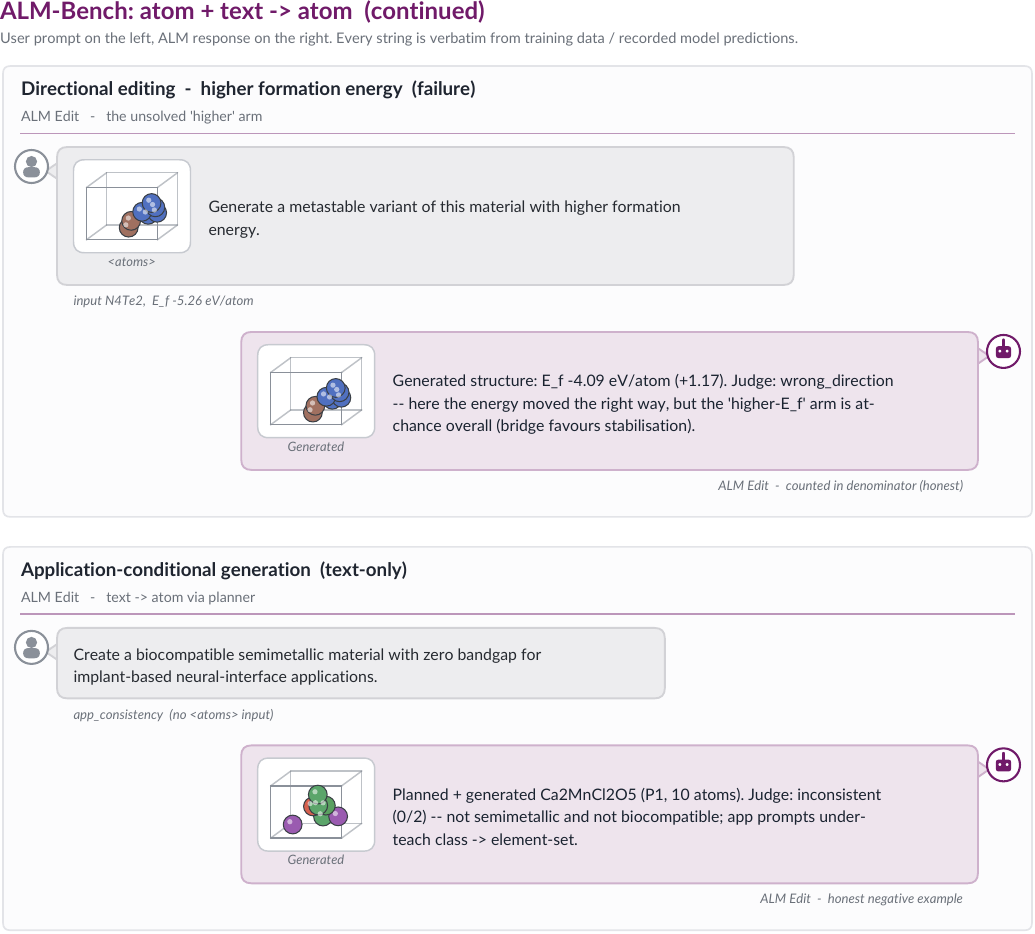}
\caption{\textbf{ALM-Bench chat examples (continued).}}
\label{fig:chat_examples_almbench_cont}
\end{figure}

%%% ===== frag_IV_i.tex =====
\section{Metrics}
\label{app:stage2_eval}\label{sec:s2-evals}

We lay out exact descriptions of each metric used to quantify performance in our work and enable \textbf{ALM Core, Edit,} and \textbf{Gen} to serve as comparable baselines for future work.  

\subsection{Property prediction metric definitions}
\label{app:stage2_metric_defs}

\paragraph{MAD\,$/$\,MAE.} For a regression property we report the ratio of the test-set Mean Absolute Deviation to the model's Mean Absolute Error,
\begin{equation}
\mathrm{MAD/MAE} \;=\;
\frac{\frac{1}{n}\sum_{i} \lvert y_i - \bar{y}\rvert}
     {\frac{1}{n}\sum_{i} \lvert \hat{y}_i - y_i\rvert},
\end{equation}
where $\bar{y}$ is the test-set mean. This is the scale-free skill score of LLM4Mat-Bench~\cite{rubungo_llm4mat-bench_2024}: \emph{higher is better}, $\mathrm{MAD/MAE}{=}1$ is no better than the mean predictor, and $\ge 5$ is the paper-defined ``good model'' threshold (ratios $<1$ are reported only for completeness). Being scale-free, the single $\ge 5$ threshold is comparable across all 28 property slices.

\paragraph{Mean Absolute Error (MAE).} Raw $\frac{1}{n}\sum_i \lvert \hat{y}_i - y_i\rvert$ in the property's native units (eV, eV/atom, g/cm$^3$, \AA, etc.); \emph{lower is better}. MAE is the quantity reported against external regression baselines (Mat2Props, MatText, GNoME-FE, the MatterChat regression tasks) where those baselines publish MAE rather than a skill ratio. As a result, we also show MAEs for selected property prediction tasks in Table~\ref{tab:stage2-llm4mat-mae}.

\begin{table*}[h!]
  \centering
  \caption{\textbf{Raw MAE on LLM4Mat-Bench}: ALM Core (canonical understanding $r{=}128$+IT) vs CIF structure-input baselines, in the physical units of the unit row. Lower is better; \textbf{bold} marks columns ALM Core wins. These are the absolute errors behind the skill ratios of Table~\ref{tab:stage2-llm4mat-mae}; generative-LLM baselines are omitted as there.}
  \label{tab:stage2-llm4mat-mae}
  \scriptsize
  \setlength{\tabcolsep}{3.2pt}
  \begin{tabular}{l rrrr rr rr rr rrrr}
  \toprule
   & \multicolumn{4}{c}{MP} & \multicolumn{2}{c}{JARVIS} & \multicolumn{2}{c}{OQMD} & \multicolumn{2}{c}{GNoME} & \multicolumn{4}{c}{hMOF} \\
  \cmidrule(lr){2-5}\cmidrule(lr){6-7}\cmidrule(lr){8-9}\cmidrule(lr){10-11}\cmidrule(lr){12-15}
  Model & $E_f$ & gap & $E_h$ & $\rho$ & $E_f$ & $E_h$ & $E_f$ & gap & $E_f$ & gap & void & LCD & PLD & CO$_2$ \\
   & {\tiny eV/at} & {\tiny eV} & {\tiny eV/at} & {\tiny g/cc} & {\tiny eV/at} & {\tiny eV/at} & {\tiny eV/at} & {\tiny eV} & {\tiny eV/at} & {\tiny eV} & {\tiny --} & {\tiny \AA} & {\tiny \AA} & {\tiny mol/kg} \\
  \midrule
  \rowcolor{almCoreBg}\textbf{ALM Core} & $0.070$ & $\mathbf{0.300}$ & $0.052$ & $0.212$ & $0.078$ & $0.063$ & $0.056$ & $0.093$ & $0.034$ & $0.051$ & $\mathbf{0.057}$ & $\mathbf{1.25}$ & $\mathbf{1.43}$ & $3.25$ \\
  CGCNN (CIF) & $0.123$ & $0.366$ & $0.058$ & $0.246$ & $0.063$ & $0.170$ & $0.033$ & $0.062$ & $0.014$ & $0.045$ & $0.062$ & $1.76$ & $2.04$ & $1.31$ \\
  MatBERT (CIF) & $0.091$ & $0.348$ & $0.059$ & $0.207$ & $0.084$ & $0.055$ & $0.053$ & $0.058$ & $0.020$ & $0.042$ & $0.110$ & $2.27$ & $2.42$ & $1.58$ \\
  LLM-Prop (CIF) & $0.070$ & $0.317$ & $0.103$ & $0.156$ & $0.066$ & $0.101$ & $0.039$ & $0.129$ & $0.017$ & $0.098$ & $0.095$ & $2.00$ & $2.50$ & $1.44$ \\
  \bottomrule
  \end{tabular}
\end{table*}

\paragraph{Leak rate.} The fraction of rows whose generation is flagged by the parser as a Markdown/image-URL emission (\verb|![...]| or an imgur/\texttt{materialsproject.org} link), rather than a materials answer. Leaks count as property prediction failure (never excluded from the denominator).

\paragraph{Accuracy (multiple choice / classification).} For MaScQA, Mat2MCQ, and the MatterChat classification tasks, fraction-correct accuracies are reported. For MatterChat's five classification tasks we use weighted-F1 (w-F1) to account for class imbalance (e.g.\ \emph{is\_metal}, \emph{is\_magnetic}). MaScQA's numerical-answer items are scored by MAE.

\paragraph{MatterChat 9-task scoring.} The MatterChat benchmark~\cite{tang_multimodal_2026} bundles five classification tasks (including crystal system, \texttt{is\_magnetic}, and \texttt{is\_metal}) scored by w-F1 ($\uparrow$) and three regression tasks (formation energy and energy-above-hull in eV/atom, band gap in eV) scored by MAE/RMSE ($\downarrow$), $n{=}1000$ test rows per task. ALM and all four MatterChat-paper variants are free-form text-generation models (cross-entropy on token probabilities, no per-task regression heads), so the comparison is output-mechanism-matched; the architectural difference is the input-side structural-encoder path and the dominant confound is per-task training exposure, not the head.

\begin{figure}[ht]
\centering
\includegraphics[width=\linewidth]{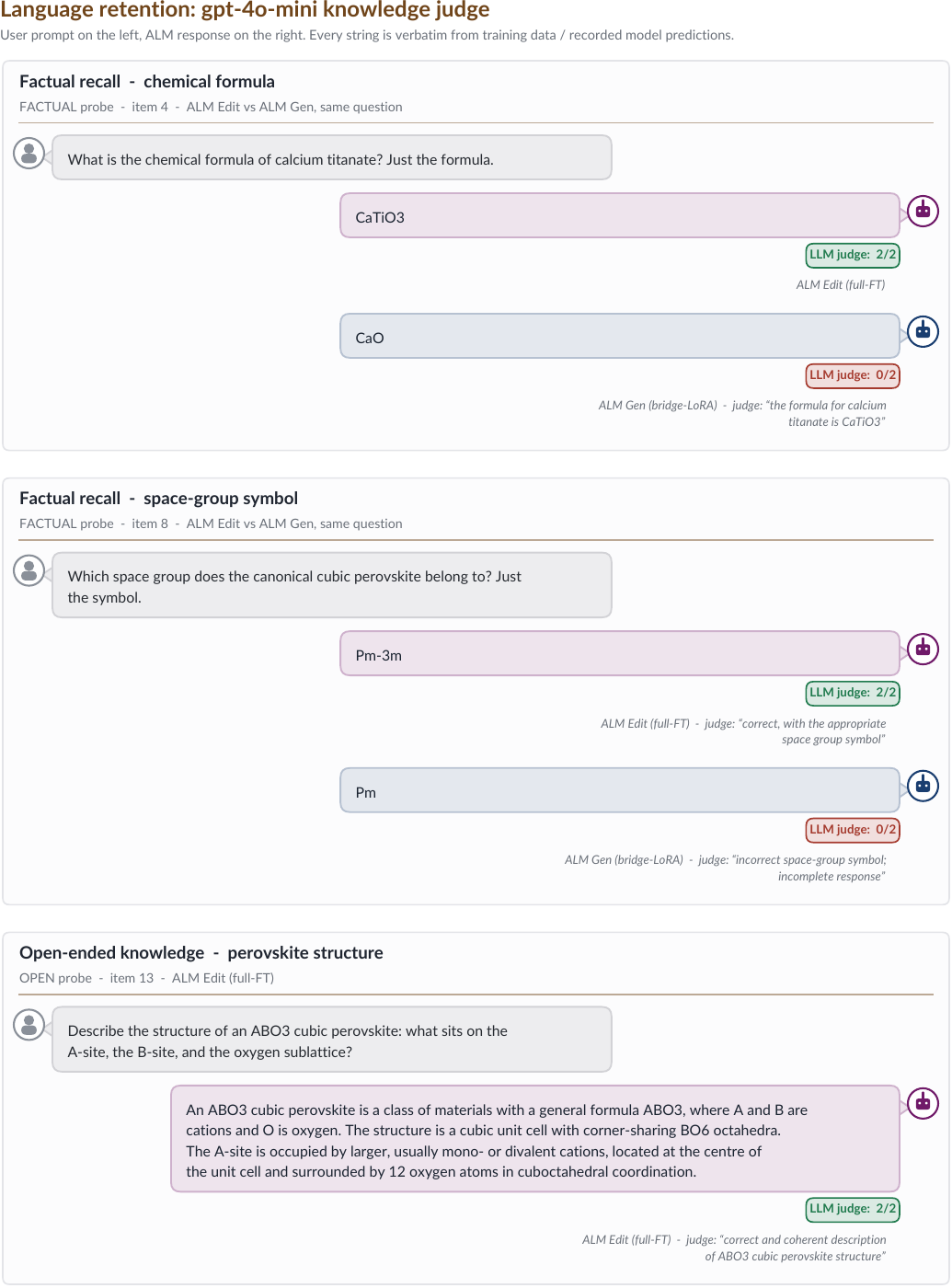}
\caption{\textbf{Materials-knowledge retention judge: verbatim graded exchanges.}}
\label{fig:chat_examples_retention}
\end{figure}

\begin{figure}[ht]
\centering
\includegraphics[width=\linewidth]{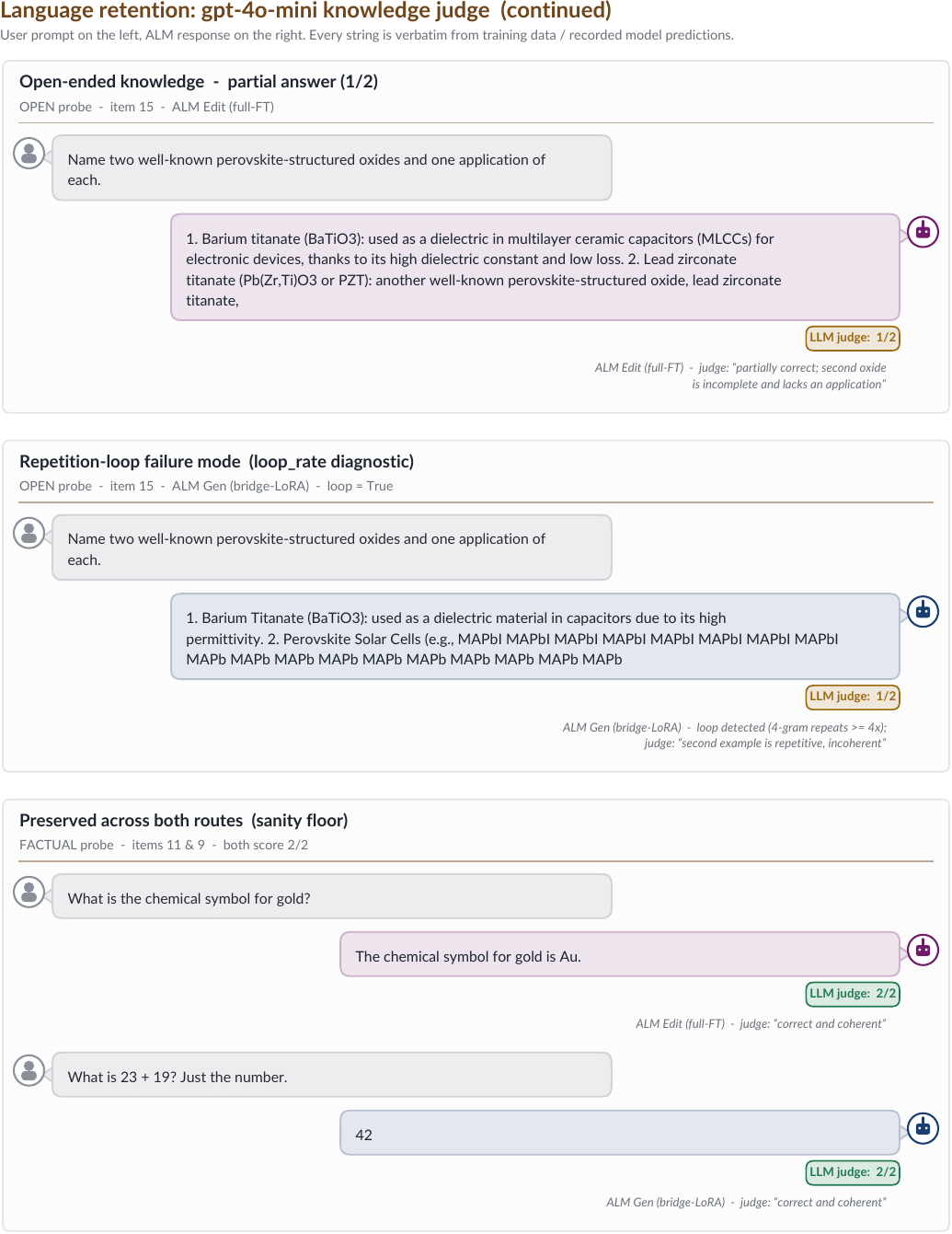}
\caption{\textbf{Materials-knowledge retention judge (continued).}}
\label{fig:chat_examples_retention_cont}
\end{figure}

\paragraph{LLM judge.} The LLM \emph{Judge} column of Table~\ref{tab:lang_retention_results} tests something the three accuracy benchmarks do not --- whether crystal-specialization erodes the model's ability to \emph{discuss} materials in natural language. We pose a fixed, in-house set of $190$ materials-science questions: $130$ short closed-form recall probes (e.g.\ the general perovskite formula $\mathrm{ABX_3}$; magnetite $=\mathrm{Fe_3O_4}$; the cubic-perovskite space group $Pm\bar{3}m$; the B-site location in $\mathrm{ABO_3}$) and $60$ open-ended free-text questions (e.g.\ describe the $\mathrm{ABO_3}$ perovskite structure; why B-site doping shifts the electronic structure; what makes a good thermoelectric). Each answer is decoded without thinking enabled ($64$ tokens for closed-form responses, $128$ for open-ended) and graded by \texttt{gpt-4o-mini} (temperature $0$) on a $0$--$2$ rubric: \textbf{2} $=$ correct \emph{and} coherent; \textbf{1} $=$ partially correct, with a minor error; \textbf{0} $=$ wrong, hallucinated, incoherent, empty, or stuck in a repetition loop. Degenerate or malformed responses are scored to $0$. Alongside the mean score (normalized to $[0,1]$ in Table~\ref{tab:lang_retention_results}) we track an independent degeneracy diagnostic, \emph{loop-rate}: the fraction of answers whose most-frequent $4$-gram repeats $\ge 4$ times. Figures~\ref{fig:chat_examples_retention} and~\ref{fig:chat_examples_retention_cont} demonstrate examples of the LLM judge. 

\subsection{Crystal structure prediction metric details}
\label{app:csp_metrics} 
This section fixes the exact, reproducible definitions of the CSP metrics reported for \textbf{ALM Edit} throughout the paper.

\paragraph{Matcher.} A generated structure is declared a match to its reference if pymatgen's \texttt{StructureMatcher} (wrapped by MatterGen's \texttt{OrderedStructureMatcher}) returns a fit at the CDVAE/CrystaLLM tolerances~\cite{xie_crystal_2022,antunes_crystal_2024}, \texttt{ltol}=0.3,\qquad \texttt{stol}=0.5,\qquad \texttt{angle\_tol}=10. The match test is \texttt{matcher.fit(gen,\,ref)} and the matched-pair RMSD is read from \texttt{matcher.get\_rms\_dist(gen,\,ref)}, which returns a \((\mathrm{rms},\,\mathrm{max\_dist})\) pair. Following CDVAE, we report the \(\mathrm{rms}\) component, in \AA, normalized as pymatgen normalizes by \(\big(V/N_p\big)^{1/3}\) (cell volume per atom).

\paragraph{Per-target aggregation: Match@$1$, Match@$K$, RMSE@$1$, RMSE@$K$.} For one reference we draw \(K\) i.i.d.\ generations \(\{g_1,\dots,g_K\}\) (the ``\(n{=}K\) generations per row'' of \Cref{tab:body_csp_mp20}) and compute:
\begin{itemize}\itemsep2pt
\item \textbf{Match@$1$} --- indicator that the \emph{first} generation \(g_1\) matches the reference. \textbf{RMSE@$1$} is the matched RMSD of \(g_1\) on a successful match. 
\item \textbf{Match@$K$} --- indicator that \emph{any} of the \(K\) generations matches the reference. \textbf{RMSE@$K$} is the \emph{minimum} RMSD over the matched subset \(\min_{\,i:\,g_i\text{ matches}}\mathrm{rms}(g_i,\mathrm{ref})\).
\end{itemize}
Reported Match@$1$/Match@$K$ are the means of these indicators over the \(n_{\text{rows}}\) targets; reported RMSE@$1$/RMSE@$K$ are means over matched rows only. CSP rows are reproduced verbatim from Table~1 of~\cite{seong_mcflow_2026}. 

%%% ===== frag_IV_iii.tex =====
\subsection{\textit{De novo} generation metric details}
\label{app:metric_dng}
\textbf{ALM Gen} is still a fundamentally conditional model, as it requires a prompt to the base language model to produce an output. However, its conditioning is deliberately weak, producing structures that are biased towards, but do not exactly follow, inputted prompts. To measure DNG performance, natural-language structural narratives are sampled from the GPT-Narratives~\cite{park_15_2024} dataset (\texttt{dft\_3d}, \texttt{mp\_3d\_2020}, \texttt{aflow2}, and \texttt{oqmd} narratives). 

\paragraph{Matcher.} Uniqueness $U$ and novelty $N$ are computed by the matcher against the generated set (for $U$) and against the MP-2020 reference set (for $N$). To determine whether or not a generated structure matches a set of reference structures, we use MatterGen's \texttt{DefaultDisorderedStructureMatcher} with default tolerances $\ell_{\mathrm{tol}}{=}0.2$, $s_{\mathrm{tol}}{=}0.3$, angle tolerance $5^{\circ}$, the same matcher used by \texttt{mattergen-evaluate} and by the LeMat-GenBench protocol. 

\paragraph{Stability.} The energy-above-hull $E_{\text{hull}}$ used by each stability gate in Table~\ref{tab:body_dng_mpts52} is the MP-2020-corrected hull energy from a MatterSim single-point relaxation. We report stability at three $E_h$ cut-offs, applied to the same generated structures:
\begin{itemize}
\setlength{\itemsep}{1pt}
\item \textbf{$E_{\text{hull}} \le 0.10$~eV/atom: Metastable ($MS$) $\to$ MSUN.} This is the gate that MatterGen~\cite{zeni_generative_2025}, and Crys-JEPA~\cite{liu_crysjepa_2026} label ``stable,'' which we show in Table~\ref{tab:dng_meta_mp20}.
\item \textbf{$E_{\text{hull}} \le 0.016$~eV/atom: Stable ($S$) $\to$ SUN.} This is the stricter CrystalReasoner~\cite{wu_crystalreasoner_2026} convention.
\item \textbf{$E_{\text{hull}} \le 0$: strict Stable $\to$ strict SUN.} This is the stability metric for LeMat-GenBench~\cite{betala_lemat-genbench_2026}. 
\end{itemize}
In addition, for LeMat-GenBench, $E_{\text{hull}}$ is evaluated and averaged by a three-MLIP ensemble (MACE-MP $+$ UMA $+$ Orb-V3), each building a self-consistent convex hull from the broader-chemistry \textbf{LeMat-Bulk} reference. All generations are pre-relaxed before scoring. 

\begin{table*}[ht]
\centering
\caption{\textbf{De-novo generation on MP-20, metastable-$MS$ convention} ($E_\mathrm{hull} \le 0.10$~eV/atom).}
\label{tab:dng_meta_mp20}
\footnotesize
\setlength{\tabcolsep}{3pt}
\begin{tabular}{lrrr}
\toprule
Method & $MS$ (\%) $\uparrow$ & $N$ (\%) $\uparrow$ & \textbf{MSUN} (\%) $\uparrow$ \\
\midrule
SymmCD$^{\dagger}$                                                              & $34.7\ci{1.4}$ & $85.1\ci{1.5}$ & $19.0\ci{0.9}$ \\
SGEquiDiff$^{\dagger}$                                                          & $46.5\ci{1.9}$ & $74.8\ci{1.2}$ & $23.5\ci{0.9}$ \\
DiffCSP++$^{\dagger}$                                                           & $39.7\ci{2.0}$ & $82.8\ci{1.0}$ & $23.8\ci{1.3}$ \\
CrysLLMGen-7B$^{\dagger}$                                                       & $35.1\ci{1.9}$ & $\underline{87.4}\ci{1.1}$ & $22.9\ci{0.9}$ \\
FlowMM$^{\dagger}$                                                              & $40.8\ci{2.0}$ & $83.1\ci{1.1}$ & $25.3\ci{1.6}$ \\
FlowLLM$^{\dagger}$                                                             & $36.5\ci{1.5}$ & $86.4\ci{1.5}$ & $25.1\ci{0.6}$ \\
CDVAE~\cite{xie_crystal_2022}$^{\dagger}$                                       & $29.9\ci{1.2}$ & $\mathbf{96.5}\ci{0.6}$ & $27.0\ci{1.3}$ \\
DiffCSP~\cite{jiao_crystal_2023}$^{\dagger}$                                    & $45.9\ci{1.8}$ & $83.6\ci{0.6}$ & $30.9\ci{1.8}$ \\
ADiT$^{\dagger}$                                                                & $69.5\ci{1.1}$ & $58.9\ci{0.9}$ & $30.3\ci{0.8}$ \\
MatterGen~\cite{zeni_generative_2025} (\textit{Crys-JEPA reprod.})$^{\dagger}$  & $47.0\ci{1.1}$ & $86.6\ci{1.0}$ & $34.6\ci{1.2}$ \\
Crys-JEPA-full$^{\dagger}$                                                      & $\mathbf{76.6}\ci{1.1}$ & $83.3\ci{1.1}$ & $\mathbf{45.2}\ci{1.4}$ \\
\midrule
MatterGen-Base (our reprod., $g{=}0$ uncond)                                    & $\underline{71.9}\ci{1.2}$ & $62.5\ci{0.8}$ & $36.8\ci{1.1}$ \\
\rowcolor{almGenBg}\textbf{ALM Gen} ($g{=}0.5$) & $68.3\ci{2.6}$ & $67.8\ci{1.9}$ & $\underline{39.0}\ci{3.5}$ \\
\rowcolor{almGenBg}\textbf{ALM Gen} ($g{=}0.5$) $+$ \textbf{T2C-FK} & $65.1\ci{3.0}$ & $58.9\ci{3.9}$ & $23.9\ci{1.9}$ \\
\bottomrule
\end{tabular}
\end{table*}

\subsection{Representational alignment metrics}
\label{app:repr_align_metrics}

The representational analysis of Figure~\ref{fig:discussion_scaling_law}B quantifies how information is shared across \textbf{ALM Edit}'s internal latent spaces. For $N{=}2{,}000$ \textbf{ALM Bench} prompts passed through \textbf{ALM Edit}, four representations $f$ are extracted: the frozen OrbV3 node-wise embeddings, the output of the MLP that projects each token into the LLM embedding space (i.e., turning them into soft tokens), the language model's $K{=}8$ atomistic token embeddings, and the language-to-atomistic producer embedding output $\mathbf{C}$. Their pairwise alignment is measured with two complementary metrics, both bounded to $[0,1]$: information imbalance (global, asymmetric)~\cite{glielmo_ranking_2022} and CKNNA (local)~\cite{huh_platonic_2024}.

\paragraph{Information imbalance.} This is an asymmetric measure of how much more information one representation holds than another~\cite{glielmo_ranking_2022}, built on the premise that a representation's nearest-neighbor ranking is more informative than per-coordinate distances. Let $r^f_{ij}$ be the nearest-neighbor rank of $f(x_j)$ with respect to $f(x_i)$ (rank $1$ is the nearest), and let $c_f \approx r^f/N$ be the associated copula (cumulative) variable. The information imbalance from representation $f$ to $g$ is
\begin{equation}
\Delta(f \rightarrow g) \;=\; 2 \lim_{\epsilon \rightarrow 0} \big\langle\, c_g \,\big|\, c_f = \epsilon \,\big\rangle,
\end{equation}
the average $g$-rank of the points that are nearest neighbors in $f$. Because $r^f_{ij}\neq r^g_{ij}$, the pair $\big(\Delta(f\!\rightarrow\!g),\,\Delta(g\!\rightarrow\!f)\big)$ is asymmetric: both near $0$ (bottom-left of an II plot) means the two spaces encode identical information; both near $1$ (top-right) means orthogonal information; $\Delta(f\!\rightarrow\!g)\!\approx\!0$ with $\Delta(g\!\rightarrow\!f)\!\approx\!1$ (top-left) means $g$ is contained within $f$; and $\Delta\!\approx\!0.5$ on both axes means the spaces share information without subsuming one another. Ranks are computed by exact cosine $k$-nearest-neighbor ordering over all $N$ points (full-rank, $\mathrm{ii}\text{-}k{=}50$).

\paragraph{Centered kernel nearest-neighbor alignment (CKNNA).} This is a local latent space similarity that is high when two representations agree on which points are mutual nearest neighbors~\cite{huh_platonic_2024}. With centered inner-product kernels $\bar K_{ij}=\langle f(x_i),f(x_j)\rangle - \mathbb{E}\!\left[\langle f(x_i),f(x_j)\rangle\right]$ and $\bar L_{ij}$ defined analogously for $g$, alignment is restricted to mutual $k$-nearest-neighbor pairs,
\begin{equation}
\mathrm{Align}(K,L) \;=\; \sum_{i,j} \alpha(i,j)\,\bar K_{ij}\,\bar L_{ij},
\end{equation}
where $\alpha(i,j){=}1$ iff $f(x_j)$ is among the $k$ nearest neighbors of $f(x_i)$ \emph{and} $g(y_j)$ is among the $k$ nearest neighbors of $g(y_i)$, and $0$ otherwise; CKNNA is the normalized form $\mathrm{Align}(K,L)\,/\,\sqrt{\mathrm{Align}(K,K)\,\mathrm{Align}(L,L)}$, bounded to $[0,1]$. We use $k{=}25$. II and CKNNA agree across all representation pairs (Figure~\ref{fig:discussion_scaling_law}B), so the global information-content reading is corroborated by local neighborhood structure.

% TODO(user): insert crystal-visualization figure here --- side-by-side % renderings of representative ALM Gen de-novo structures at the g=0.5 % operating point (e.g. via ASE/VESTA), illustrating the stable-and-novel % outputs the MSUN/SUN-strict gates credit. Use \label{fig:dng_crystal_viz}.

\end{document}